\documentclass[lettersize,journal]{IEEEtran}
\usepackage{amsmath,amsfonts}
\usepackage{amssymb}
\usepackage{mathtools}
\usepackage{algorithmic}
\usepackage{algorithm}
\usepackage{array}
\usepackage[caption=false,font=normalsize,labelfont=sf,textfont=sf]{subfig}
\usepackage{textcomp}
\usepackage{stfloats}
\usepackage{url}
\usepackage{verbatim}
\usepackage{graphicx}
\usepackage{pifont}
\usepackage{cite}
\usepackage{hyperref}
\usepackage{cleveref} 
\usepackage{amsmath}
\usepackage{booktabs}
\usepackage{color}
\usepackage{xspace}
\usepackage{graphicx}
\usepackage{adjustbox}
\usepackage{tabularx}
\usepackage{pifont}
\newcommand{\cmark}{\ding{51}}
\newcommand{\xmark}{\ding{55}}
\usepackage[table]{xcolor}
\definecolor{thirdcolor}{RGB}{0,153,76} 
\usepackage{multirow}

\begin{document}

\title{HDRAgent: An Agentic Framework for Multi-Exposure HDR Imaging}

\author{Weiyu Zhou, Tao Hu, Yijian Wang, Xiaogang Xu, Ruixing Wang and Qingsen Yan
\thanks{Weiyu Zhou, Tao Hu, Yijian Wang and Qingsen Yan are with the School of Computer Science, Northwestern Polytechnical University, Xi'an 710129, China. (e-mail: weiyvzhou@gmail.com; taohu@mail.nwpu.edu.cn; wangyijian1017@163.com; qingsenyan@nwpu.edu.cn).}
\thanks{Qingsen Yan is also with the Shenzhen Research Institute, Northwestern Polytechnical University, Shenzhen, China.}
\thanks{Xiaogang Xu is with Zhejiang University, Hangzhou, China.}
\thanks{Ruixing Wang is with the Camera Group, DJI, Shenzhen, China.}
}


\maketitle

\begin{abstract}
Most existing multi-exposure HDR methods follow a fixed feed-forward reconstruction paradigm, making them prone to ghosting artifacts in complex dynamic scenes. To address this issue, we propose HDRAgent, the first agent-driven framework for HDR imaging, which adaptively selects reconstruction strategies according to the current scene conditions. 
Specifically, to provide scene-specific prior knowledge, we introduce a fine-grained contextual knowledge matching (FCM) module. This module leverages multimodal large language model (MLLM)-derived scene perception to retrieve relevant historical cases and tool knowledge, organizing them into structured evidence for MLLM-based adaptive tool scheduling.
In addition, we propose a perception--distortion feedback mechanism that transforms post-execution quality assessment and artifact diagnosis into structured feedback, which is accumulated in historical memory to help subsequent contextual knowledge refinement and strategy selection.
Furthermore, considering that extreme motion can invalidate alignment methods, we design an agent-guided generative alignment strategy that uses MLLM-based dynamic-region parsing to reconstruct unreliable contents in non-reference frames under reference-frame guidance. 
Experiments demonstrate that HDRAgent effectively reduces ghosting and local artifacts while achieving competitive or superior objective performance and visual quality.
\end{abstract}

\begin{IEEEkeywords}
High dynamic range, multi-exposed imaging, agent-driven reconstruction, multimodal large language model 
\end{IEEEkeywords}

\section{Introduction}
\IEEEPARstart{M}{ulti-exposure} high dynamic range (HDR) imaging reconstructs HDR scenes by fusing complementary information from images captured under different exposure settings~\cite{debevec2023recovering,kalantari2017deep}. Compared with single-image HDR reconstruction, multi-exposure imaging can better preserve shadow details and highlight structures by exploiting exposure diversity~\cite{debevec2023recovering,Chen_2025_CVPR_UltraFusion}. However, robust HDR reconstruction in dynamic scenes remains challenging, as saturated highlights, large object motion, and occlusions introduce unreliable and inconsistent observations across input exposures, often leading to ghosting and local fusion artifacts~\cite{hu2013hdr,wu2018deep,yan2019attention}. Therefore, how to adaptively exploit cross-exposure information and suppress ghosting under complex dynamic conditions remains a central problem in multi-exposure HDR imaging.

\begin{figure}[t]
\centering
\includegraphics[width=\columnwidth]{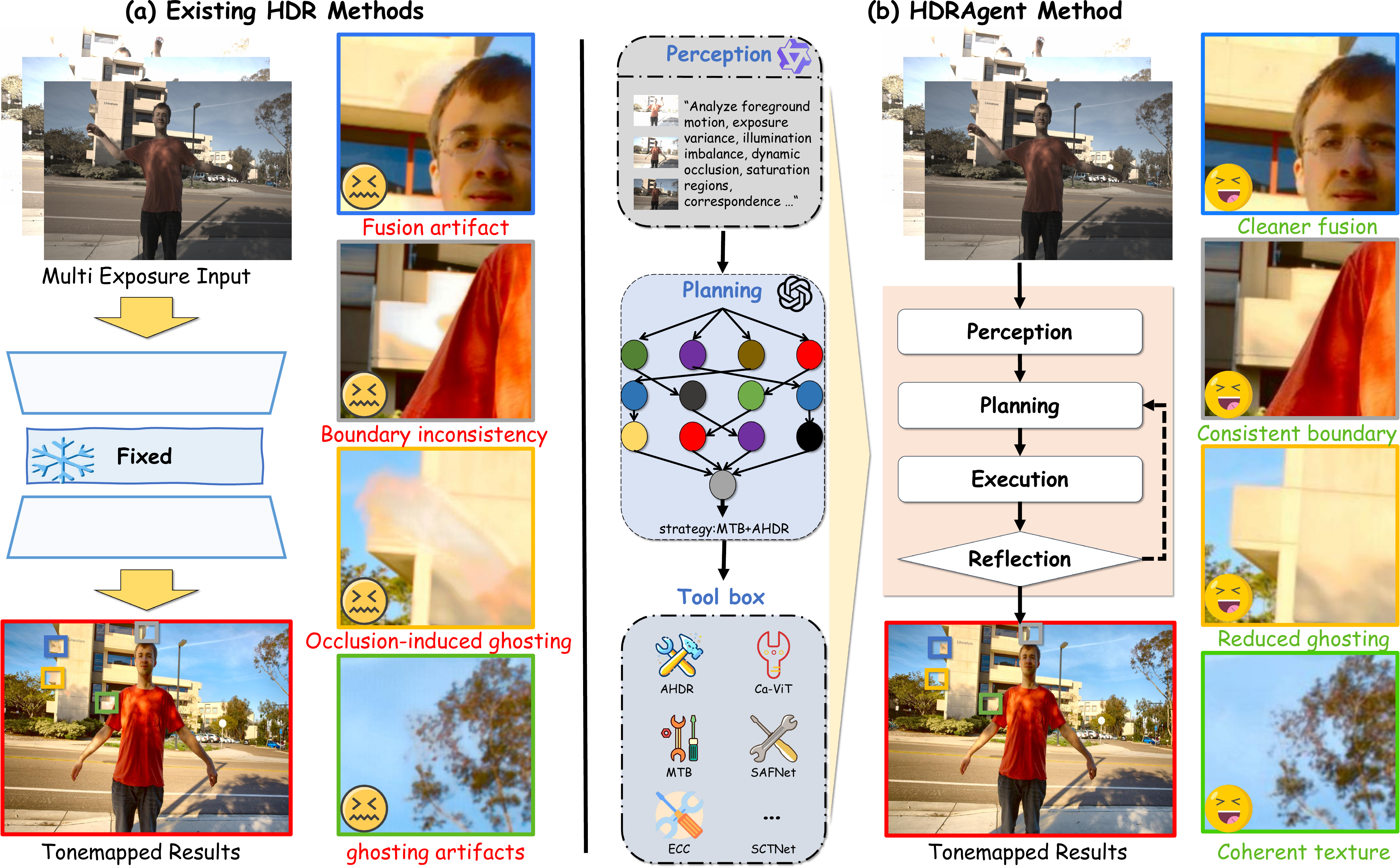}
\caption{Paradigm comparison between existing multi-exposure HDR methods and HDRAgent. Existing methods typically perform alignment and fusion along a fixed feed-forward reconstruction paradigm, making them difficult to adapt to diverse dynamic scene conditions. HDRAgent reformulates HDR reconstruction as an agent-driven iterative process that integrates perception, planning, execution, and feedback, enabling scene-adaptive deghosting and progressive artifact correction.}
\label{fig:paradigm}
\vspace{-0.5cm}
\end{figure}

Early multi-exposure HDR methods are usually based on physical imaging models and hand-crafted registration-and-fusion pipelines, such as camera response recovery, motion detection, and patch-based alignment~\cite{debevec2023recovering,sen2012robust,hu2013hdr}. These methods provide certain physical interpretability, but they often rely on brightness consistency, locally matchable textures, and limited-motion assumptions, making them prone to alignment failures and ghosting artifacts in complex dynamic scenes~\cite{grosch2006fast,lee2014ghost,hu2013hdr}. In recent years, deep learning methods have significantly advanced multi-exposure HDR imaging. CNN-based methods learn cross-exposure correspondence modeling and feature-domain fusion in an end-to-end manner to alleviate motion and exposure discrepancies~\cite{kalantari2017deep,wu2018deep,yan2019attention,prabhakar2020towards}. Transformer-based architectures further enhance long-range dependency modeling and global feature interaction~\cite{liu2022ghost,song2022selective}. Meanwhile, generative models and visual priors from large-scale models provide new possibilities for restoring severely degraded regions~\cite{niu2021hdr,yan2023toward,hu2024generating,Wang_2025_CVPR_LEDiff,Lin_2024_ECCV_DiffBIR}. Despite these advances, most existing methods still embed their alignment, fusion, or generative capability into a fixed feed-forward reconstruction paradigm. Once trained or designed, different inputs are processed by the same predefined inference procedure, as illustrated in Fig.~\ref{fig:paradigm}(a). As a result, a single fixed inference path can hardly adapt to diverse scene conditions involving different combinations of motion, saturation, occlusion, and exposure imbalance.

\begin{figure}[t]
    \centering
    \includegraphics[width=1\linewidth]{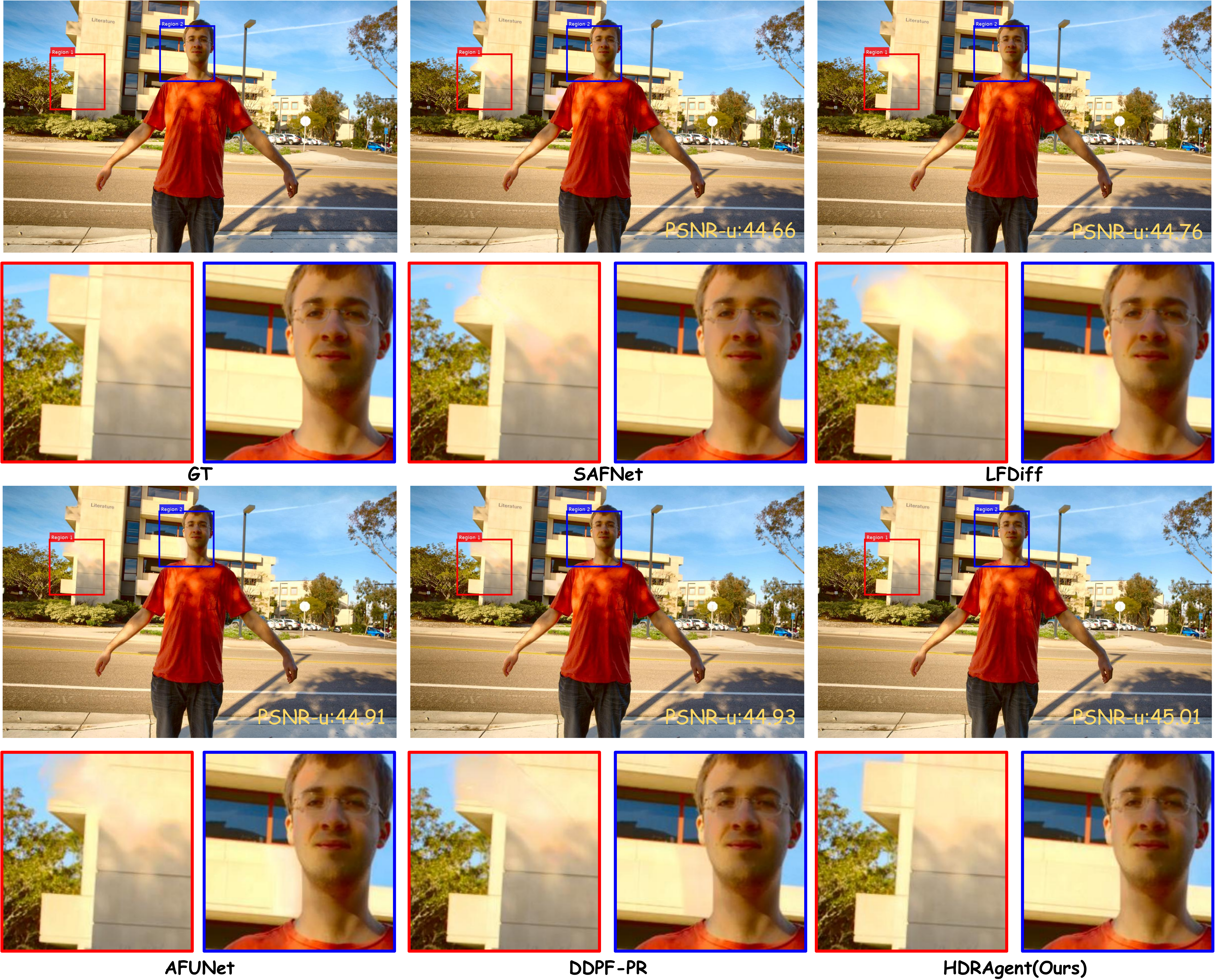} 
    \caption{Visual motivation of HDRAgent. Although recent state-of-the-art methods, such as SAFNet~\cite{kong2024safnet}, LFDiff~\cite{hu2024generating}, AFUNet~\cite{li2025afunet}, and DDPF-PR~\cite{zhou2026high}, achieve favorable quantitative performance, their zoomed-in results may still exhibit ghosting and local fusion artifacts in challenging dynamic regions. In contrast, HDRAgent produces cleaner local structures by adaptively selecting reconstruction strategies and performing feedback-guided refinement.}
    \label{fig:intro2} 
\vspace{-0.5cm}
\end{figure}

This limitation is further amplified by the spatial heterogeneity of dynamic scene degradations. Regions with mild displacement and discriminative textures can often be handled by standard alignment-and-fusion strategies, since reliable cross-exposure correspondences remain accessible. Regions with larger spatial displacement or broader contextual ambiguity require stronger global context modeling, such as transformer-based long-range feature interaction, to capture non-local dependencies. Regions affected by severe saturation, occlusion, or extreme structural inconsistency may lack reliable cross-exposure content, making reconstruction more dependent on generative structure completion than either local alignment or global correspondence modeling. These heterogeneous requirements call for scene-adaptive strategy selection rather than a single fixed reconstruction pipeline. Moreover, appropriate strategy selection alone does not guarantee artifact-free reconstruction. As shown in Fig.~\ref{fig:intro2}, several recent methods achieve favorable metric scores, yet visible ghosting and local fusion artifacts may still remain in dynamic regions. This indicates that image-level quantitative metrics may obscure small but perceptually significant local defects, motivating a feedback mechanism that inspects reconstruction outcomes and provides corrective cues for subsequent strategy refinement.

Recent multimodal large language models (MLLMs) have demonstrated strong capabilities in visual understanding and cross-modal reasoning~\cite{openai2023gpt4v,geminiteam2023gemini,liu2023llava,bai2023qwenvl}. Meanwhile, agentic systems show that MLLM can serve as task controllers that integrate perception, planning, tool invocation, and feedback correction into a unified reasoning process~\cite{yao2023react,shen2023hugginggpt,wu2023visualchatgpt}. Recent works have introduced this paradigm into image restoration, where degradation perception, task scheduling, and expert model selection are used to handle complex degradation problems~\cite{chen2024restoreagent,zhu2025agenticir,lin2025jarvisir,wang2026paagent}. These advances suggest a promising direction for multi-exposure HDR reconstruction: instead of applying a fixed reconstruction pipeline to all inputs, an MLLM-driven agent can serve as a high-level controller that interprets scene conditions, selects suitable reconstruction tools, and refines decisions according to reconstruction feedback.

Motivated by the above analysis, we propose HDRAgent, an agent-driven framework for multi-exposure HDR reconstruction. Specifically, an MLLM first describes the multi-exposure inputs from the perspectives of foreground motion, exposure variation, saturation, and occlusion. Based on this description, we introduce a fine-grained contextual knowledge matching (FCM) module to organize scene descriptions, candidate reconstruction tools, historical failure cases, and feedback-updated reconstruction experience as structured decision evidence. An MLLM dynamic router then selects suitable alignment and fusion strategies according to the input description and accumulated contextual knowledge. To correct residual ghosting and local artifacts after tool execution, we further introduce a perception--distortion feedback mechanism that transforms artifact diagnosis into contextual knowledge for iterative correction on the original multi-exposure inputs. Moreover, for extreme motion cases where alignment becomes unreliable, we design an agent-guided generative alignment strategy to restore reference-consistent structures in non-reference frames. Specifically, the MLLM identifies dynamic foreground objects, generates segmentation prompts, and invokes segmentation and generative tools to obtain correction masks and perform reference-conditioned masked generation. Through this agentic reconstruction process, HDRAgent effectively handles saturated highlights, occlusions, and severe motion, achieving competitive or superior objective and visual quality.

The main contributions of our work can be summarized as follows:
\begin{itemize}
    \item We propose HDRAgent, to the best of our knowledge the first agent-driven framework for multi-exposure HDR reconstruction, which reformulates fixed feed-forward alignment and fusion as a perception--planning--execution--feedback dynamic reconstruction process.

    \item We introduce a fine-grained contextual knowledge matching (FCM) module that organizes scene evidence, tool applicability, and historical feedback into structured decision evidence, enabling an MLLM-based dynamic router to adaptively select reconstruction strategies.

    \item We design a perception--distortion feedback mechanism for structured artifact diagnosis and iterative refinement, together with an agent-guided generative alignment strategy that handles extreme-motion regions via reference-conditioned masked generation.

    \item Extensive experiments on benchmark datasets demonstrate that HDRAgent effectively reduces ghosting and local artifacts, achieving competitive or superior objective performance and visual quality in complex dynamic scenes.
\end{itemize}

\section{Related Work}
\subsection{Multi-exposure HDR Imaging}
Traditional multi-exposure HDR methods typically rely on physical imaging models and hand-crafted preprocessing pipelines, including radiometric recovery, exposure fusion, and registration or patch matching for dynamic scenes~\cite{debevec2023recovering,mertens2007exposure,gallo2009artifact,ward2003fast,sen2012robust,hu2013hdr}. Although interpretable, these methods heavily depend on preprocessing accuracy and locally matchable textures, making them less effective under large motion, occlusion, and saturation~\cite{hu2013hdr,wu2018deep}. In recent years, deep learning methods have become the dominant solution for multi-exposure HDR reconstruction. CNN-based methods learn feature-domain alignment and fusion in an end-to-end manner, and attention mechanisms are further introduced to enhance region selection and cross-exposure interaction for dynamic-scene deghosting~\cite{kalantari2017deep,wu2018deep,yan2019attention,chen2022attention,prabhakar2020towards}. Transformer-based methods further exploit global contextual modeling and selective region interaction to strengthen long-range correspondence reasoning~\cite{liu2022ghost,song2022selective,tel2023alignment}. Meanwhile, optical-flow-guided approaches explicitly estimate inter-frame motion for misalignment compensation~\cite{xu2024hdrflow,kong2024safnet}. Generative methods, including GANs and diffusion models, have also been explored to recover missing content and improve perceptual quality in severely degraded regions~\cite{niu2021hdr,yan2023toward,yan2024dynamic,Wang_2025_CVPR_LEDiff,Chen_2025_CVPR_UltraFusion}. Nevertheless, most existing methods still follow a fixed one-shot alignment-and-fusion paradigm, lacking dynamic planning for different motion, saturation, and occlusion conditions, as well as explicit feedback correction for residual ghosting.

\subsection{Agentic Image Restoration}
Recent advances in LLM- and VLM-based agentic systems have driven image restoration from fixed feed-forward inference toward a dynamic decision-making paradigm. Unlike conventional single-degradation restoration networks and All-in-One restoration models~\cite{zamir2022restormer,PromptIR,Conde_2024_ECCV_InstructIR}, agentic methods perceive input degradations, plan restoration strategies, and progressively refine results through tool invocation and quality feedback. Existing automatic and instruction-guided restoration methods provide conditional information through degradation recognition, blind quality assessment, or text prompts~\cite{PromptIR,Conde_2024_ECCV_InstructIR,jiang2024autodir}. More recent agentic frameworks further formulate image restoration as a perception--planning--execution--feedback process, showing strong adaptability in complex image restoration, super-resolution, deraining, and mixed-degradation removal~\cite{yao2023react,shen2023hugginggpt,wu2023visualchatgpt,Chen_2024_NeurIPS_RestoreAgent,Zhu_2025_ICLR_AgenticIR,Jiang_2025_arxiv_MAIR,Li_2025_arxiv_HybridAgent,Zuo_2025_NeurIPS_4KAgent,Lin_2025_CVPR_JarvisIR}.

These advances motivate us to introduce the agentic restoration paradigm into multi-exposure HDR imaging. However, existing methods mainly focus on single-image degradation or general mixed-degradation restoration, without considering the radiometric relationships, cross-exposure correspondences, and dynamic ghosting issues in multi-exposure inputs. To this end, HDRAgent formulates multi-exposure HDR reconstruction as an agentic decision-making process for cross-exposure dynamic scenes, achieving dynamic ghost suppression through scene-adaptive context and feedback updating.

\begin{figure*}[ht!]
\centering
\includegraphics[width=\textwidth]{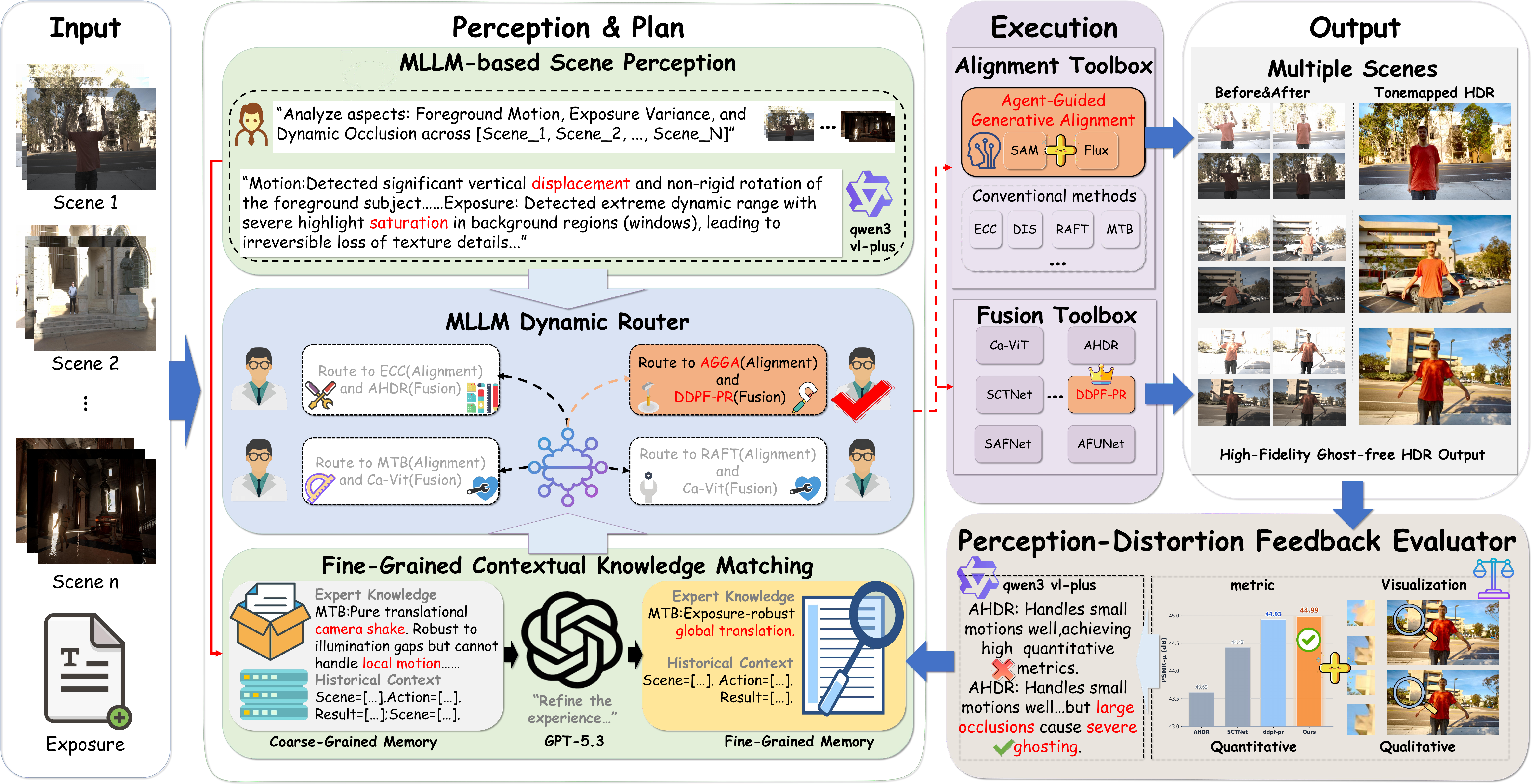}
\caption{Overall architecture of the proposed HDRAgent framework for multi-exposure HDR reconstruction. HDRAgent first uses MLLM-based scene perception to analyze motion, exposure variation, saturation, and occlusion, and then performs fine-grained contextual knowledge matching to organize scene evidence, tool applicability, and historical feedback. The dynamic router selects suitable reconstruction tools according to the contextual evidence, with agent-guided generative alignment handling local large-motion and high-mismatch regions. The perception--distortion evaluator further detects residual ghosting and local artifacts, providing structured feedback for iterative reconstruction refinement.}
\label{overview}
\end{figure*}

\section{Methodology}

\subsection{Overview of Proposed Framework}

The overall architecture of HDRAgent is shown in Fig.~\ref{overview}. 
Given multi-exposure LDR images $\mathcal{I}=\{I_i\}_{i=1}^{N}$ and exposure information $\mathcal{E}=\{e_i\}_{i=1}^{N}$, the goal is to reconstruct a ghost-free HDR image $\hat{H}$ with faithful radiance and consistent local structures. 
Compared with fixed feed-forward HDR pipelines, HDRAgent formulates reconstruction as a closed-loop agentic process that integrates scene perception, contextual knowledge matching, dynamic routing, toolbox-based execution, and feedback-driven correction.

HDRAgent first employs Qwen3-VL-Plus~\cite{qwen3vl} to analyze motion, exposure variation, saturation, occlusion, and cross-exposure mismatch. 
FCM then organizes the scene state, expert tool knowledge, historical context, and feedback memory into structured decision evidence. 
Given this evidence, Qwen3.6-Plus~\cite{qwen36plus} dynamically selects suitable alignment and fusion/reconstruction tools. 
The reconstructed result is further assessed by a perception--distortion feedback evaluator, whose feedback is written back into contextual memory to refine subsequent routing and correction. 
This closed-loop design enables HDRAgent to reduce ghosting and local fusion artifacts in complex dynamic scenes.

\subsection{Fine-Grained Contextual Knowledge Matching}

HDRAgent employs FCM to organize scene perception results, expert tool knowledge, and historical feedback into structured contextual evidence, which provides decision support for subsequent dynamic routing.

Given the multi-exposure inputs $\mathcal{I}=\{I_i\}_{i=1}^{N}$, exposure information $\mathcal{E}=\{e_i\}_{i=1}^{N}$~\cite{kalantari2017deep,hu2013hdr,tel2023alignment}, and a scene-analysis instruction $\mathcal{P}_{\mathrm{scene}}$, the scene perception module first produces an input-specific scene-state description:
\begin{equation}
S = \Phi_{\mathrm{perc}}(\mathcal{I}, \mathcal{E}, \mathcal{P}_{\mathrm{scene}}),
\label{eq:scene_perception}
\end{equation}
where $\Phi_{\mathrm{perc}}$ denotes the MLLM-based scene perception process instantiated with Qwen3-VL-Plus~\cite{qwen3vl} and guided by the scene-analysis instruction. $S$ denotes the scene-state description produced by the MLLM, including global motion dynamics, exposure distribution, saturation locations, and occlusion changes.

After obtaining the scene state, the FCM module organizes the scene state $S$, expert tool knowledge base $\mathcal{K}$, and historical feedback $\mathcal{H}_{t-1}$ into structured contextual evidence for the $t$-th reconstruction round under the context organization instruction $\mathcal{P}_{\mathrm{ctx}}$:
\begin{equation}
\begin{aligned}
C_t &= \Phi_{\mathrm{fcm}}(S,\mathcal{P}_{\mathrm{ctx}},\mathcal{K},\mathcal{H}_{t-1}) \\
    &= \{C_t^{\mathrm{scene}},C_t^{\mathrm{tool}},C_t^{\mathrm{fb}}\},
\end{aligned}
\label{eq:fcm}
\end{equation}
where $\Phi_{\mathrm{fcm}}$ denotes the contextual knowledge matching process instantiated with GPT-5.3~\cite{openai_gpt53}. $\mathcal{P}_{\mathrm{ctx}}$ constrains the structure and focus of the generated contextual evidence. $\mathcal{K}$ denotes the expert tool knowledge base, which records the functions, applicability conditions, and potential failure risks of candidate alignment and fusion tools. In this process, GPT-5.3 first reformulates the scene-state description $S$ into scene-conditioned reconstruction evidence $C_t^{\mathrm{scene}}$, then matches the expert tool knowledge base $\mathcal{K}$ with $C_t^{\mathrm{scene}}$ to obtain tool applicability evidence $C_t^{\mathrm{tool}}$, and finally uses $C_t^{\mathrm{scene}}$ as a query to match the historical feedback memory $\mathcal{H}_{t-1}$ and obtain scene-relevant feedback evidence $C_t^{\mathrm{fb}}$. Thus, $C_t$ represents the fine-grained contextual evidence used in the $t$-th reconstruction round, providing structured decision evidence for the dynamic router.

The scene-conditioned reconstruction evidence $C_t^{\mathrm{scene}}$ describes the reconstruction constraints implied by the current scene. It mainly includes alignment reliability, texture observability, and fusion risk. Alignment reliability reflects whether reliable cross-exposure correspondences can be established under the observed motion and occlusion conditions. Texture observability indicates whether useful textures and structures are preserved in saturated or under-exposed regions. Fusion risk estimates whether aggressive fusion may introduce ghosting, structural inconsistency, or local artifacts. These scene-side evidence entries provide the dynamic router with reconstruction-aware constraints for selecting a suitable strategy.

The tool applicability evidence $C_t^{\mathrm{tool}}$ provides scene-relevant tool knowledge for the dynamic router. The expert tool knowledge base $\mathcal{K}$ contains general descriptions of all candidate tools, including their functions, applicable conditions, and potential failure modes. Directly feeding the entire $\mathcal{K}$ to the router may introduce irrelevant tool information and increase the ambiguity of strategy selection. Therefore, FCM matches $\mathcal{K}$ with the current scene-conditioned evidence $C_t^{\mathrm{scene}}$ and extracts only the tool knowledge relevant to the current reconstruction constraints. Specifically, $C_t^{\mathrm{tool}}$ summarizes which types of tools are potentially suitable, under what conditions they are reliable, and under what conditions they may fail. In this way, $C_t^{\mathrm{tool}}$ filters and reformulates the global tool knowledge into compact tool-aware evidence, while leaving the final strategy selection to the dynamic router.

The feedback evidence $C_t^{\mathrm{fb}}$ is obtained by matching the current scene-conditioned evidence with the historical feedback memory. Specifically, the historical feedback memory before the $t$-th reconstruction round is denoted as
\begin{equation}
\mathcal{H}_{t-1}
=
\mathcal{H}_{0}
\cup
\{F_{\tau}\}_{\tau=1}^{t-1},
\label{eq:history}
\end{equation}                                
where $\mathcal{H}_{0}$ denotes the initial feedback memory constructed from the training dataset without updating model parameters, and $F_{\tau}$ denotes the structured feedback generated by the perception--distortion feedback evaluator at round $\tau$. Given the current scene evidence $C_t^{\mathrm{scene}}$, FCM uses GPT-5.3 to retrieve relevant feedback cases from $\mathcal{H}_{t-1}$. The retrieved feedback cases provide correction-oriented evidence for the dynamic router, indicating which failure patterns or artifact risks should be avoided in the current reconstruction round.

Finally, the dynamic router generates the execution strategy for the $t$-th round according to the contextual evidence and candidate tool set:
\begin{equation}
\mathcal{R}_t
=
\Phi_{\mathrm{router}}(C_t,\mathcal{T}),
\quad
\mathcal{R}_t \in \mathcal{T},
\label{eq:router}
\end{equation}
where $\Phi_{\mathrm{router}}$ denotes the dynamic routing process instantiated with Qwen3.6-Plus~\cite{qwen36plus}, $\mathcal{T}$ denotes the candidate tool set, and $\mathcal{R}_t$ denotes the image-level alignment and fusion strategy selected for the current reconstruction round. Through FCM, HDRAgent transforms scene perception results, expert tool knowledge, and historical feedback into structured decision evidence, allowing the dynamic router to adaptively select an overall reconstruction strategy for different dynamic scenes.

\subsection{Perception--Distortion Feedback Evaluation}

Global objective metrics can measure overall HDR reconstruction quality, but may overlook local visual defects in dynamic regions, such as residual ghosting, boundary misalignment, and fusion artifacts. To provide more informative feedback for iterative correction, HDRAgent introduces a perception--distortion feedback mechanism, which combines metric-based quality scores with MLLM-based perceptual diagnosis and converts them into structured feedback for context update and dynamic routing.

Given the reconstruction result $\hat{H}_t$ at the $t$-th round, on the training set we map both $\hat{H}_t$ and the ground-truth HDR image $H^{gt}$ into the SDR domain before metric computation:
\begin{equation}
Y_t = \mathcal{T}(\hat{H}_t),
\quad
Y^{gt} = \mathcal{T}(H^{gt}),
\label{eq:tonemapping_train}
\end{equation}
where $\mathcal{T}(\cdot)$ denotes the tone-mapping operation, $Y_t$ denotes the tone-mapped result of $\hat{H}_t$, and $Y^{gt}$ denotes the tone-mapped result of $H^{gt}$.

The distortion branch computes both full-reference and no-reference metrics on the tone-mapped results:
\begin{equation}
D_t^{\mathrm{train}}
=
\left\{
\Phi_{\mathrm{fr}}(Y_t,Y^{gt}),
\Phi_{\mathrm{nr}}(Y_t)
\right\},
\label{eq:train_distortion}
\end{equation}
where $\Phi_{\mathrm{fr}}$ denotes the full-reference evaluation process, and $\Phi_{\mathrm{nr}}$ denotes the no-reference evaluation process.

In parallel, the perceptual branch uses an MLLM to diagnose local visual artifacts from the tone-mapped result:
\begin{equation}
P_t
=
\Phi_{\mathrm{vlm}}(Y_t,\mathcal{P}_{\mathrm{eval}}),
\label{eq:perceptual_eval_train}
\end{equation}
where $\Phi_{\mathrm{vlm}}$ denotes the perceptual evaluation process instantiated with Qwen3-VL-Plus~\cite{qwen3vl}, and $\mathcal{P}_{\mathrm{eval}}$ is the evaluation instruction. $P_t$ records perceptual artifact descriptions, such as residual ghosting, boundary misalignment, fusion artifacts, and structural discontinuity.

The metric scores and perceptual diagnosis are then summarized into unified feedback:
\begin{equation}
F_t
=
\Phi_{\mathrm{eval}}(D_t^{\mathrm{train}},P_t),
\label{eq:train_feedback}
\end{equation}
where $\Phi_{\mathrm{eval}}$ denotes the feedback summarization process instantiated with Qwen3-VL-Plus~\cite{qwen3vl}, and $F_t$ denotes the structured feedback at round $t$. $F_t$ records not only quality scores, but also the type, severity, and possible cause of local reconstruction failures. On the training set, the generated feedback is accumulated to form the initial historical feedback memory $\mathcal{H}_{0}$, which stores typical failure patterns and correction-oriented experience without updating model parameters.

On the test set, ground-truth HDR images are unavailable, so full-reference metrics are excluded from the feedback loop. Since the feedback is used to identify reconstruction failures and guide subsequent strategy selection rather than provide pixel-level supervision, no-reference metrics from $Y_t$ together with MLLM-based perceptual diagnosis provide sufficient cues for context matching and dynamic routing.

\subsection{Agent-Guided Generative Alignment}

\begin{figure*}[ht!]
\centering
\includegraphics[width=\textwidth]{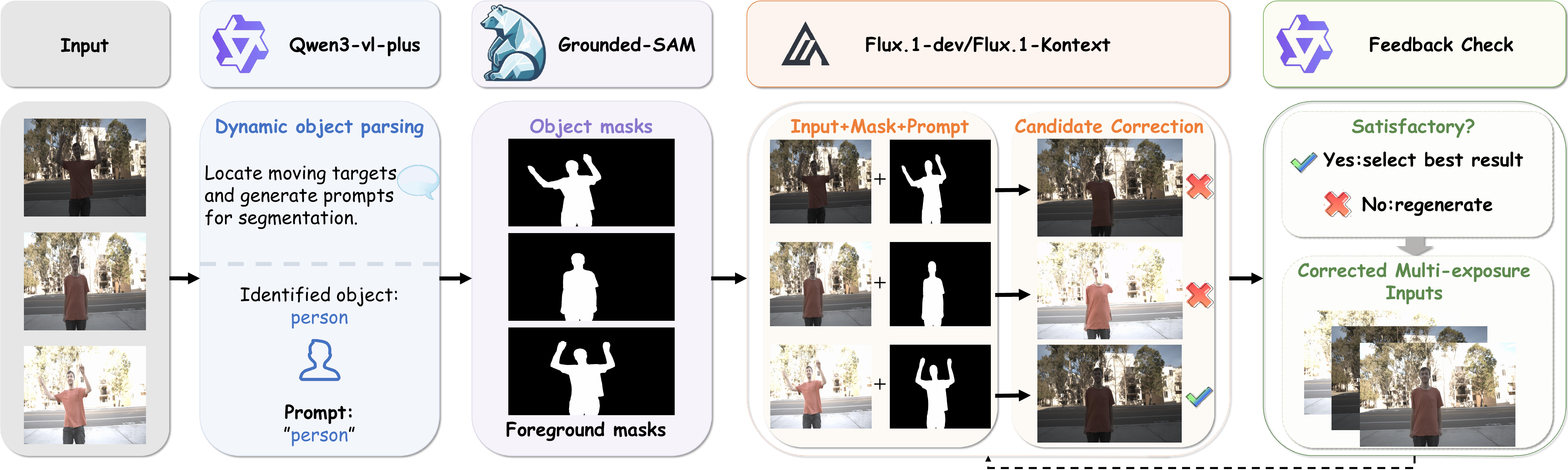}
\caption{Illustration of the proposed agent-guided generative alignment module. Qwen3-VL-Plus first parses the dynamic foreground target and generates a text prompt. Grounded-SAM then produces foreground masks for the reference and non-reference frames. The reference frame is kept unchanged and used only as a condition, while FLUX.1 Fill/Kontext performs mask-guided generation for non-reference frames. A feedback checker further verifies whether the generated result satisfies the reference-consistency requirement.}
\label{fig:AGGA}
\vspace{-0.4cm}
\end{figure*}

To handle extreme motion and large occlusion where existing alignment methods are unreliable, we design an agent-guided generative alignment (AGGA) module, which leverages generative priors to locally reconstruct unreliable dynamic regions in non-reference frames. As shown in Fig.~\ref{fig:AGGA}, AGGA follows a perception--segmentation--generation--feedback pipeline. Specifically, Qwen3-VL-Plus~\cite{qwen3vl} first parses the moving foreground object and generates a textual prompt for segmentation. Grounded-SAM~\cite{liu2024groundingdino,kirillov2023segment} then localizes the prompted object and produces foreground masks. Based on the generated masks and prompt, FLUX.1 Fill/Kontext~\cite{blackforestlabs2024fluxfill,blackforestlabs2025fluxkontext} performs reference-conditioned masked generation to correct unreliable regions in non-reference frames. Finally, Qwen3-VL-Plus is further used as a feedback checker to verify the structural consistency of the generated result.

Let $I_r$ denote the reference frame and $I_i$ denote a non-reference frame. Given the instruction $\mathcal{P}_{\mathrm{agga}}$, Qwen3-VL-Plus~\cite{qwen3vl} analyzes the multi-exposure inputs and produces a dynamic-object prompt:
\begin{equation}
q = \Phi_{\mathrm{vlm}}(I_r,I_i,\mathcal{P}_{\mathrm{agga}}),
\label{eq:agga_prompt}
\end{equation}
where $\Phi_{\mathrm{vlm}}$ denotes the MLLM-based target parsing process, and $q$ denotes the textual prompt describing the moving foreground object, e.g., ``person''. This prompt provides semantic guidance for subsequent open-vocabulary segmentation.

Based on the prompt $q$, Grounded-SAM~\cite{liu2024groundingdino,kirillov2023segment} segments the corresponding foreground object in both the reference and non-reference frames:
\begin{equation}
M_r = \Phi_{\mathrm{gsam}}(I_r,q),
\quad
M_i = \Phi_{\mathrm{gsam}}(I_i,q),
\label{eq:agga_mask}
\end{equation}
where $\Phi_{\mathrm{gsam}}$ denotes the Grounded-SAM segmentation process, and $M_r,M_i \in \{0,1\}^{H\times W}$ denote the foreground masks of the reference and non-reference frames, respectively. To cover the potentially misaligned dynamic region, we construct a correction mask by merging the two masks:
\begin{equation}
M_i^{u} = M_r \cup M_i.
\label{eq:agga_union_mask}
\end{equation}
The union mask indicates the region in the non-reference frame that may require generative correction.

AGGA then performs reference-conditioned masked generation on the non-reference frame. Specifically, FLUX.1 Fill and FLUX.1 Kontext~\cite{blackforestlabs2024fluxfill,blackforestlabs2025fluxkontext} take the non-reference frame $I_i$, the reference condition $I_r$, the correction mask $M_i^{u}$, and the inpainting prompt $q$ as inputs:
\begin{equation}
G_i = \Phi_{\mathrm{flux}}(I_i,I_r,M_i^{u},q),
\label{eq:agga_flux}
\end{equation}
where $\Phi_{\mathrm{flux}}$ denotes the reference-conditioned masked generation process, and $G_i$ denotes the generated non-reference candidate. The reference frame provides structural guidance, while the mask restricts the editable area to the unreliable foreground region. To preserve reliable image content, only the generated masked region is used for correction:
\begin{equation}
\tilde{I}_i
=
M_i^{u}\odot G_i + (1-M_i^{u})\odot I_i,
\label{eq:agga_composition}
\end{equation}
where $\tilde{I}_i$ denotes the corrected non-reference frame and $\odot$ denotes element-wise multiplication.

Finally, HDRAgent uses Qwen3-VL-Plus as a feedback checker to verify whether the corrected non-reference frame satisfies the reference-consistency requirement:
\begin{equation}
v_i = \Phi_{\mathrm{check}}(\tilde{I}_i,I_r,M_i^{u},\mathcal{P}_{\mathrm{check}}),
\label{eq:agga_check}
\end{equation}
where $\Phi_{\mathrm{check}}$ denotes the MLLM-based consistency checking process, $\mathcal{P}_{\mathrm{check}}$ denotes the checking instruction, and $v_i\in\{0,1\}$ indicates whether the correction is accepted. If the result is unsatisfactory, the feedback is used to refine the prompt and re-invoke the generative correction process.

Repeating AGGA for all non-reference frames yields the corrected multi-exposure sequence:
\begin{equation}
\tilde{\mathcal{I}}
=
\{I_r\}
\cup
\{\tilde{I}_i\}_{i\neq r},
\label{eq:agga_sequence}
\end{equation}
which is subsequently fed into the HDR fusion module. In this way, AGGA converts difficult geometric alignment in large-motion regions into reference-conditioned masked generation, reducing ghosting and local fusion artifacts while preserving reliable background content.

\begin{figure*}[t]
\centering

\setlength{\tabcolsep}{1pt}
\renewcommand{\arraystretch}{0.0}

\newcommand{\scenegap}{2pt}

\newcommand{\methodrowgap}{6pt}

\newcommand{\namegap}{-4pt}

\newcommand{\methodtwoscenes}[3]{%
\begin{minipage}[t]{0.142\textwidth}
    \centering
    \includegraphics[width=\linewidth]{#1}\\[\scenegap]
    \includegraphics[width=\linewidth]{#2}\\[\namegap]
    \makebox[\linewidth][c]{\scriptsize\textbf{#3}}%
\end{minipage}
}

\resizebox{\textwidth}{!}{%
\begin{tabular}{@{}ccccccc@{}}

\methodtwoscenes
{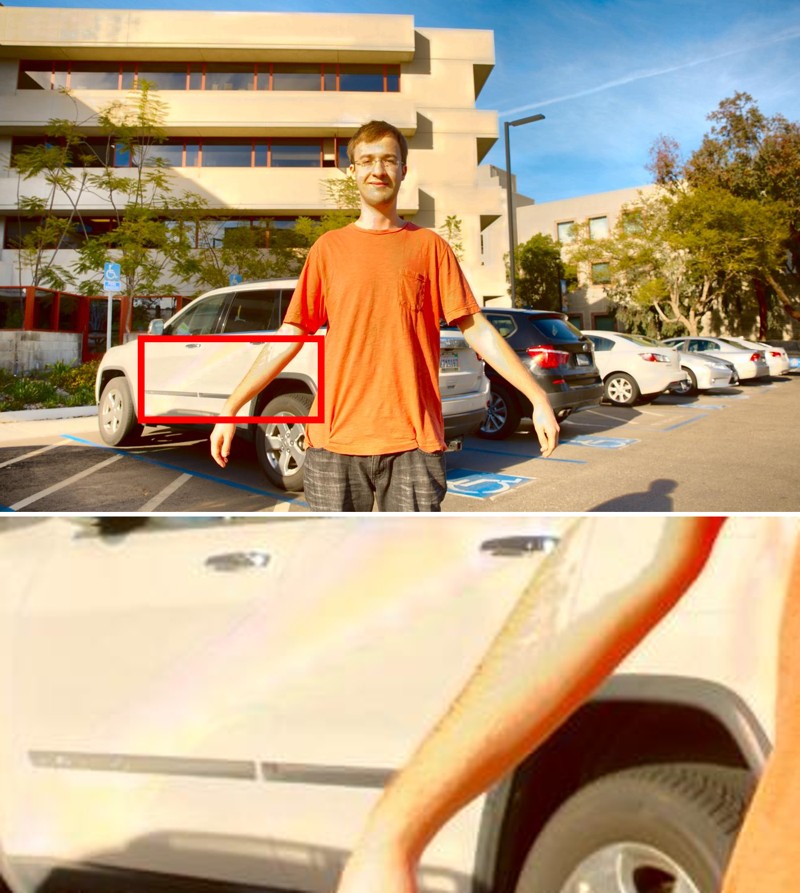}
{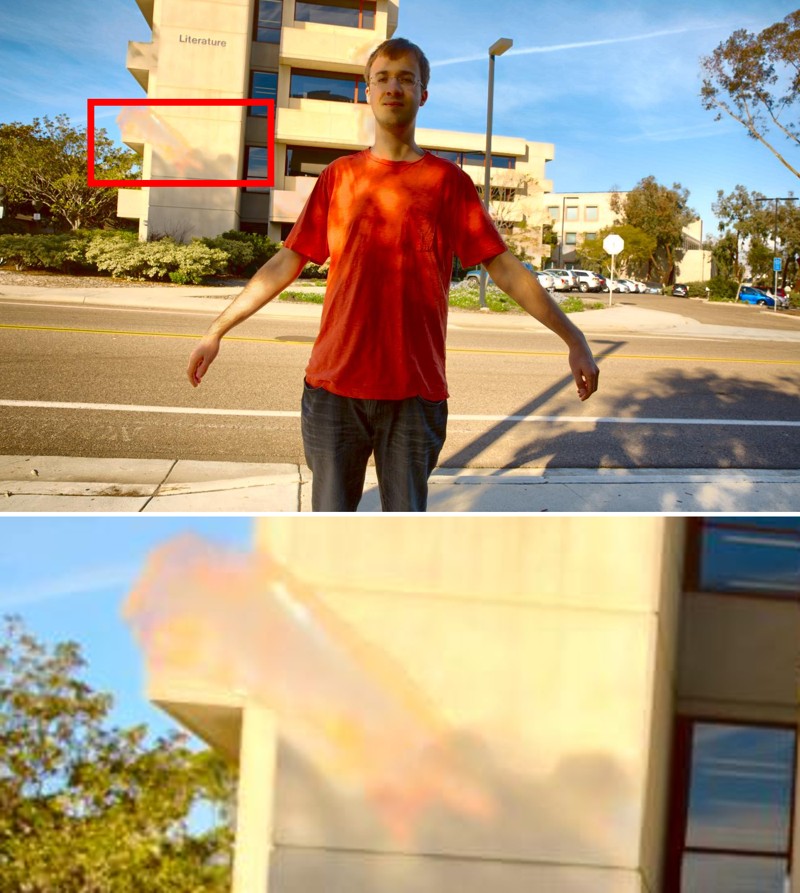}
{(a) APNet~\cite{chen2022attention}}
&
\methodtwoscenes
{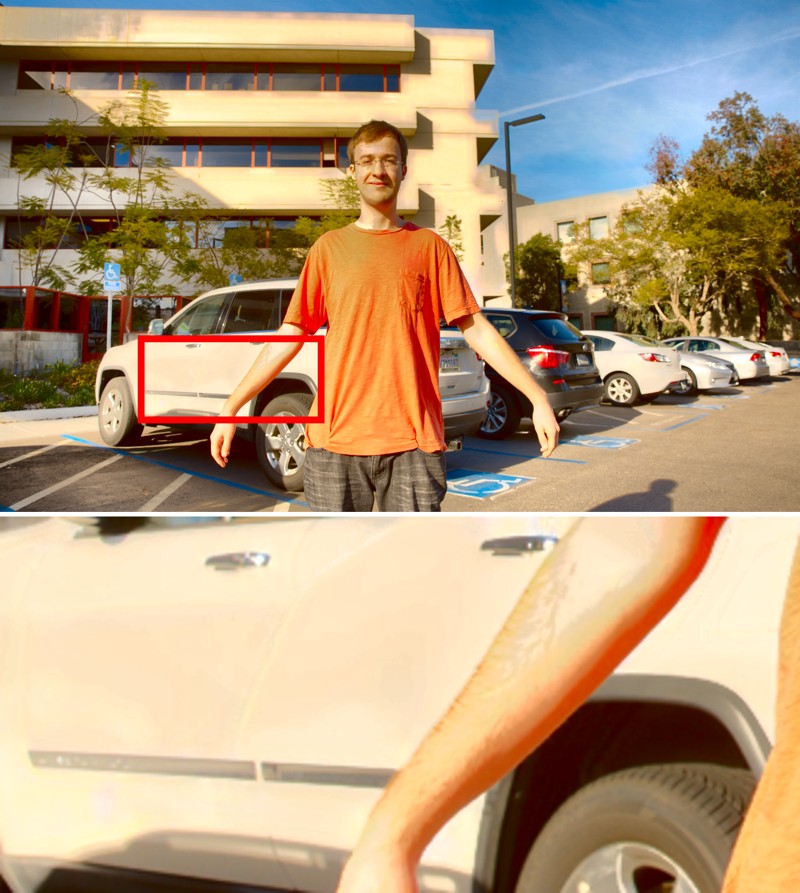}
{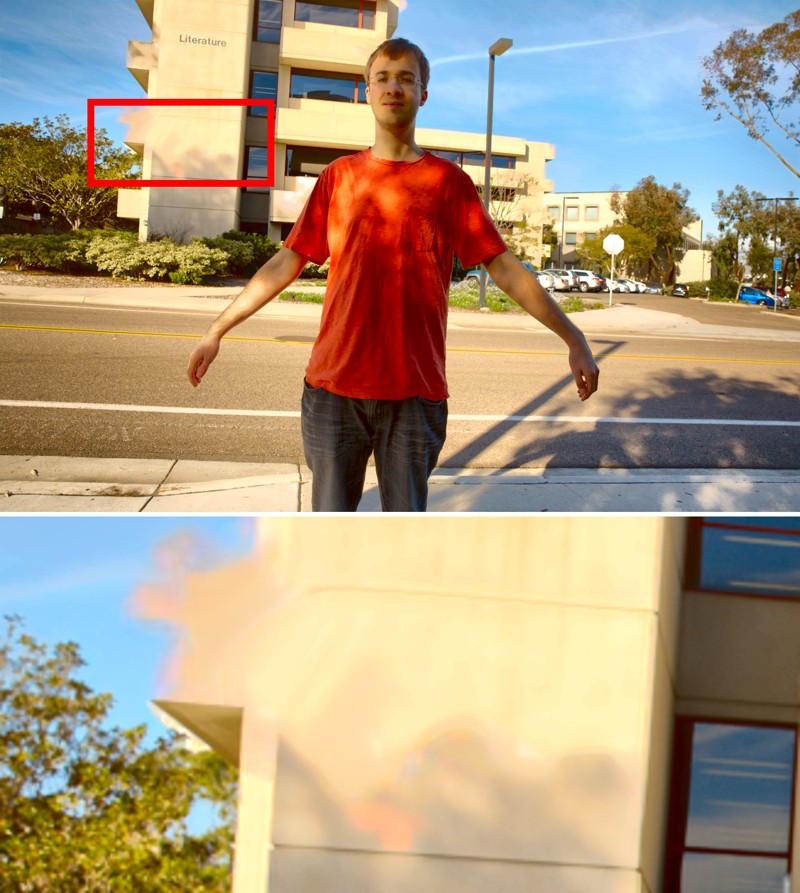}
{(b) AHDR~\cite{yan2019attention}}
&
\methodtwoscenes
{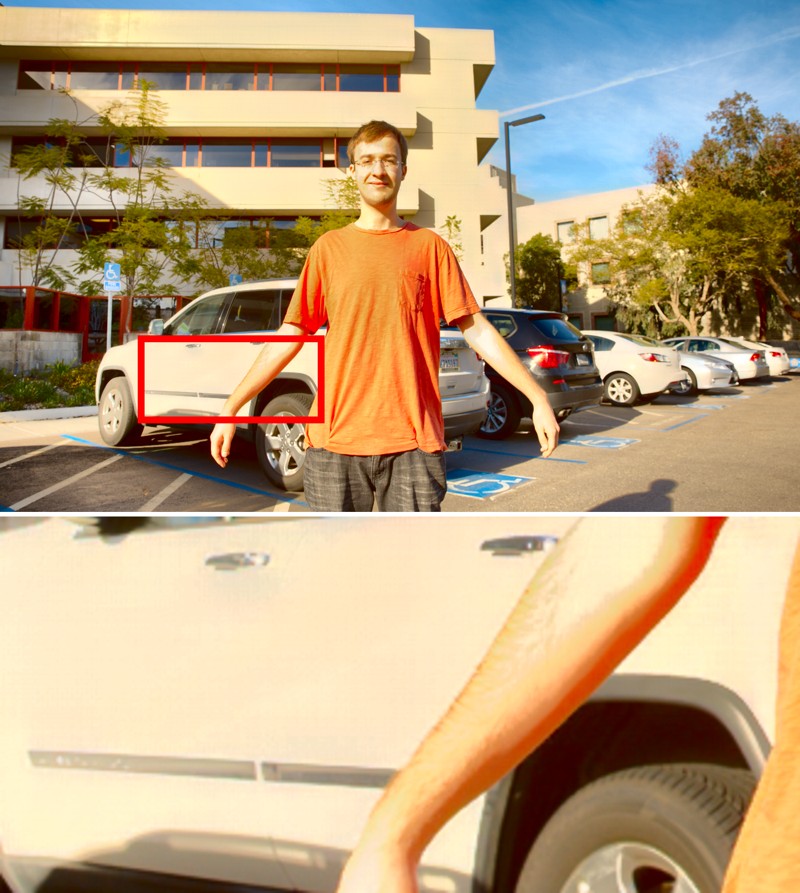}
{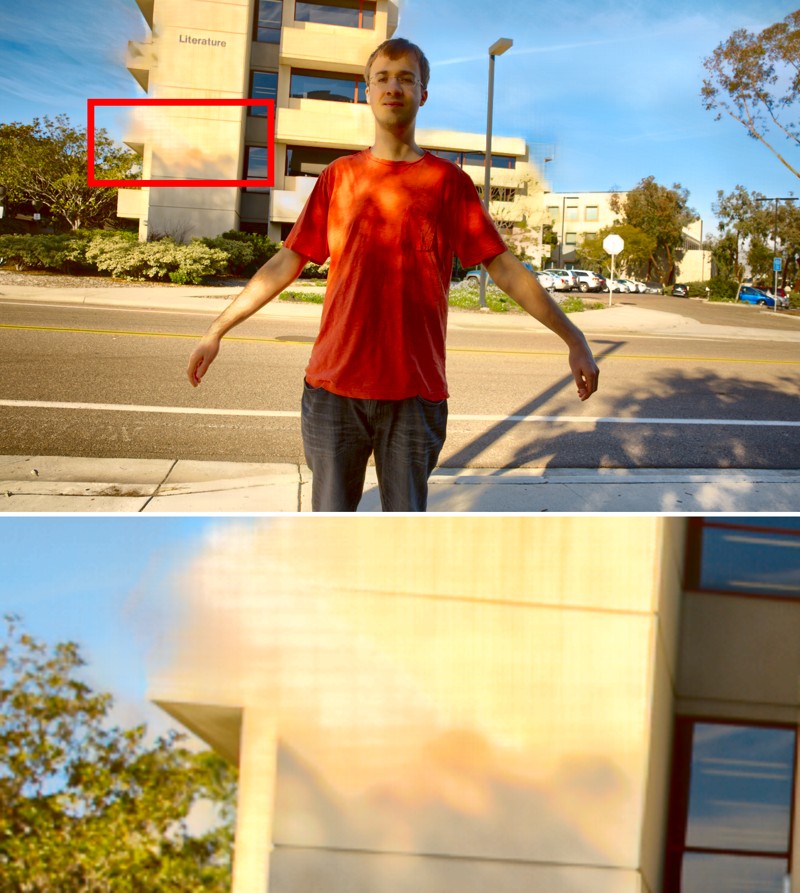}
{(c) NHDRR~\cite{yan2020deep}}
&
\methodtwoscenes
{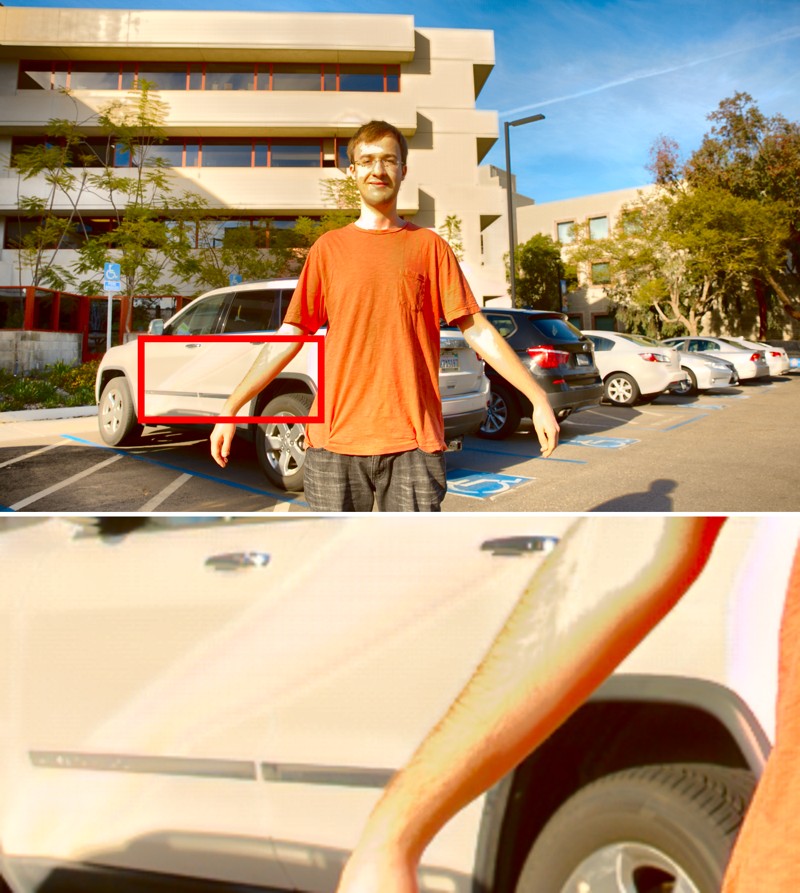}
{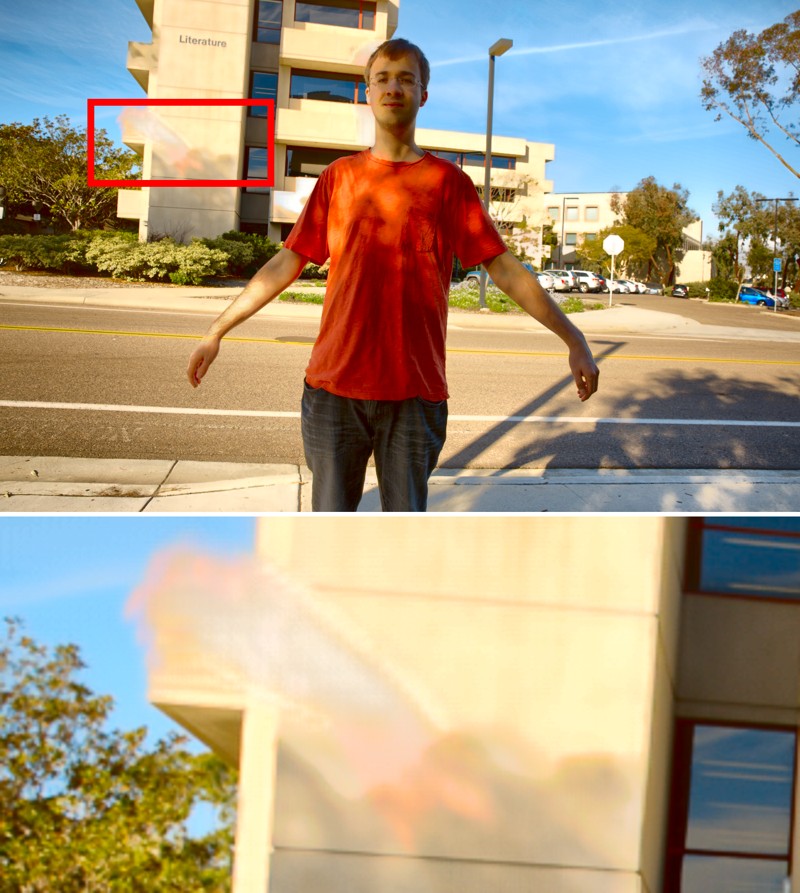}
{(d) DHDR~\cite{wu2018deep}}
&
\methodtwoscenes
{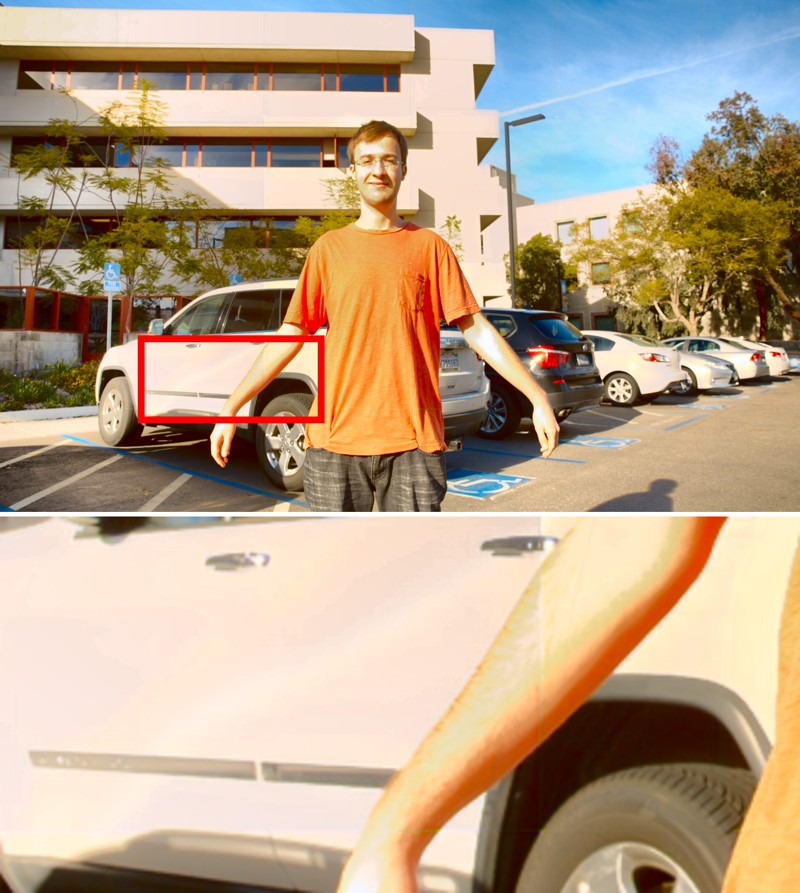}
{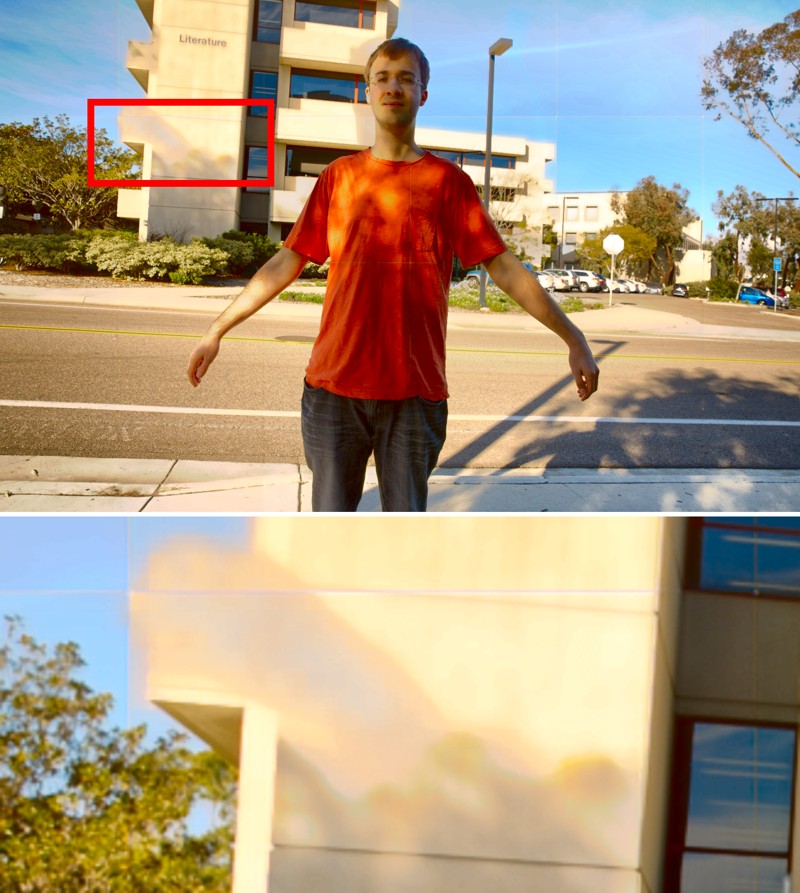}
{(e) SCTNet~\cite{tel2023alignment}}
&
\methodtwoscenes
{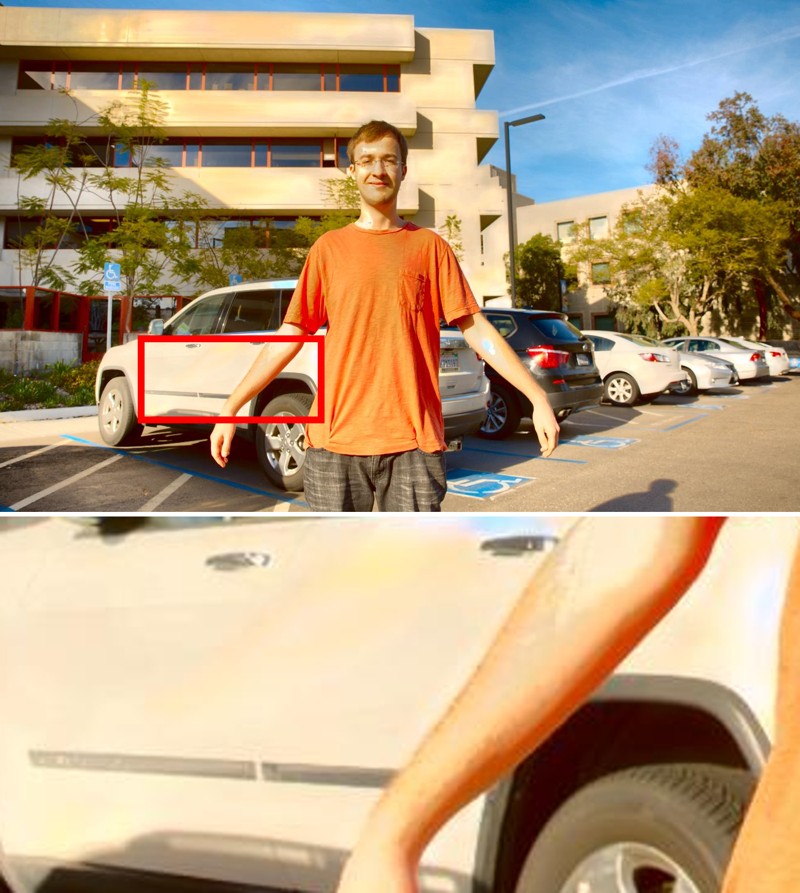}
{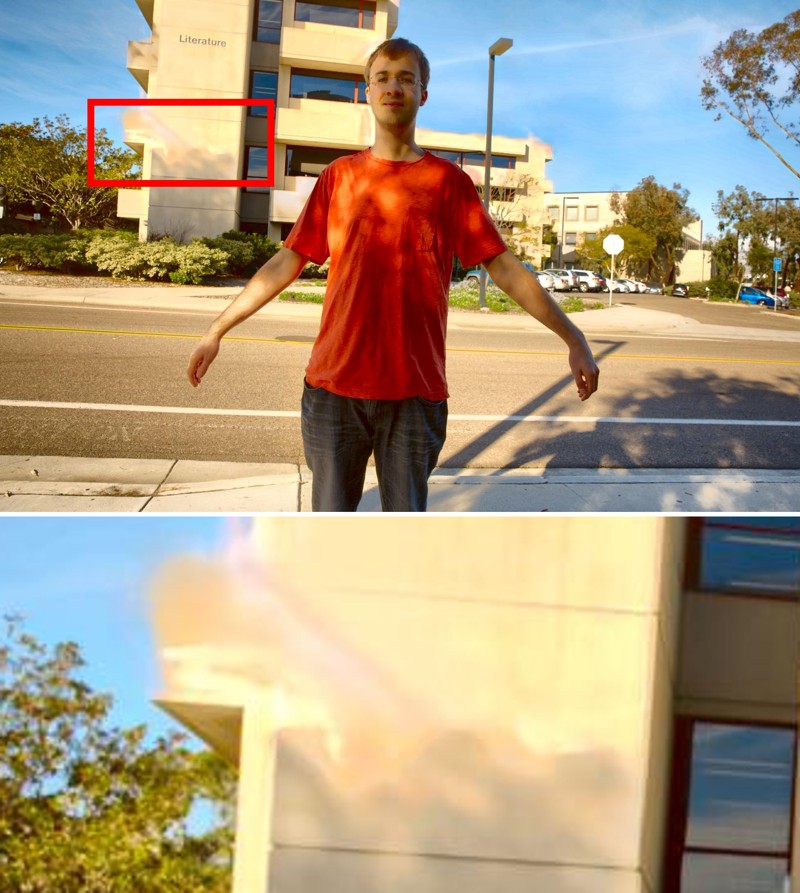}
{(f) HDR-GAN~\cite{niu2021hdr}}
&
\methodtwoscenes
{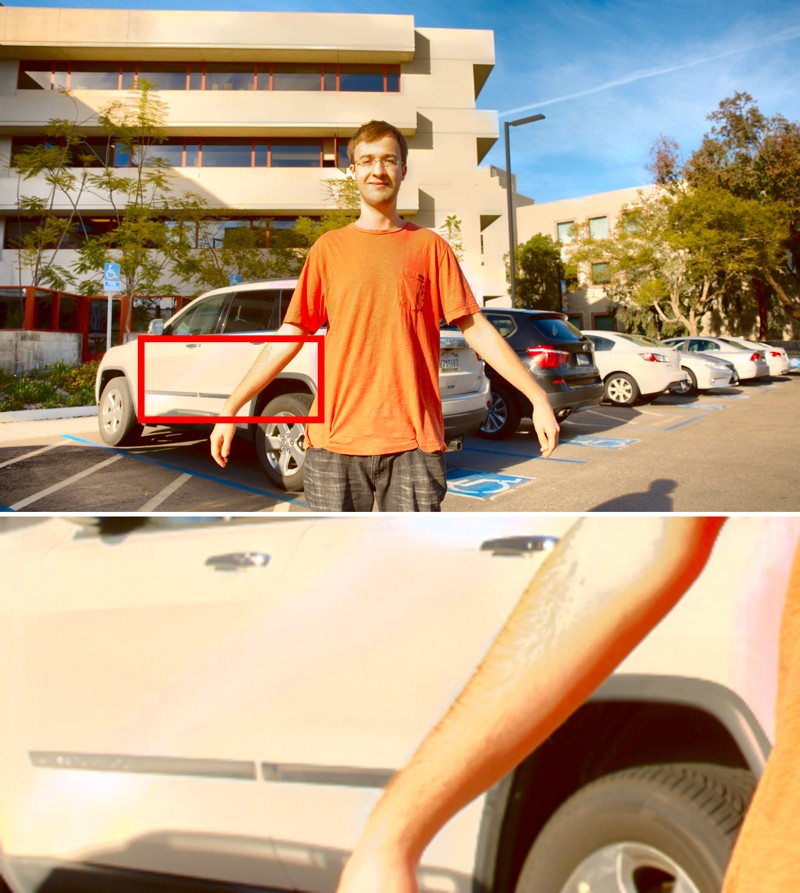}
{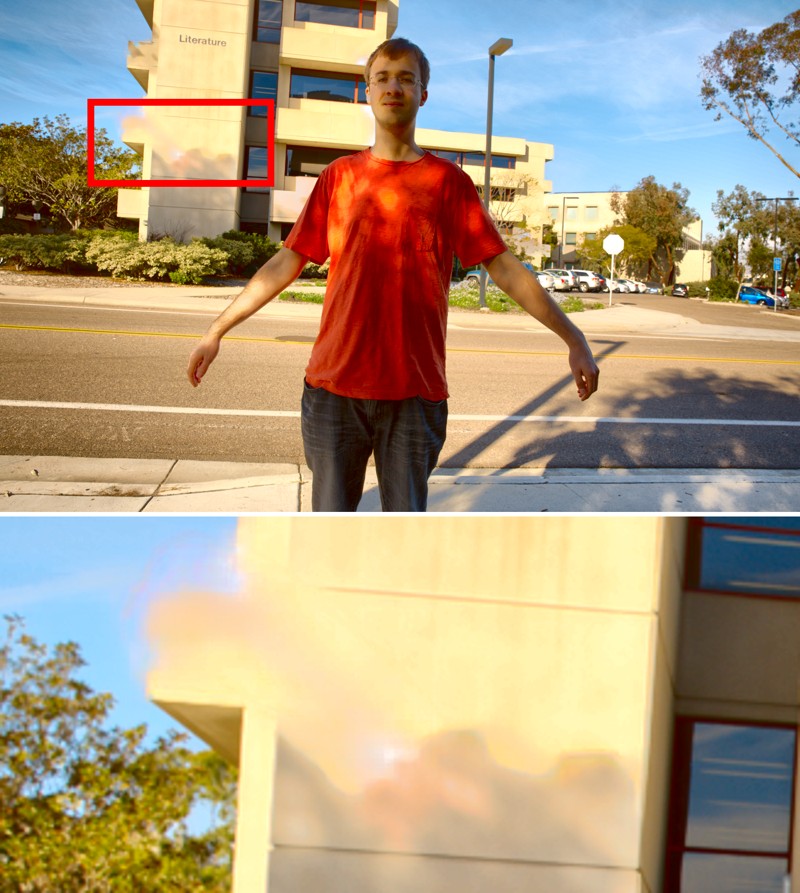}
{(g) HDRTrans~\cite{liu2022ghost}}
\\[\methodrowgap]

\methodtwoscenes
{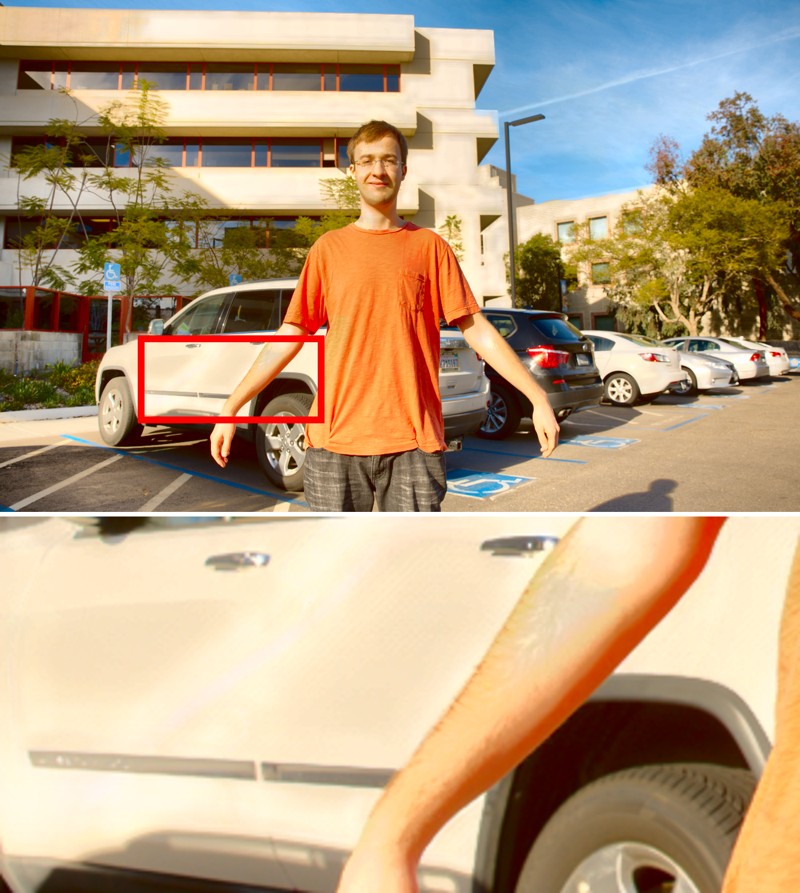}
{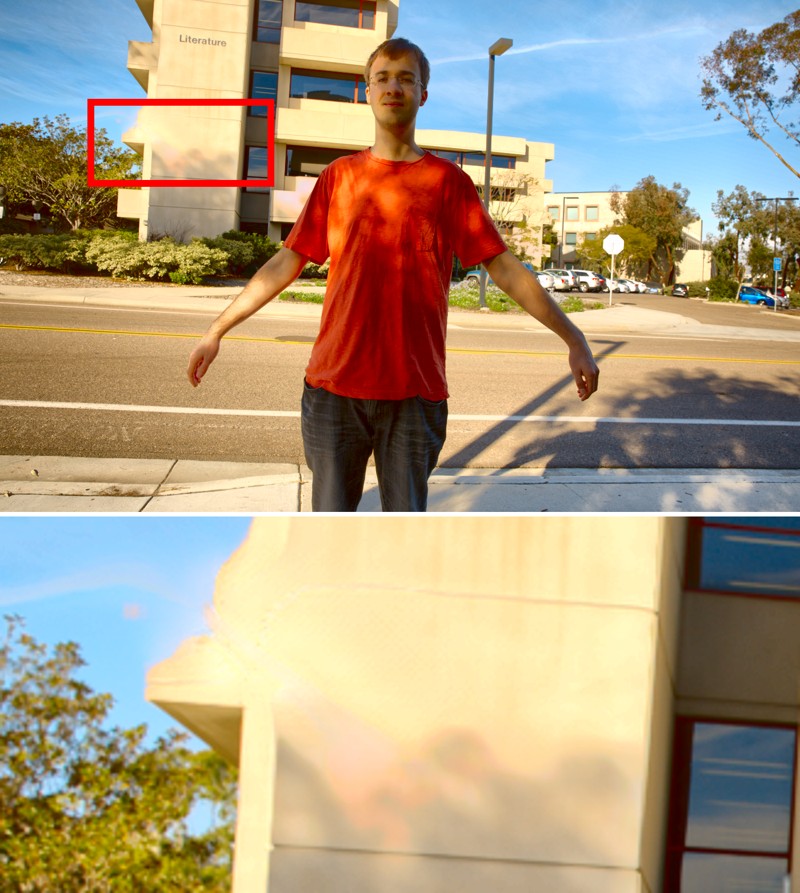}
{(h) SAFNet~\cite{kong2024safnet}}
&
\methodtwoscenes
{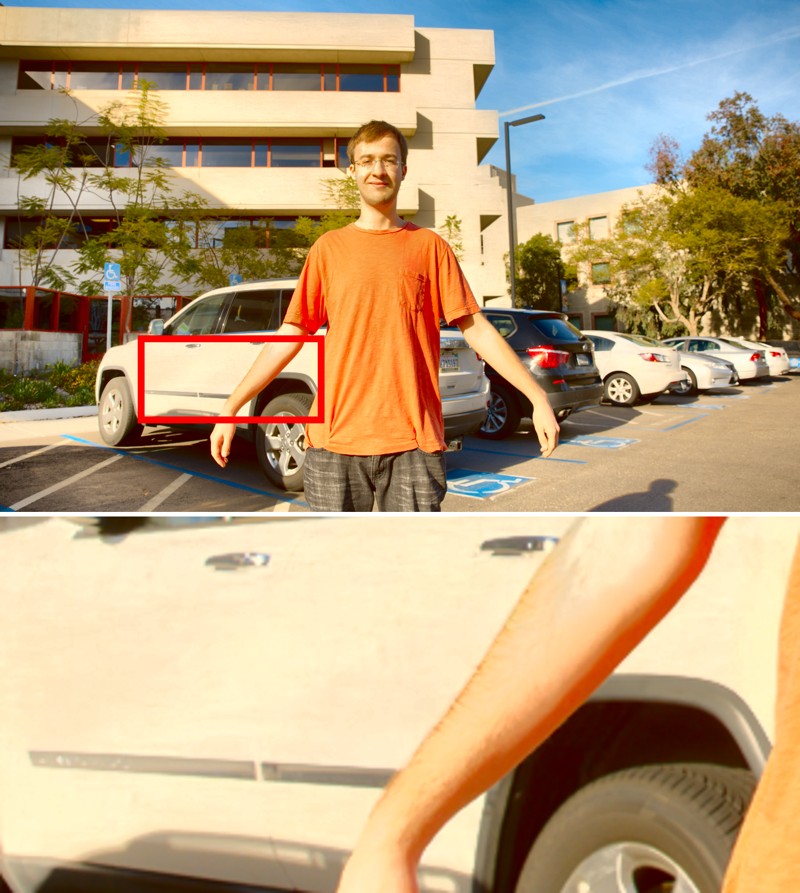}
{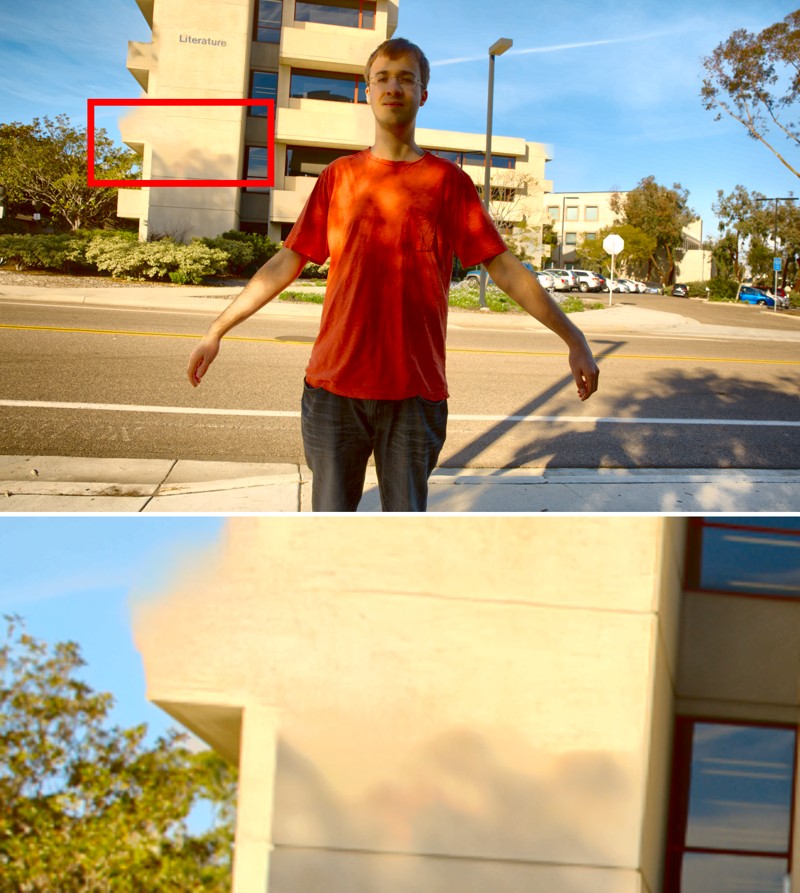}
{(i) DiffHDR~\cite{yan2023toward}}
&
\methodtwoscenes
{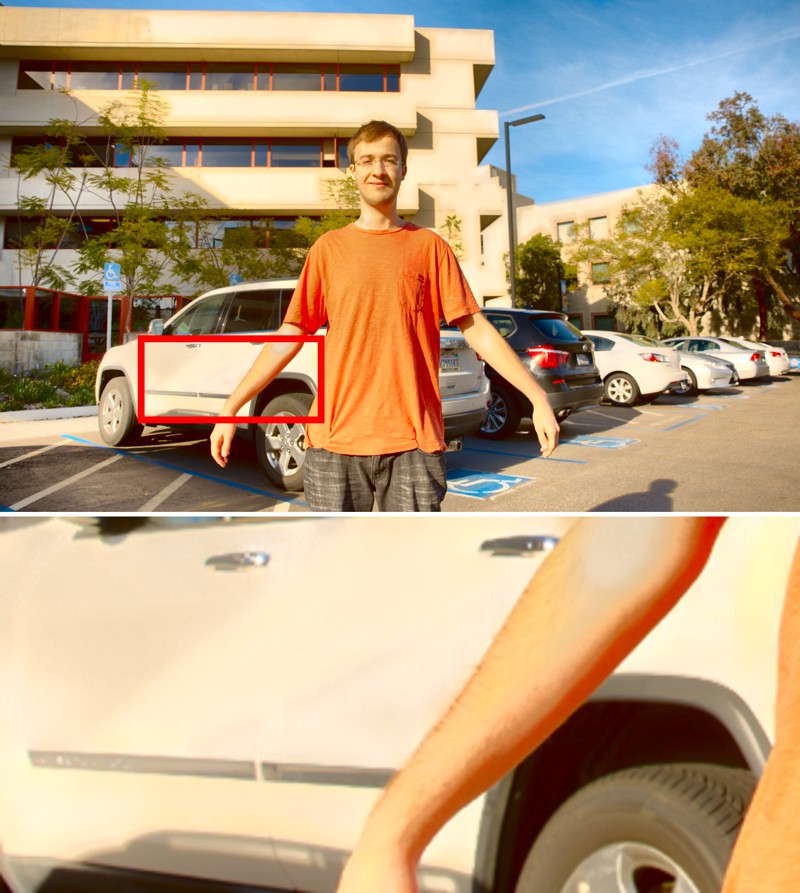}
{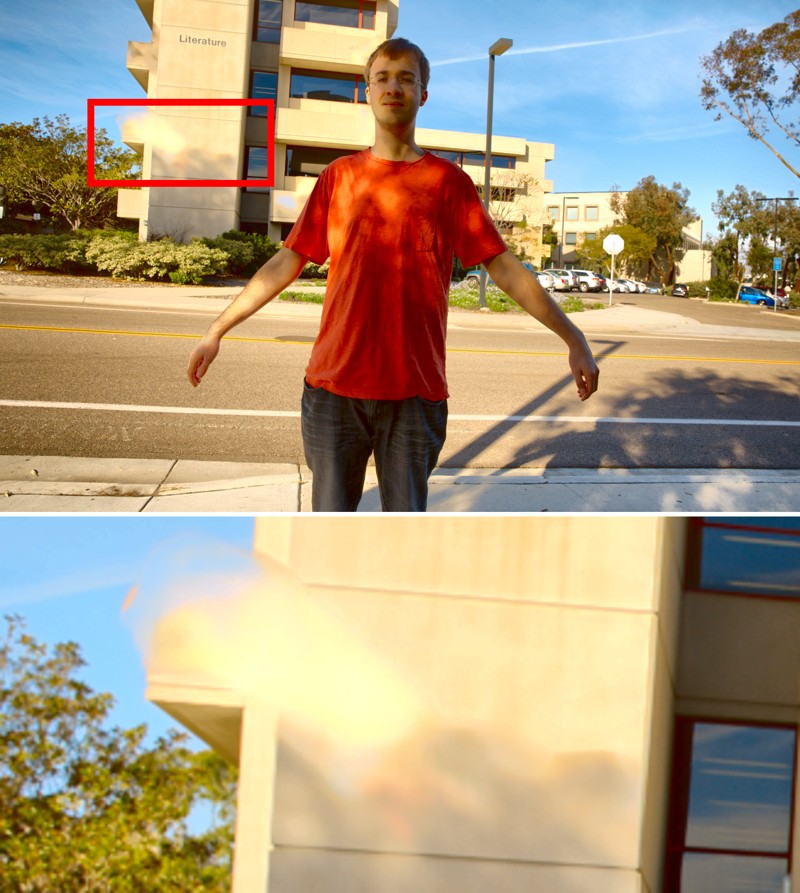}
{(j) LFDiff~\cite{hu2024generating}}
&
\methodtwoscenes
{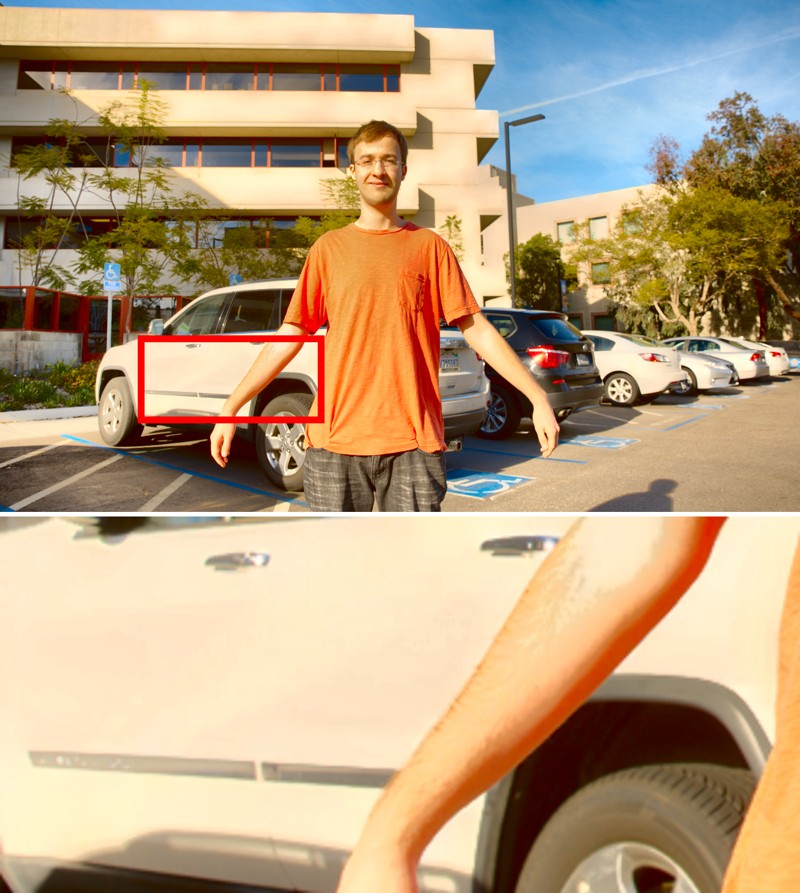}
{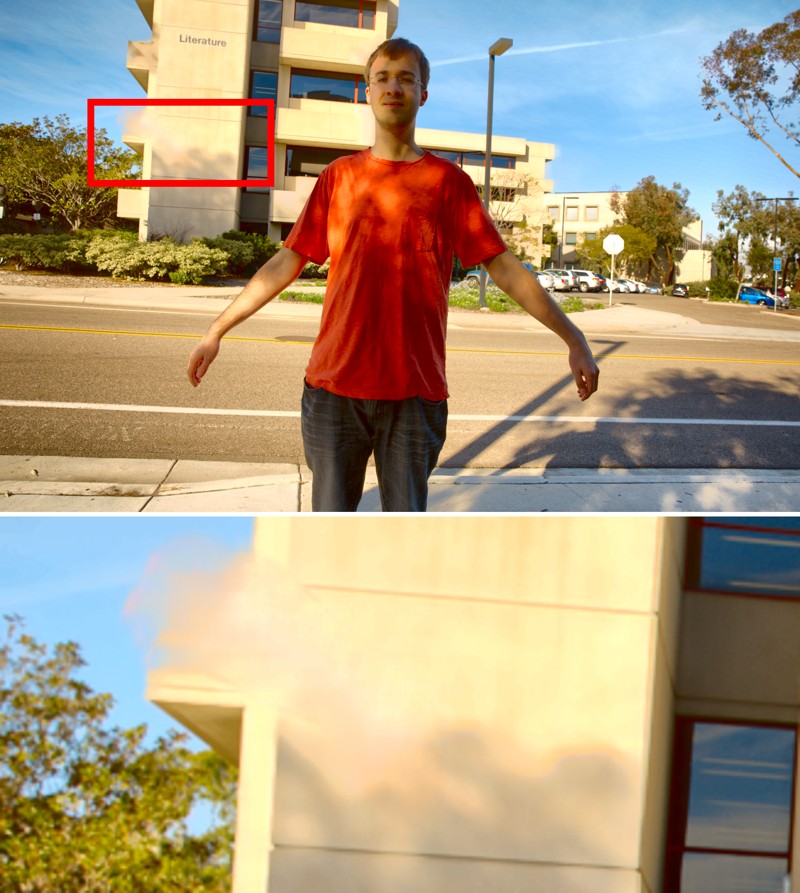}
{(k) AFUNet~\cite{li2025afunet}}
&
\methodtwoscenes
{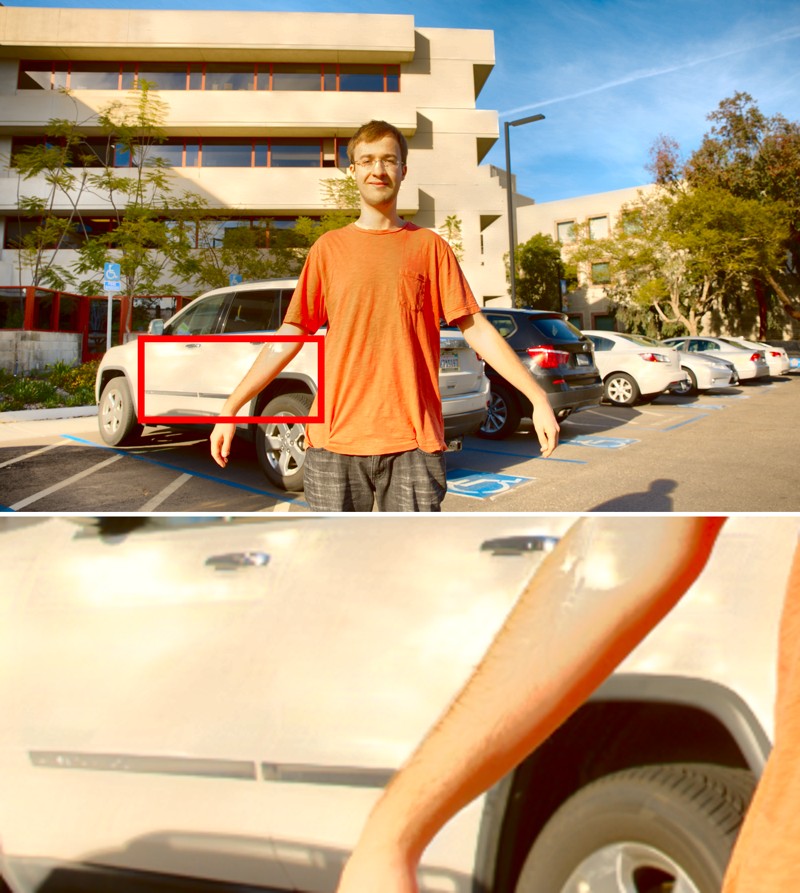}
{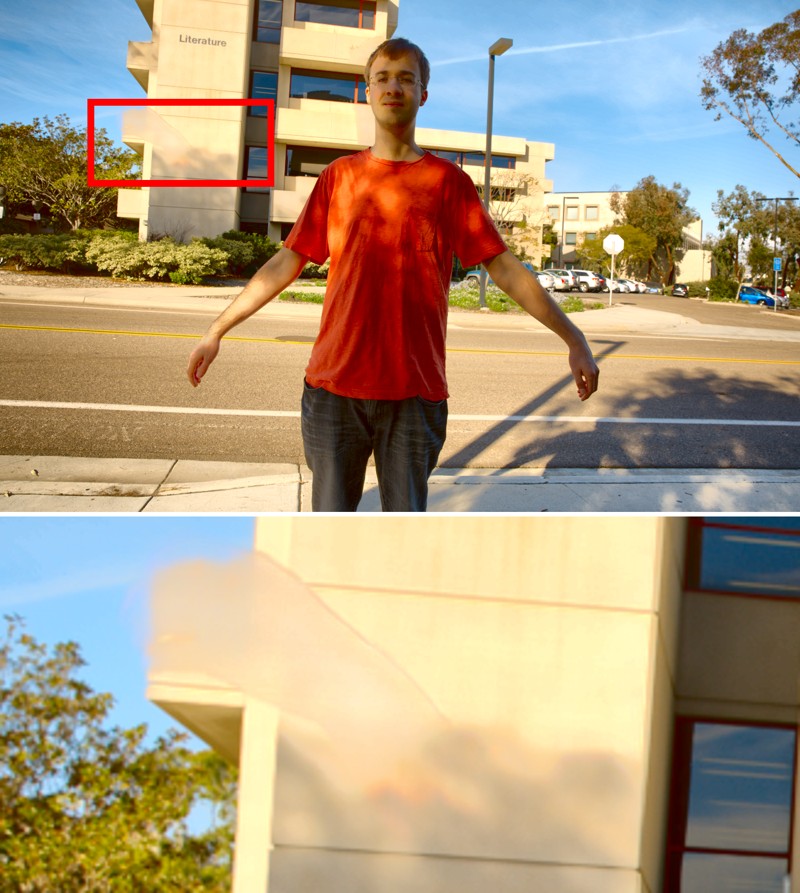}
{(l) DDPF-PR~\cite{zhou2026high}}
&
\methodtwoscenes
{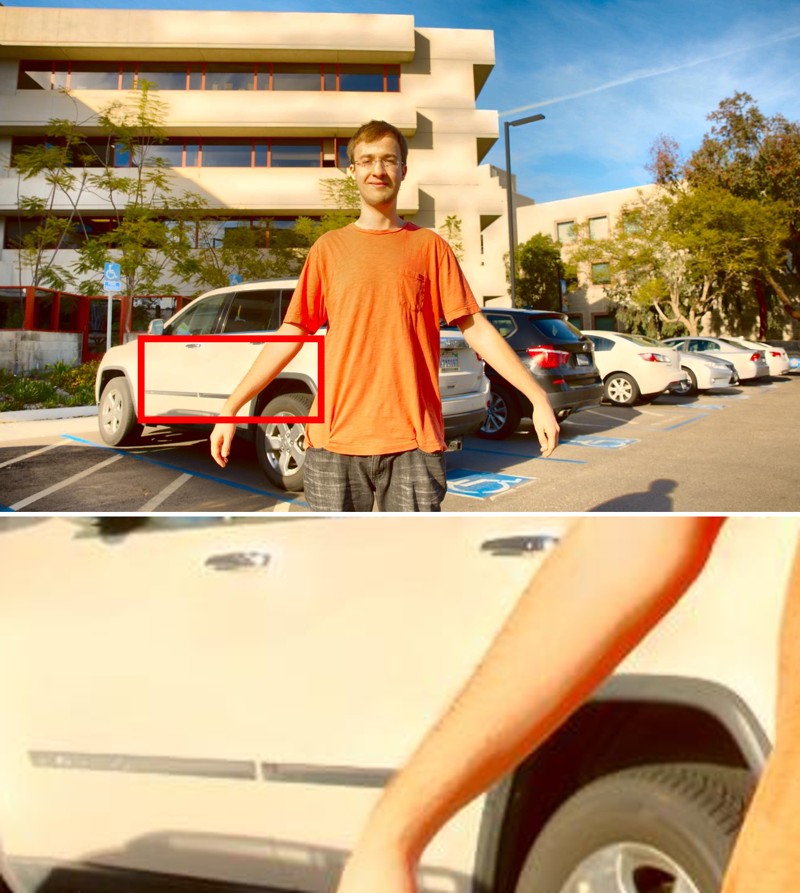}
{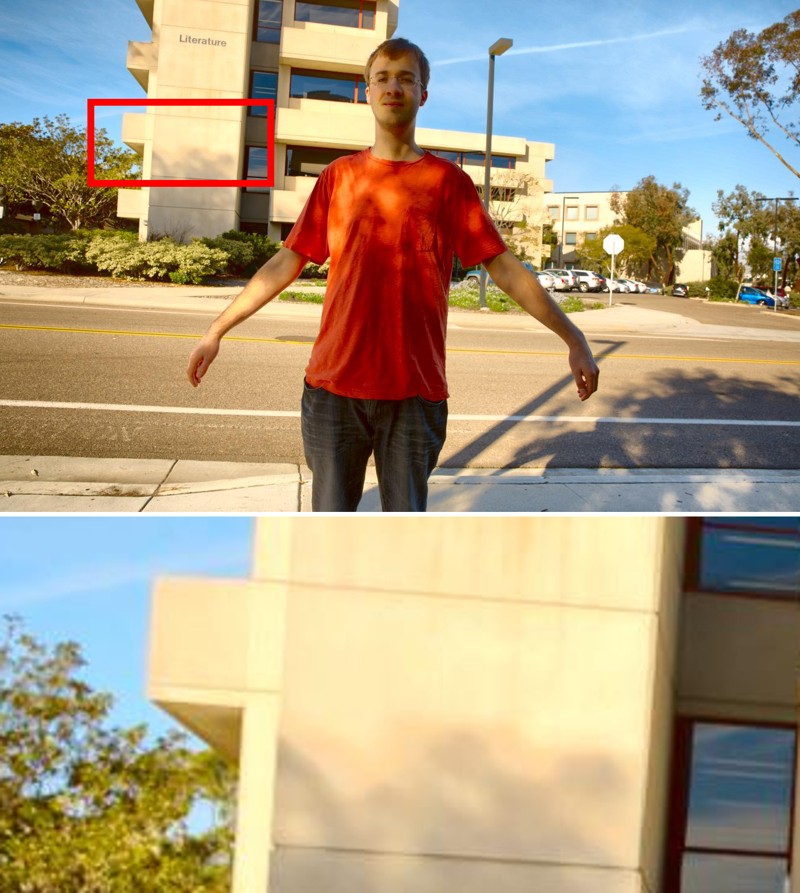}
{(m) HDRAgent}
&
\methodtwoscenes
{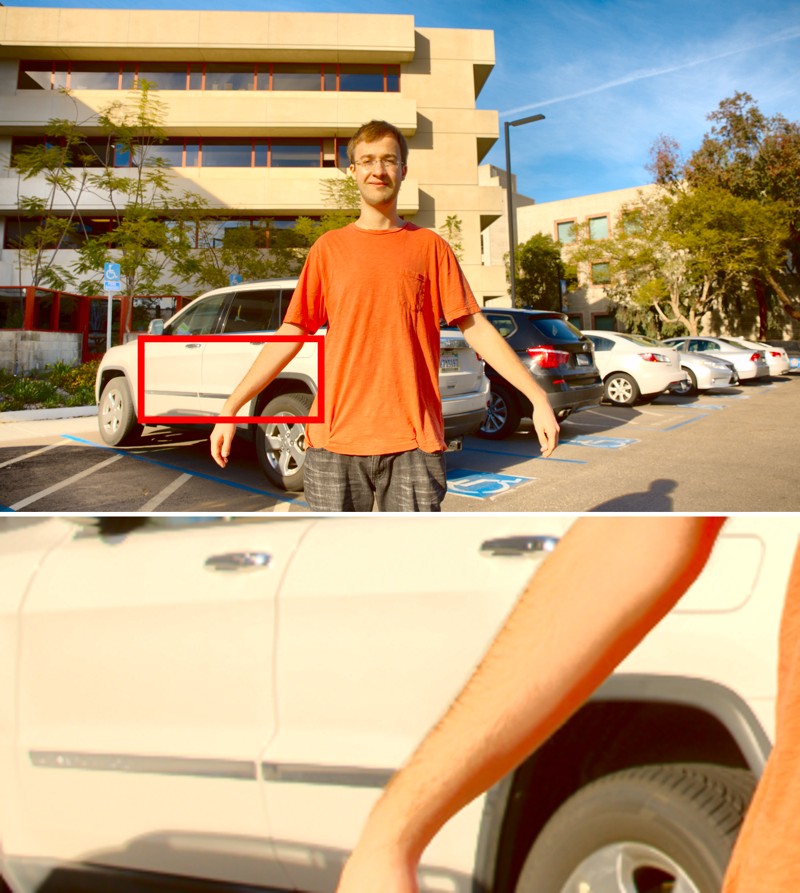}
{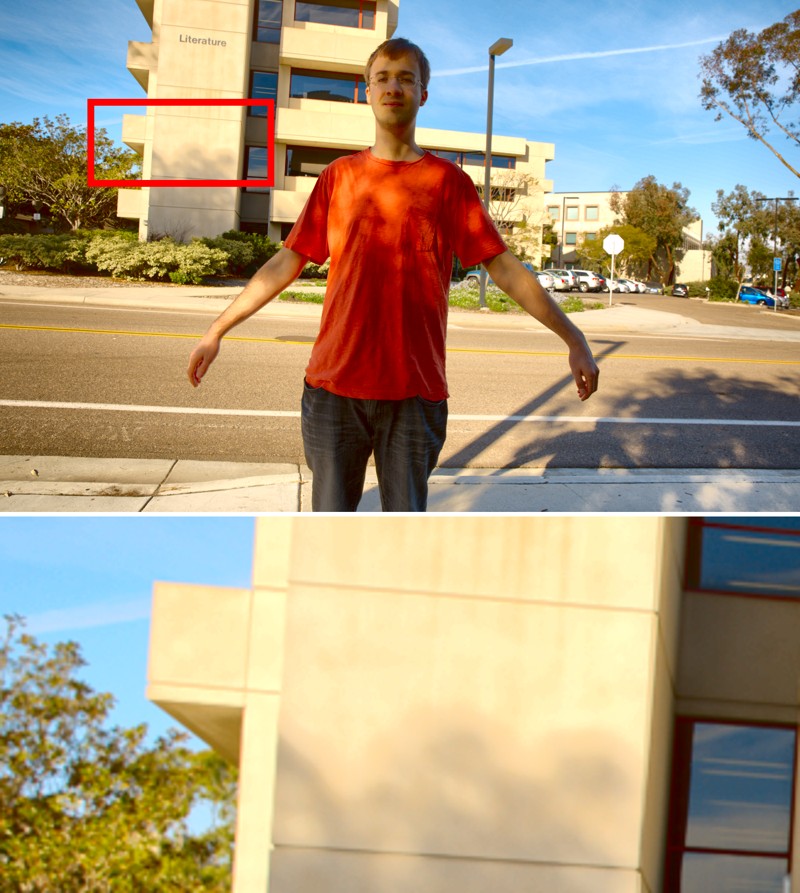}
{(n) GT}

\end{tabular}
}

\caption{Visual comparison of different methods on dynamic scenes from the Kalantari's dataset~\cite{kalantari2017deep}.
Our HDRAgent yields cleaner local structures and fewer ghosting artifacts, with results closer to the reference HDR images.
}
\label{fig:visual_comparison_sig}
\vspace{-0.4cm}
\end{figure*}

\section{Experiments}
\subsection{Experimental Settings}

\noindent\textbf{Implementation Details.}
HDRAgent uses Qwen3-VL-Plus~\cite{qwen3vl} for visual scene perception, dynamic foreground parsing, artifact assessment, and feedback checking, Qwen3.6-Plus~\cite{qwen36plus} as the dynamic router, and GPT-5.3~\cite{openai_gpt53} for fine-grained contextual knowledge matching. In AGGA, Grounded-SAM~\cite{liu2024groundingdino,kirillov2023segment} generates foreground masks from MLLM-produced prompts, while FLUX.1 Fill/Kontext~\cite{blackforestlabs2024fluxfill,blackforestlabs2025fluxkontext} performs reference-conditioned masked generation. The feedback accumulated on the training set is stored as the initial historical feedback memory $\mathcal{H}_{0}$ without model-parameter updates. Experiments are conducted on a single NVIDIA RTX 4090 GPU, and part of the FLUX.1 model states are offloaded to CPU memory during generative alignment.

\noindent\textbf{Datasets.}
We evaluate HDRAgent on three representative multi-exposure HDR benchmarks: Kalantari's dataset~\cite{kalantari2017deep}, Tel's dataset~\cite{tel2023alignment}, and Hu's dataset~\cite{hu2013hdr}. These datasets cover a wide range of exposure differences, motion patterns, and scene dynamics. Kalantari's dataset provides real dynamic HDR sequences captured under natural lighting, including 74 training samples and 15 testing samples. Tel's dataset further increases the diversity of motion and illumination conditions, with 108 training examples and 36 testing examples. In addition, Hu's dataset is a synthetic benchmark generated with a physics-based game engine, where each scene is rendered under three exposure settings to simulate realistic sensor responses. 

\noindent\textbf{Evaluation Metrics.}
We evaluate the reconstruction quality from both fidelity-oriented and perceptual perspectives. Following common multi-exposure HDR evaluation protocols, we report five full-reference metrics: PSNR-$\mu$, PSNR-$L$, SSIM-$\mu$, SSIM-$L$, and HDR-VDP-2~\cite{kalantari2017deep,wang2004image,luo2023refusion}. PSNR-$L$ and SSIM-$L$ are computed in the linear HDR domain, while PSNR-$\mu$ and SSIM-$\mu$ are measured after $\mu$-law tone mapping. HDR-VDP-2 further evaluates perceptual visibility differences in HDR images. These metrics jointly assess the numerical fidelity, structural consistency, and perceptual quality of reconstructed HDR results under full-reference evaluation.

To further assess visual quality without relying on ground-truth HDR images, we adopt six no-reference image quality metrics, including MUSIQ~\cite{ke2021musiq}, MANIQA~\cite{yang2022maniqa}, LIQE~\cite{zhang2023liqe}, HyperIQA~\cite{su2020hyperiqa}, DBCNN~\cite{zhang2020dbcnn}, and NIMA~\cite{talebi2018nima}. Unless otherwise specified, all no-reference metrics are computed on tone-mapped images. For PSNR-$\mu$, PSNR-$L$, SSIM-$\mu$, SSIM-$L$, HDR-VDP-2, MUSIQ, MANIQA, LIQE, HyperIQA, DBCNN, and NIMA, higher values indicate better reconstruction quality.

\noindent\textbf{Compared Methods.}
To evaluate the effectiveness of HDRAgent, we compare it with representative and recent multi-exposure HDR reconstruction methods, covering CNN-based methods, Transformer-based methods, and generative or diffusion-based methods. Specifically, we include classical baselines such as DHDR~\cite{wu2018deep}, AHDR~\cite{yan2019attention}, NHDRR~\cite{yan2020deep}, HDR-GAN~\cite{niu2021hdr}, APNT~\cite{chen2022attention}, CA-ViT~\cite{liu2022ghost}, HyHDR~\cite{yan2023unified}, SCTNet~\cite{tel2023alignment}, and DiffHDR~\cite{yan2023toward}, as well as recent methods from the past two years, including LFDiff~\cite{hu2024generating}, SAFNet~\cite{kong2024safnet}, AFUNet~\cite{li2025afunet} and DDPF-PR~\cite{zhou2026high}. For a fair comparison, we follow the official implementations or reported settings whenever available, and evaluate all methods using the same dataset splits and metric protocols.

\subsection{Quantitative Assessment}

Tables~\ref{tab:quantity_kalantari}, \ref{tab:quantity_tel}, and \ref{tab:quantity_hu} report full-reference quantitative comparisons on Kalantari's dataset, Tel's dataset, and Hu's dataset, respectively. Across the three benchmarks, HDRAgent consistently achieves the best or second-best results on most fidelity-oriented metrics, including PSNR-$\mu$, PSNR-$L$, SSIM-$\mu$, SSIM-$L$, and HDR-VDP-2. These results demonstrate that the proposed agent-driven reconstruction framework can effectively improve numerical fidelity and structural consistency under diverse dynamic HDR imaging conditions.

In addition to full-reference evaluation, Table~\ref{tab:nr_kalantari} reports no-reference quality comparisons on Kalantari's dataset. HDRAgent achieves the best performance on DBCNN and NIMA, and ranks among the top three on MUSIQ, MANIQA, LIQE, and HyperIQA. This indicates that HDRAgent not only improves distortion-oriented reconstruction accuracy, but also maintains competitive perceptual quality. Overall, the full-reference and no-reference evaluations jointly verify the robustness of HDRAgent in reducing ghosting artifacts, preserving local structures, and producing visually reliable HDR results.


On Kalantari's dataset, as shown in Table~\ref{tab:quantity_kalantari}, HDRAgent achieves the best PSNR-$\mu$, SSIM-$L$, and HDR-VDP-2 scores, reaching 45.01, 0.9911, and 66.81, respectively. 
It also obtains the second-best PSNR-$L$ and SSIM-$\mu$ results, only slightly lower than SAFNet and AFUNet, respectively. 
These results show that HDRAgent provides a strong balance between tone-mapped fidelity, linear-domain reconstruction accuracy, structural consistency, and perceptual quality.

\begin{table}[ht]
  \centering
  \caption{Quantitative comparisons on Kalantari's dataset~\cite{kalantari2017deep}. The best and second-best results are highlighted in red and blue, respectively.}
  \label{tab:quantity_kalantari}
  \setlength{\tabcolsep}{3.2pt}
  \renewcommand{\arraystretch}{1.05}
  \resizebox{\columnwidth}{!}{
    \begin{tabular}{c c|ccccc}
      \toprule\toprule
      Methods & Venue & PSNR-$\mu$ & PSNR-$L$ & SSIM-$\mu$ & SSIM-$L$ & HDR-VDP-2 \\
      \hline
      DHDR~\cite{wu2018deep}         & ECCV'18  & 41.64 & 40.91 & 0.9869 & 0.9858 & 60.50 \\
      AHDR~\cite{yan2019attention}   & CVPR'19  & 43.62 & 41.03 & 0.9900 & 0.9862 & 62.30 \\
      NHDRR~\cite{yan2020deep}       & TIP'20   & 42.41 & 41.08 & 0.9887 & 0.9861 & 61.21 \\
      HDR-GAN~\cite{niu2021hdr}      & TIP'21   & 43.92 & 41.57 & 0.9905 & 0.9865 & 65.45 \\
      APNT~\cite{chen2022attention}  & TIP'22   & 43.94 & 41.61 & 0.9898 & 0.9879 & 64.05 \\
      CA-ViT~\cite{liu2022ghost}     & ECCV'22  & 44.32 & 42.18 & 0.9916 & 0.9884 & 66.03 \\
      HyHDR~\cite{yan2023unified}    & CVPR'23  & 44.64 & 42.47 & 0.9915 & 0.9894 & 66.05 \\
      SCTNet~\cite{tel2023alignment} & ICCV'23  & 44.43 & 42.21 & 0.9918 & 0.9891 & 66.64 \\
      DiffHDR~\cite{yan2023toward}   & TCSVT'23 & 44.11 & 41.73 & 0.9911 & 0.9885 & 65.52 \\
      SAFNet~\cite{kong2024safnet}   & ECCV'24  & 44.66 & \textcolor{red}{43.18} & 0.9919 & 0.9901 & 66.11 \\
      LFDiff~\cite{hu2024generating} & CVPR'24  & 44.76 & 42.59 & 0.9919 & \textcolor{blue}{0.9906} & 66.54 \\
      AFUNet~\cite{li2025afunet}     & ICCV'25  & \textcolor{blue}{44.91} & 42.59 & \textcolor{red}{0.9923} & \textcolor{blue}{0.9906} & \textcolor{blue}{66.75} \\
      Ours                           & -        & \textcolor{red}{45.01} & \textcolor{blue}{42.68} & \textcolor{blue}{0.9922} & \textcolor{red}{0.9911} & \textcolor{red}{66.81} \\
      \bottomrule\bottomrule
    \end{tabular}
  }
  \vspace{-0.4cm}
\end{table}


Table~\ref{tab:quantity_tel} presents the results on Tel's dataset, which contains more challenging motion and illumination variations. HDRAgent achieves the best results across all five full-reference metrics, with PSNR-$\mu$, PSNR-$L$, SSIM-$\mu$, SSIM-$L$, and HDR-VDP-2 scores of 43.61, 47.99, 0.9877, 0.9962, and 71.11, respectively. Compared with AFUNet, the second-best method on this dataset, HDRAgent consistently improves all metrics, indicating stronger tone-mapped fidelity, linear-domain reconstruction accuracy, structural consistency, and perceptual quality under complex dynamic conditions.

\begin{table}[ht]
  \centering
  \caption{Quantitative comparisons on Tel's dataset~\cite{tel2023alignment}. The best and second-best results are highlighted in red and blue, respectively.}
  \label{tab:quantity_tel}
  \setlength{\tabcolsep}{3.2pt}
  \renewcommand{\arraystretch}{1.05}
  \resizebox{\columnwidth}{!}{
    \begin{tabular}{c c|ccccc}
      \toprule\toprule
      Methods & Venue & PSNR-$\mu$ & PSNR-$L$ & SSIM-$\mu$ & SSIM-$L$ & HDR-VDP-2 \\
      \hline
      DHDR~\cite{wu2018deep}         & ECCV'18  & 40.05 & 43.37 & 0.9794 & 0.9924 & 67.09 \\
      AHDR~\cite{yan2019attention}   & CVPR'19  & 42.08 & 45.30 & 0.9837 & 0.9943 & 68.80 \\
      NHDRR~\cite{yan2020deep}       & TIP'20   & 36.68 & 39.61 & 0.9590 & 0.9853 & 65.41 \\
      HDR-GAN~\cite{niu2021hdr}      & TIP'21   & 41.71 & 44.87 & 0.9832 & 0.9949 & 69.57 \\
      CA-ViT~\cite{liu2022ghost}     & ECCV'22  & 42.39 & 46.35 & 0.9844 & 0.9948 & 69.23 \\
      SCTNet~\cite{tel2023alignment} & ICCV'23  & 42.55 & 47.51 & 0.9850 & 0.9952 & 70.66 \\
      DiffHDR~\cite{yan2023toward}   & TCSVT'23 & 42.18 & 45.63 & 0.9841 & 0.9946 & 69.88 \\
      SAFNet~\cite{kong2024safnet}   & ECCV'24  & 42.68 & 47.46 & 0.9792 & 0.9955 & 68.16 \\
      AFUNet~\cite{li2025afunet}     & ICCV'25  & \textcolor{blue}{43.31} & \textcolor{blue}{47.83} & \textcolor{blue}{0.9876} & \textcolor{blue}{0.9959} & \textcolor{blue}{71.08} \\
      HDRAgent                       & -        & \textcolor{red}{43.61} & \textcolor{red}{47.99} & \textcolor{red}{0.9877} & \textcolor{red}{0.9962} & \textcolor{red}{71.11} \\
      \bottomrule\bottomrule
    \end{tabular}
  }
  \vspace{-0.2cm}
\end{table}


On Hu's synthetic dataset, Table~\ref{tab:quantity_hu} shows that HDRAgent achieves the best results on four out of five metrics, including PSNR-$\mu$, PSNR-$L$, SSIM-$\mu$, and HDR-VDP-2. 
Specifically, HDRAgent reaches 49.01 in PSNR-$\mu$, 52.45 in PSNR-$L$, 0.9970 in SSIM-$\mu$, and 77.49 in HDR-VDP-2, outperforming recent competing methods on tone-mapped fidelity, linear-domain reconstruction accuracy, structural consistency, and perceptual quality. 
For SSIM-$L$, LFDiff obtains the best value of 0.9993, while HDRAgent achieves the second-best result of 0.9992. 
These results indicate that HDRAgent remains robust on synthetic scenes with strong illumination variations, saturated regions, and fine local textures.

\begin{table}[ht]
  \centering
  \caption{Quantitative comparisons on Hu's dataset~\cite{hu2020sensor}. The best and second-best results are highlighted in red and blue, respectively.}
  \label{tab:quantity_hu}
  \setlength{\tabcolsep}{3.2pt}
  \renewcommand{\arraystretch}{1.05}
  \resizebox{\columnwidth}{!}{
    \begin{tabular}{c c|ccccc}
      \toprule\toprule
      Methods & Venue & PSNR-$\mu$ & PSNR-$L$ & SSIM-$\mu$ & SSIM-$L$ & HDR-VDP-2 \\
      \hline
      DHDR~\cite{wu2018deep}         & ECCV'18  & 41.13 & 41.20 & 0.9870 & 0.9941 & 70.82 \\
      AHDR~\cite{yan2019attention}   & CVPR'19  & 45.76 & 49.22 & 0.9956 & 0.9980 & 75.04 \\
      NHDRR~\cite{yan2020deep}       & TIP'20   & 45.15 & 48.75 & 0.9956 & 0.9981 & 74.86 \\
      HDR-GAN~\cite{niu2021hdr}      & TIP'21   & 45.86 & 49.14 & 0.9945 & 0.9989 & 75.19 \\
      APNT~\cite{chen2022attention}  & TIP'22   & 46.41 & 47.97 & 0.9953 & 0.9986 & 73.06 \\
      CA-ViT~\cite{liu2022ghost}     & ECCV'22  & 48.10 & 51.17 & 0.9947 & 0.9989 & 77.12 \\
      HyHDR~\cite{yan2023unified}    & CVPR'23  & 48.46 & 51.91 & 0.9959 & 0.9991 & 77.24 \\
      DiffHDR~\cite{yan2023toward}   & TCSVT'23 & 48.03 & 50.23 & 0.9954 & 0.9989 & 76.22 \\
      SCTNet~\cite{tel2023alignment} & ICCV'23  & 48.10 & 51.03 & 0.9963 & 0.9991 & 77.14 \\
      SAFNet~\cite{kong2024safnet}   & ECCV'24  & 47.18 & 49.35 & 0.9951 & 0.9990 & 76.83 \\
      LFDiff~\cite{hu2024generating} & CVPR'24  & 48.74 & 52.10 & \textcolor{blue}{0.9968} & \textcolor{red}{0.9993} & 77.35 \\
      AFUNet~\cite{li2025afunet}     & ICCV'25  & \textcolor{blue}{48.83} & \textcolor{blue}{52.13} & \textcolor{blue}{0.9968} & 0.9991 & \textcolor{blue}{77.44} \\
      Ours                           & -        & \textcolor{red}{49.01} & \textcolor{red}{52.45} & \textcolor{red}{0.9970} & \textcolor{blue}{0.9992} & \textcolor{red}{77.49} \\
      \bottomrule\bottomrule
    \end{tabular}
  }
  \vspace{-0.2cm}
\end{table}

These quantitative results show that HDRAgent consistently performs competitively across real and synthetic HDR benchmarks. The improvements are particularly evident in PSNR-$\mu$ and HDR-VDP-2, suggesting that HDRAgent can better reconstruct visually faithful tone-mapped HDR results and suppress perceptually noticeable artifacts. Together with the qualitative comparisons, these results validate the effectiveness of the proposed agent-driven dynamic deghosting framework.

\begin{table}[t]
  \centering
  \caption{No-reference quality comparisons on Kalantari's dataset. The best, second-best, and third-best results are highlighted in red, blue, and green, respectively.}
  \label{tab:nr_kalantari}
  \setlength{\tabcolsep}{2.2pt}
  \renewcommand{\arraystretch}{1.05}
  \resizebox{\columnwidth}{!}{
    \begin{tabular}{c c|cccccc}
      \toprule\toprule
      Methods & Venue 
      & MUSIQ$\uparrow$ 
      & MANIQA$\uparrow$ 
      & LIQE$\uparrow$ 
      & HyperIQA$\uparrow$ 
      & DBCNN$\uparrow$ 
      & NIMA$\uparrow$ \\
      \hline
      AHDR~\cite{yan2019attention}     & CVPR'19  & 63.7260 & 0.3345 & 3.1729 & 0.5385 & 0.5457 & 4.3294 \\
      HDRTrans~\cite{liu2022ghost}     & ECCV'22  & 64.9563 & \textcolor{thirdcolor}{0.3447} & 3.2742 & 0.5505 & 0.5624 & \textcolor{thirdcolor}{4.3488} \\
      DiffHDR~\cite{yan2023toward}     & TCSVT'23 & \textcolor{thirdcolor}{64.9680} & \textcolor{blue}{0.3452} & \textcolor{thirdcolor}{3.2859} & \textcolor{thirdcolor}{0.5507} & \textcolor{blue}{0.5642} & \textcolor{blue}{4.3518} \\
      SAFNet~\cite{kong2024safnet}     & ECCV'24  & 64.7381 & 0.3442 & 3.2687 & 0.5472 & 0.5617 & 4.3449 \\
      LFDiff~\cite{hu2024generating}   & CVPR'24  & \textcolor{red}{65.0523} & \textcolor{red}{0.3458} & \textcolor{red}{3.2992} & \textcolor{red}{0.5526} & \textcolor{thirdcolor}{0.5641} & 4.3405 \\
      AFUNet~\cite{li2025afunet}       & ICCV'25  & 64.8090 & 0.3425 & 3.2744 & 0.5497 & 0.5591 & 4.3405 \\
      Ours                             & -        & \textcolor{blue}{65.0218} & \textcolor{thirdcolor}{0.3447} & \textcolor{blue}{3.2894} & \textcolor{blue}{0.5513} & \textcolor{red}{0.5648} & \textcolor{red}{4.3537} \\
      \bottomrule\bottomrule
    \end{tabular}
  }
  \vspace{-0.2cm}
\end{table}

\begin{table}[h!]
  \centering
  \caption{Ablation study of decision-level variants on Kalantari's dataset~\cite{kalantari2017deep}. FCM, AGGA, and FB denote fine-grained contextual knowledge matching, agent-guided generative alignment, and perception--distortion feedback, respectively. The best results are highlighted in red.}
  \label{tab:ablation_kalantari}
  \footnotesize
  \setlength{\tabcolsep}{3.8pt}
  \renewcommand{\arraystretch}{0.95}
  \resizebox{\columnwidth}{!}{
    \begin{tabular}{c|ccc|cccc}
      \toprule
      Variant & FCM & AGGA & FB & PSNR-$\mu$ & PSNR-$L$ & SSIM-$\mu$ & SSIM-$L$ \\
      \midrule
      w/ RS    & \xmark & \xmark & \xmark & 44.14 & 42.31 & 0.9917 & 0.9883 \\
      w/o CTX  & \xmark & \cmark & \cmark & 44.51 & 42.62 & 0.9918 & 0.9893 \\
      w/o GA   & \cmark & \xmark & \cmark & 44.88 & 42.57 & 0.9921 & 0.9903 \\
      Full     & \cmark & \cmark & \cmark & \textcolor{red}{45.01} & \textcolor{red}{42.68} & \textcolor{red}{0.9922} & \textcolor{red}{0.9911} \\
      \bottomrule
    \end{tabular}
  }
  \vspace{-0.4cm}
\end{table}

\begin{figure*}[t]
\centering

\setlength{\tabcolsep}{1pt}
\renewcommand{\arraystretch}{0.0}

\newlength{\TelMethodW}
\setlength{\TelMethodW}{0.170\textwidth}

\newcommand{\scenegapTel}{0pt}

\newcommand{\methodrowgapTel}{2pt}

\newcommand{\namegapTel}{-5pt}

\newcommand{\methodtwoscenesTel}[3]{%
\begin{minipage}[t]{\TelMethodW}
    \centering
    \includegraphics[width=\linewidth]{#1}\\[\scenegapTel]
    \includegraphics[width=\linewidth]{#2}\\[\namegapTel]
    \makebox[\linewidth][c]{\scriptsize\textbf{#3}}%
\end{minipage}
}

\begin{tabular}{@{}ccccc@{}}

\methodtwoscenesTel
{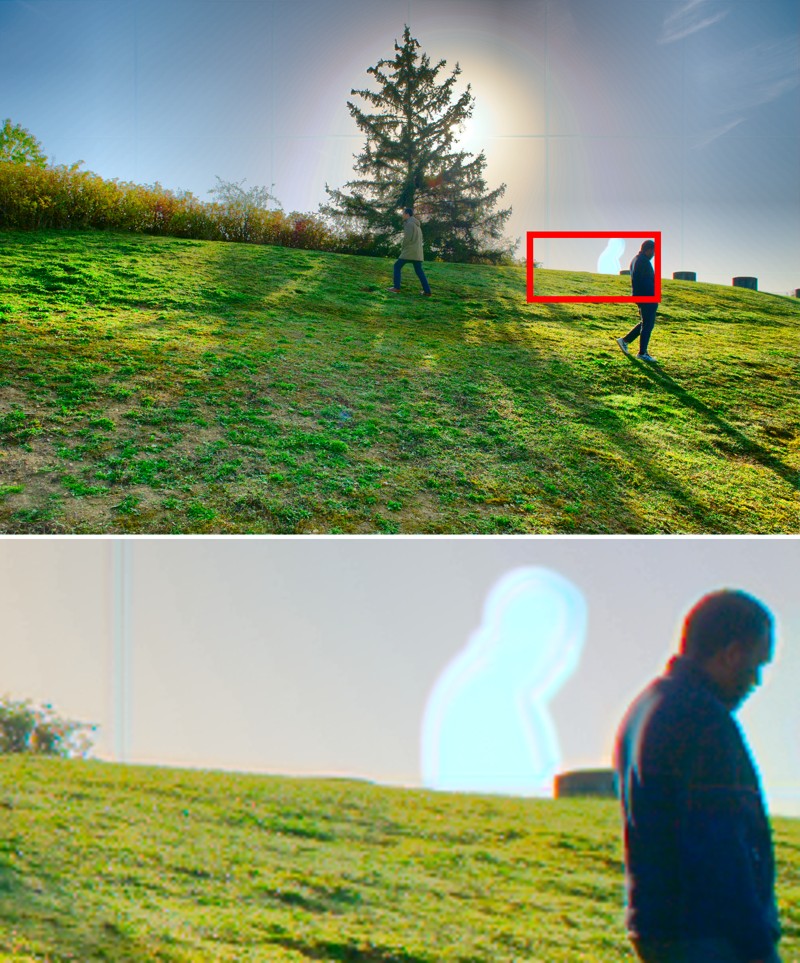}
{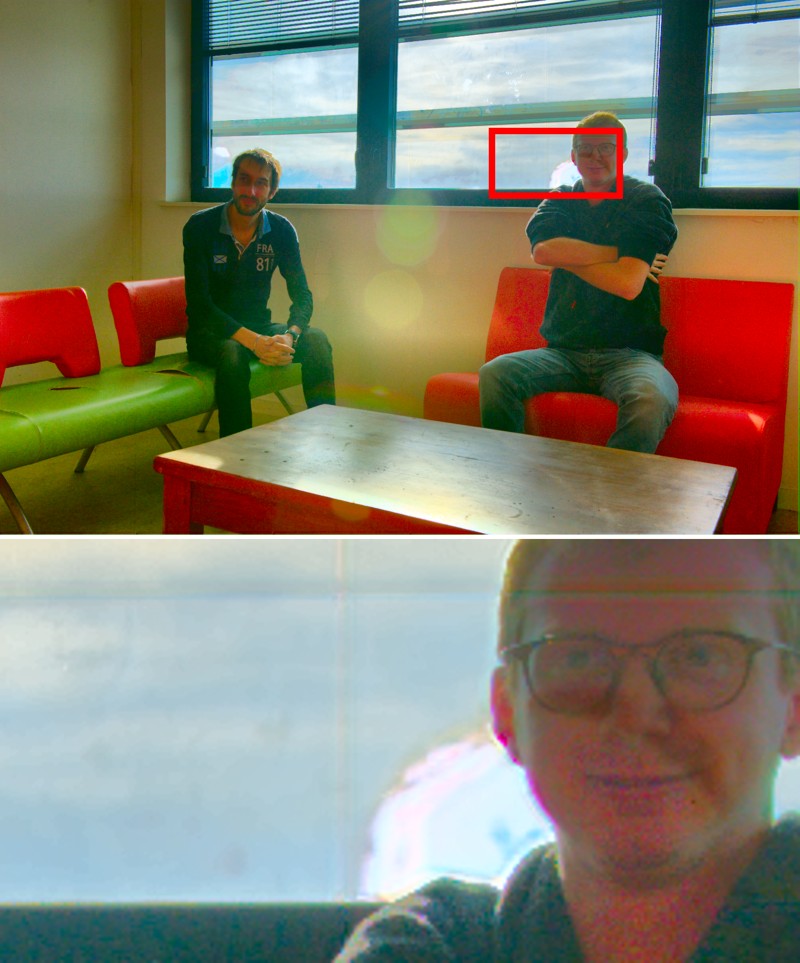}
{(a) DHDR~\cite{wu2018deep}}
&
\methodtwoscenesTel
{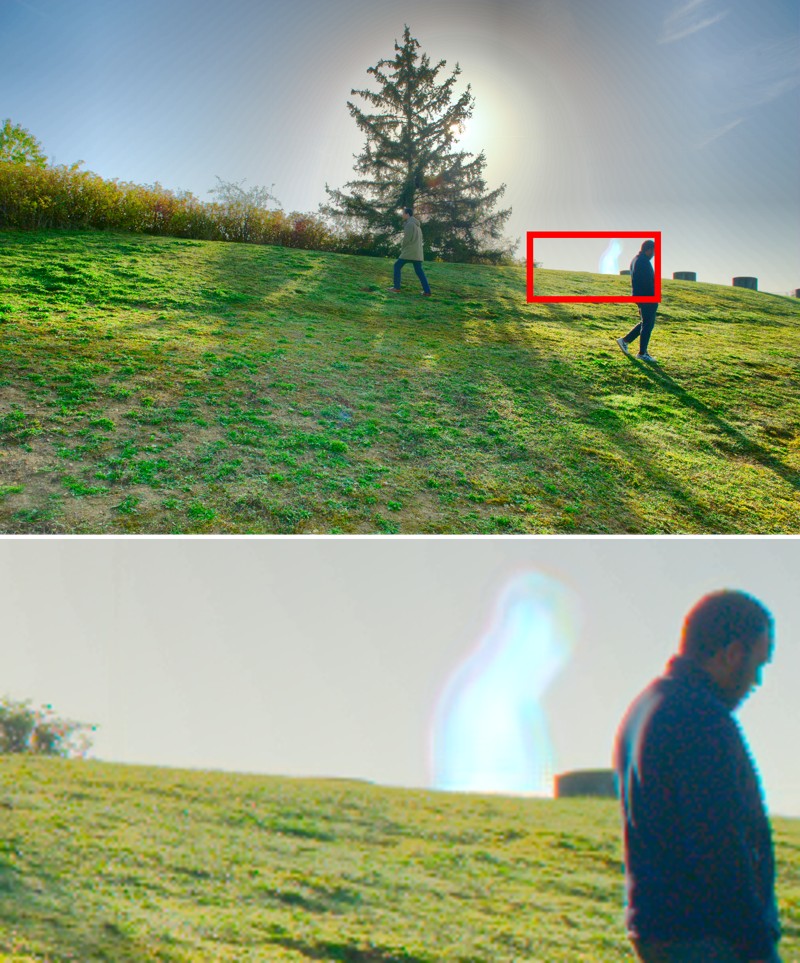}
{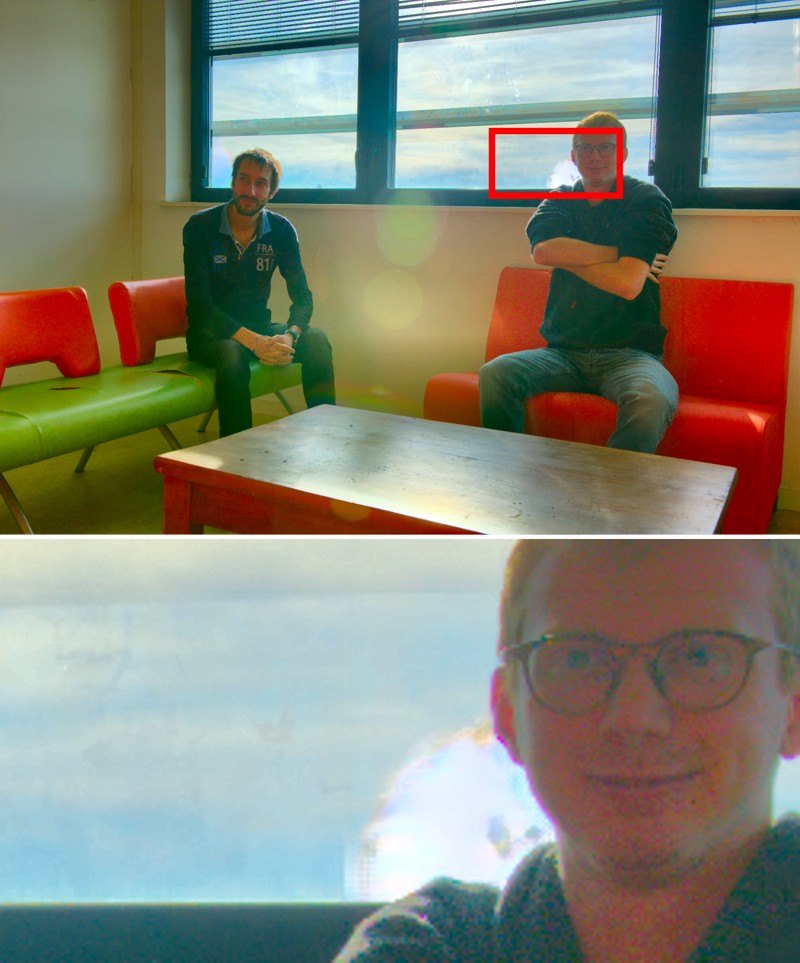}
{(b) AHDR~\cite{yan2019attention}}
&
\methodtwoscenesTel
{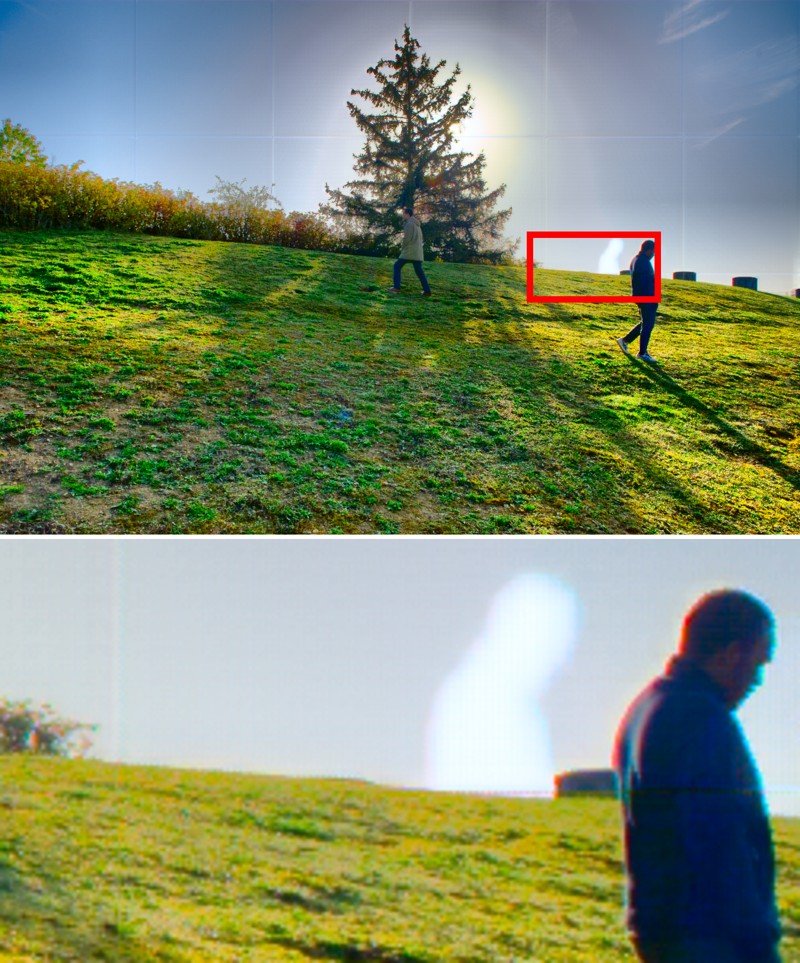}
{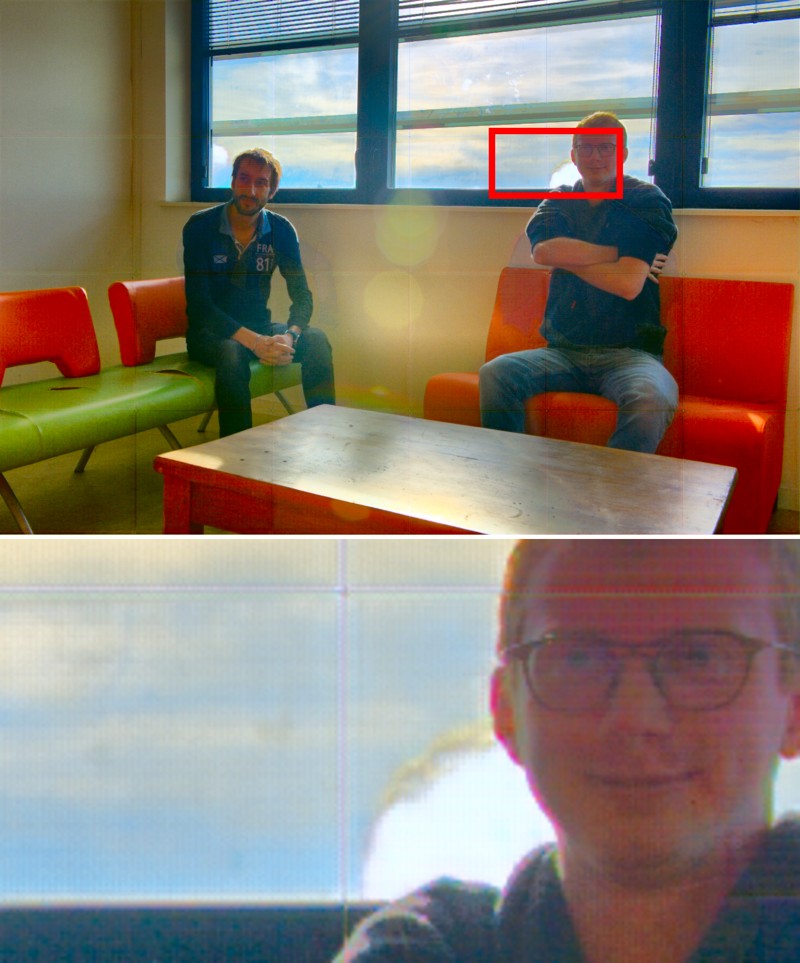}
{(c) NHDRR~\cite{yan2020deep}}
&
\methodtwoscenesTel
{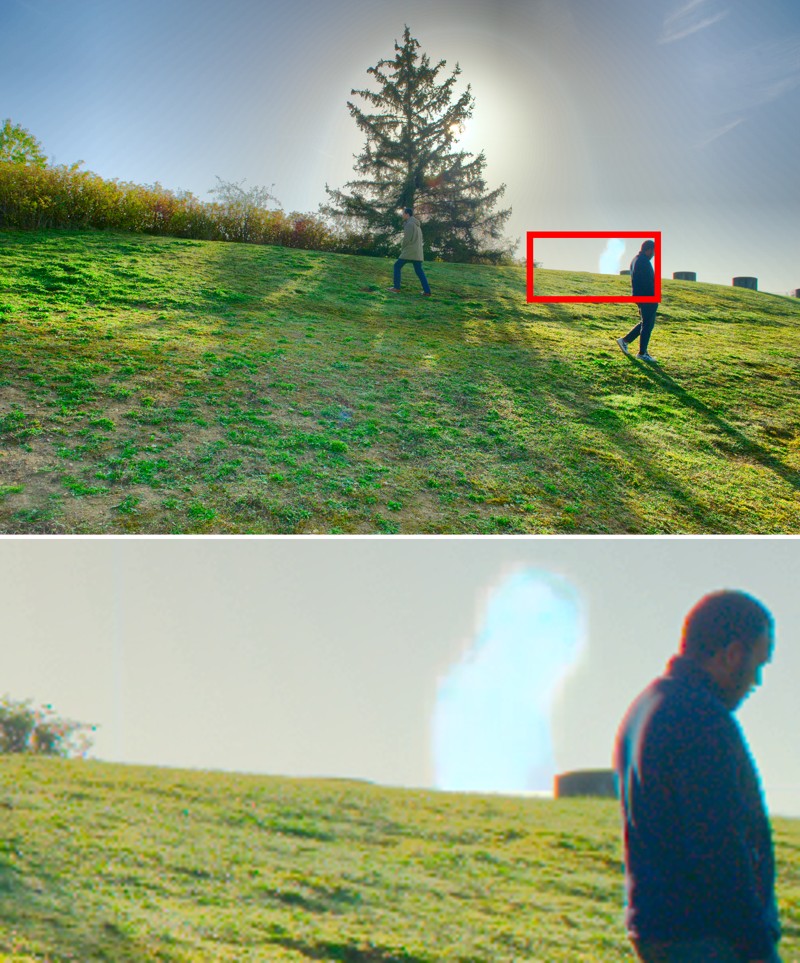}
{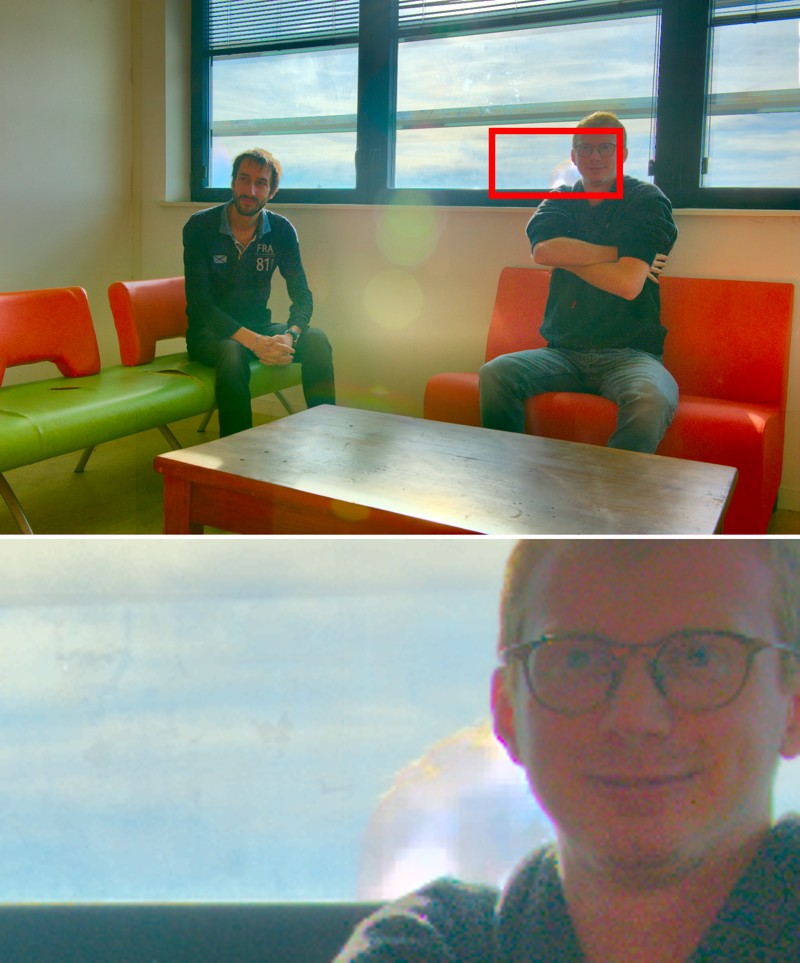}
{(d) HDRTrans~\cite{liu2022ghost}}
&
\methodtwoscenesTel
{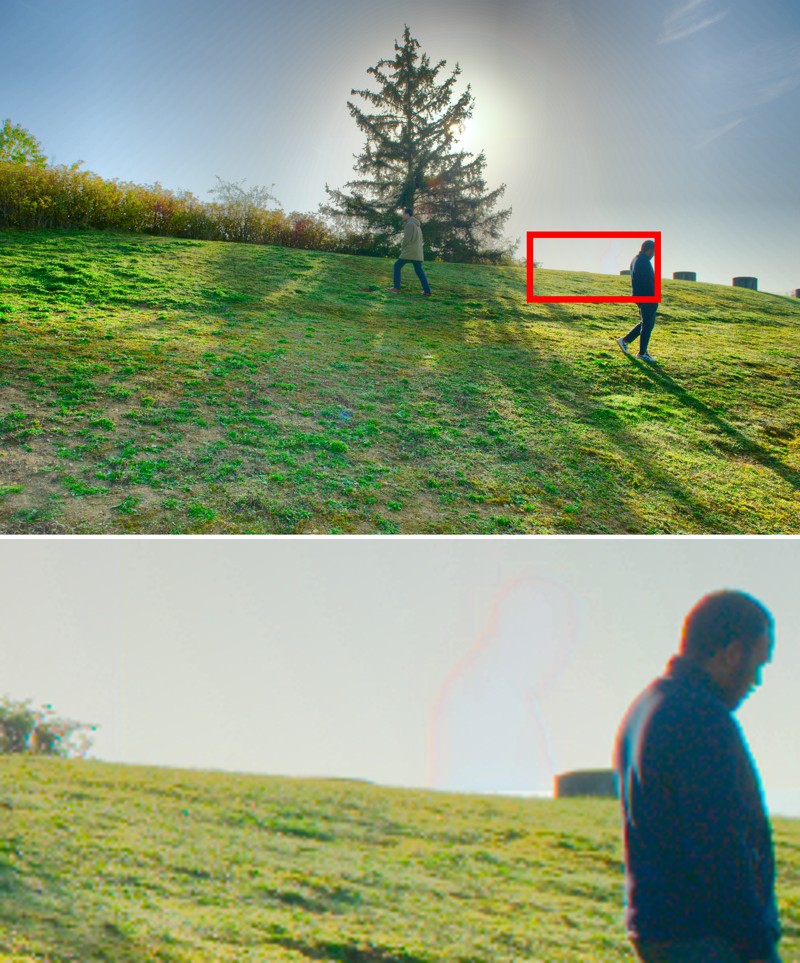}
{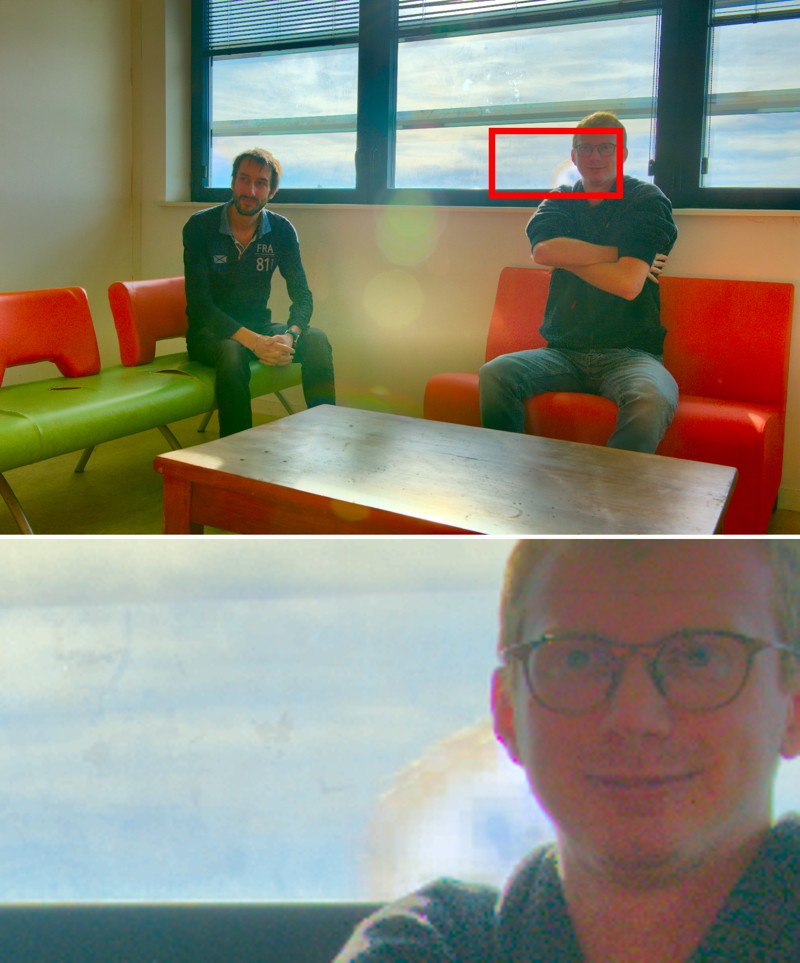}
{(e) SCTNet~\cite{tel2023alignment}}
\\[\methodrowgapTel]

\methodtwoscenesTel
{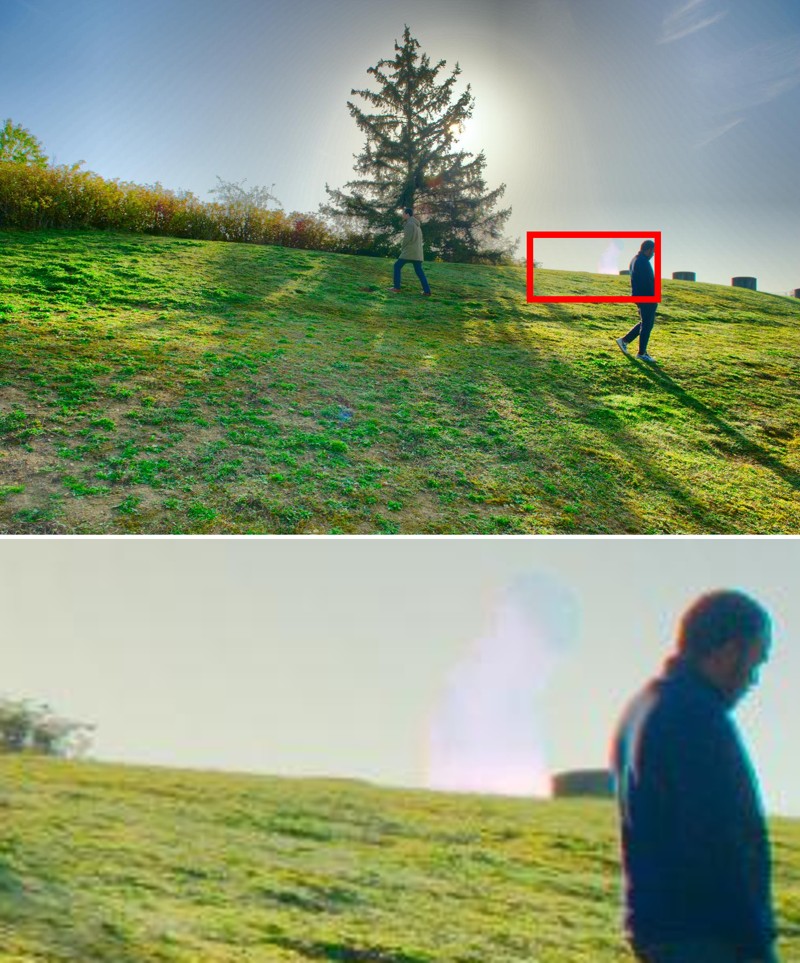}
{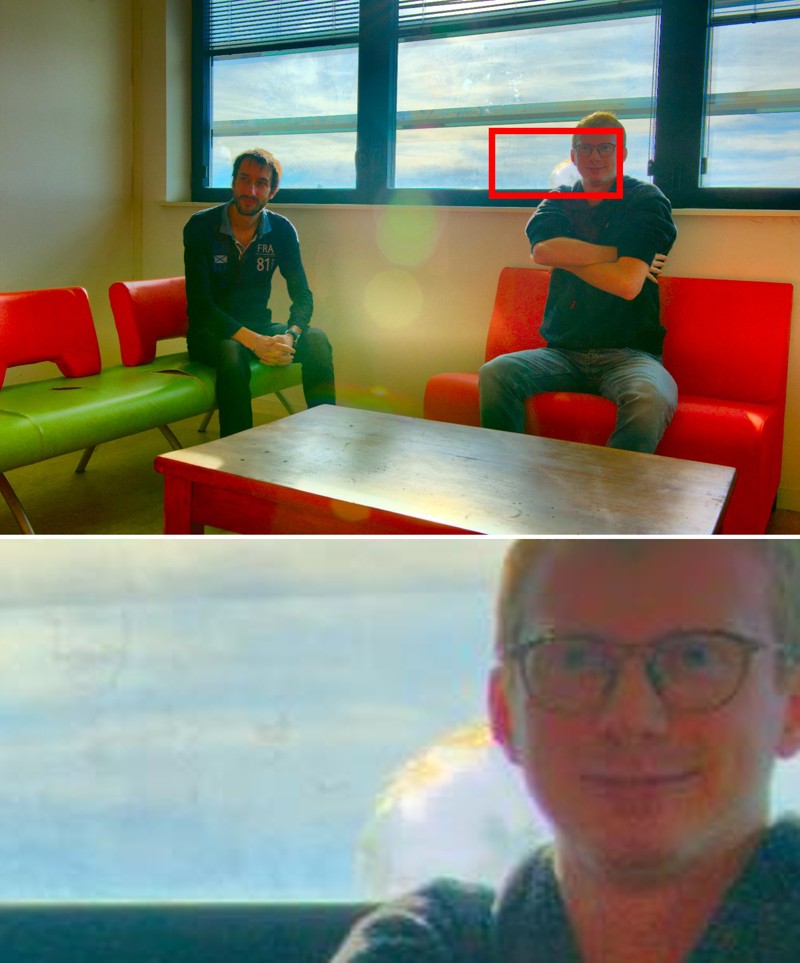}
{(f) SMHDR~\cite{ni2025semantic}}
&
\methodtwoscenesTel
{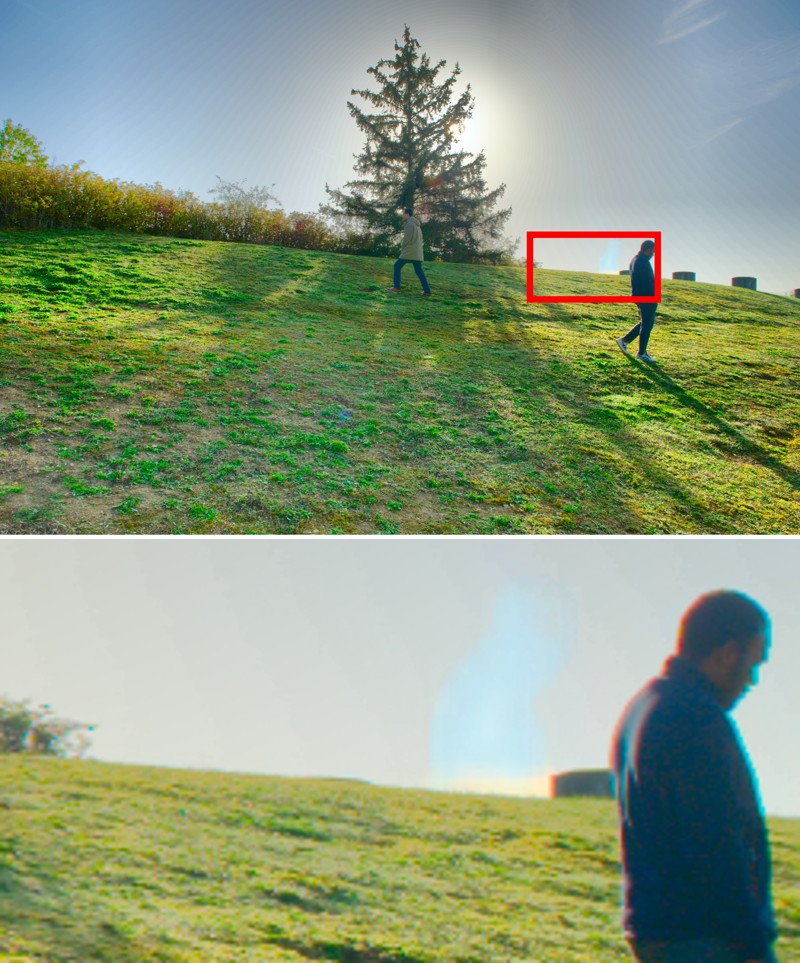}
{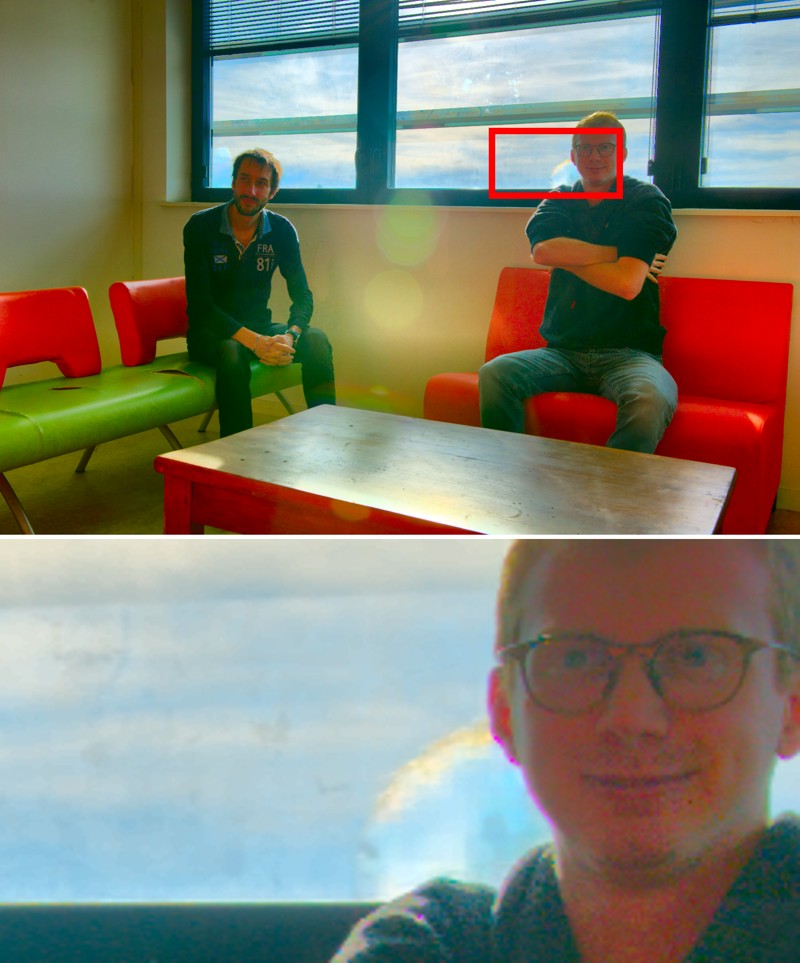}
{(g) AFUNet~\cite{li2025afunet}}
&
\methodtwoscenesTel
{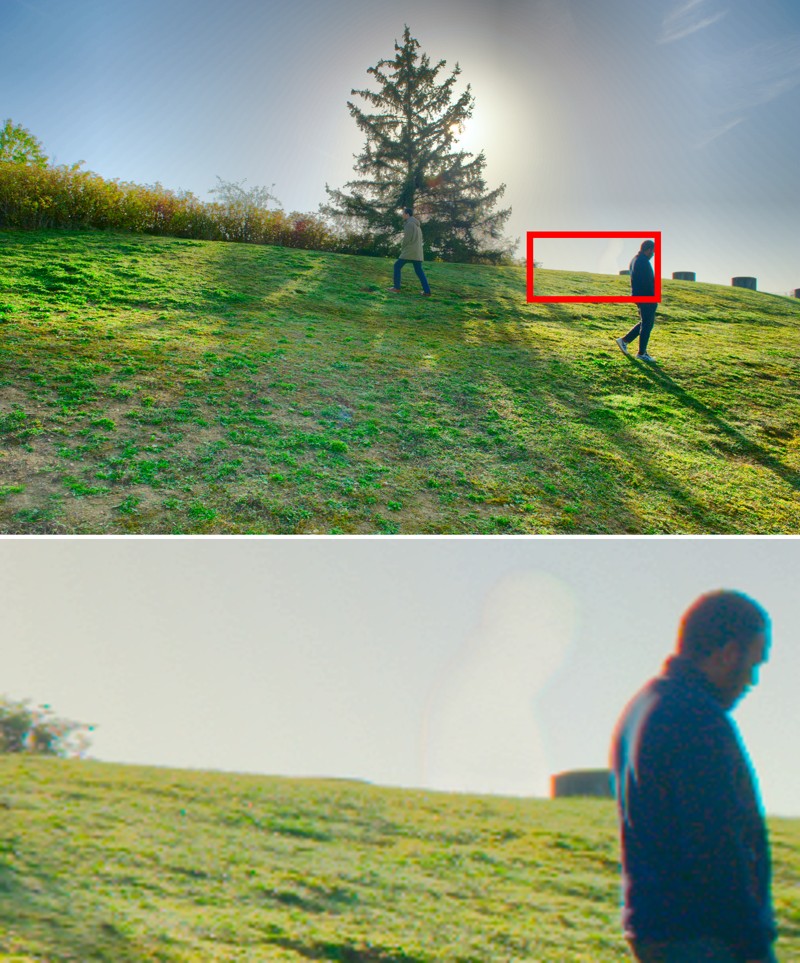}
{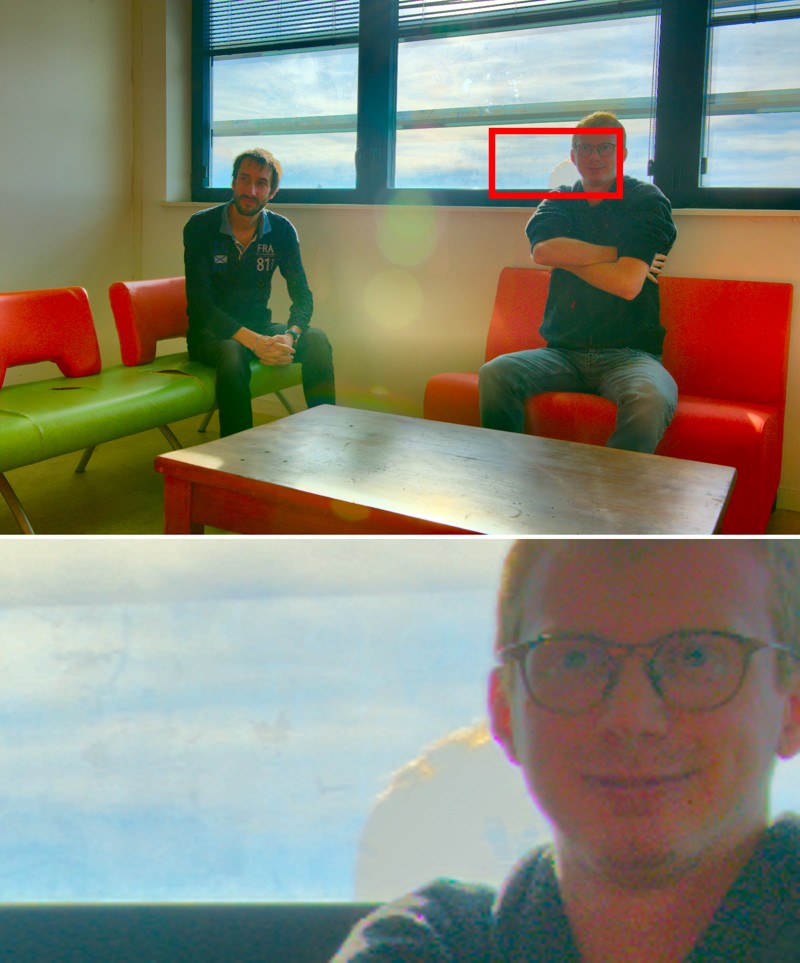}
{(h) DDPF-PR~\cite{zhou2026high}}
&
\methodtwoscenesTel
{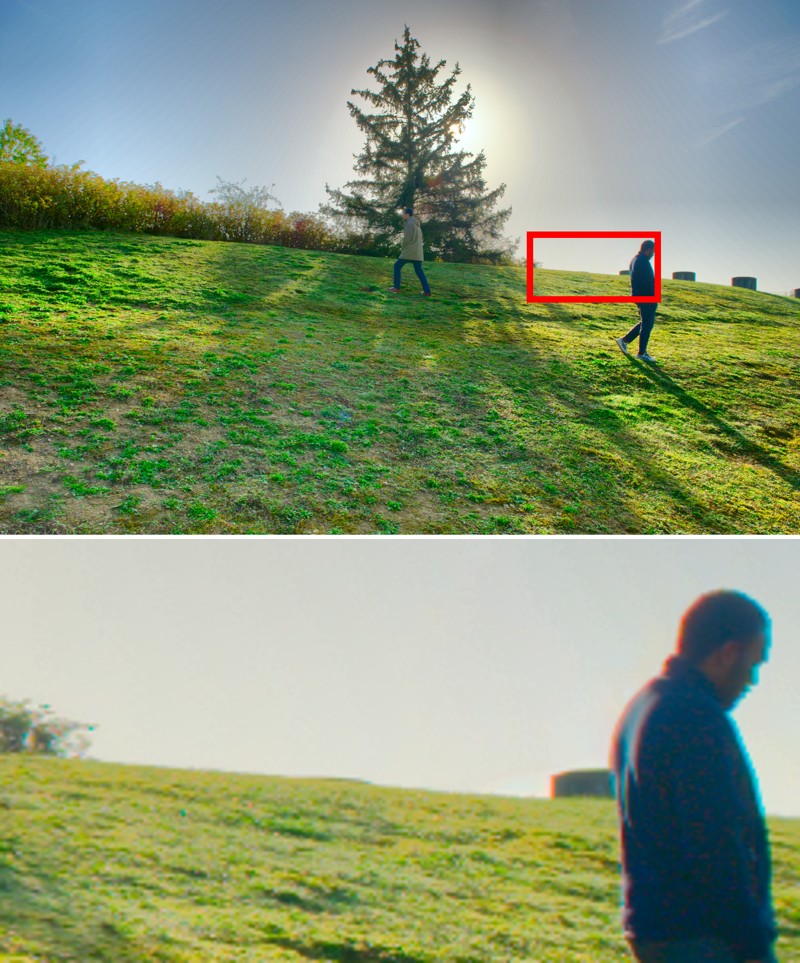}
{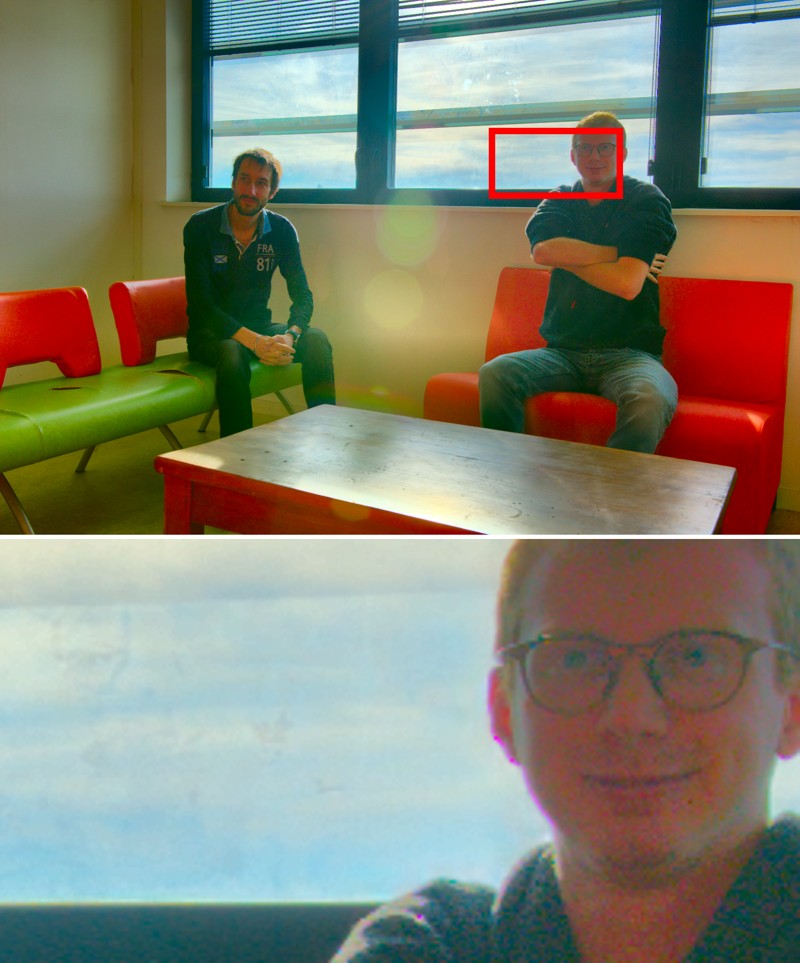}
{(i) HDRAgent}
&
\methodtwoscenesTel
{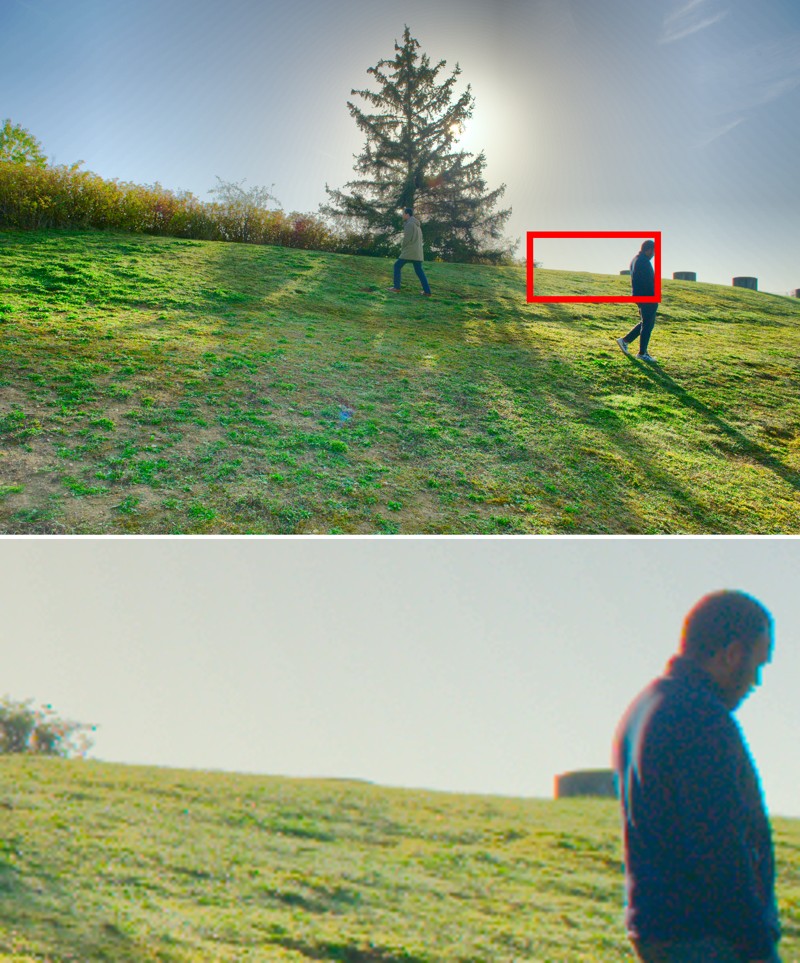}
{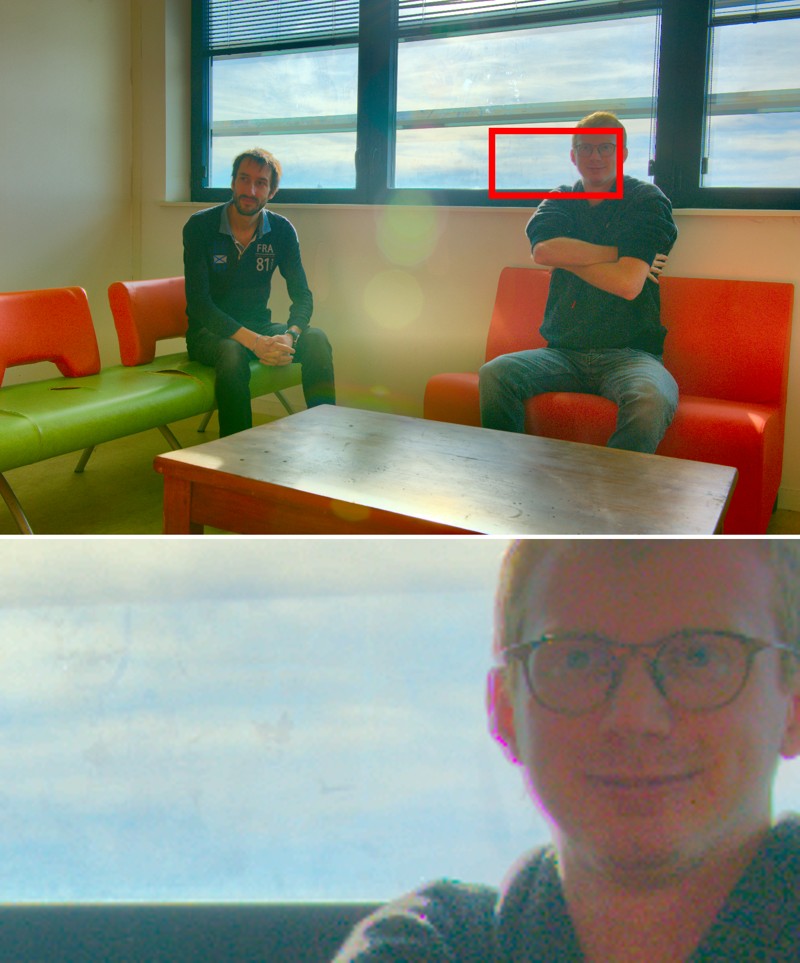}
{(j) GT}

\end{tabular}

\caption{
Visual comparison of different methods on dynamic scenes from Tel's dataset~\cite{tel2023alignment}.
Our HDRAgent yields cleaner local structures and fewer ghosting artifacts, with results closer to the reference HDR images.
}
\label{fig:visual_comparison_tel}
\vspace{-0.4cm}
\end{figure*}

\begin{figure*}[t]
\centering

\setlength{\tabcolsep}{1pt}
\renewcommand{\arraystretch}{0.0}

\newlength{\HuMethodW}
\setlength{\HuMethodW}{0.170\textwidth}

\newcommand{\scenegapHu}{0pt}

\newcommand{\methodrowgapHu}{2pt}

\newcommand{\namegapHu}{-5pt}

\newcommand{\methodtwoscenesHu}[3]{%
\begin{minipage}[t]{\HuMethodW}
    \centering
    \includegraphics[width=\linewidth]{#1}\\[\scenegapHu]
    \includegraphics[width=\linewidth]{#2}\\[\namegapHu]
    \makebox[\linewidth][c]{\scriptsize\textbf{#3}}%
\end{minipage}
}

\begin{tabular}{@{}ccccc@{}}

\methodtwoscenesHu
{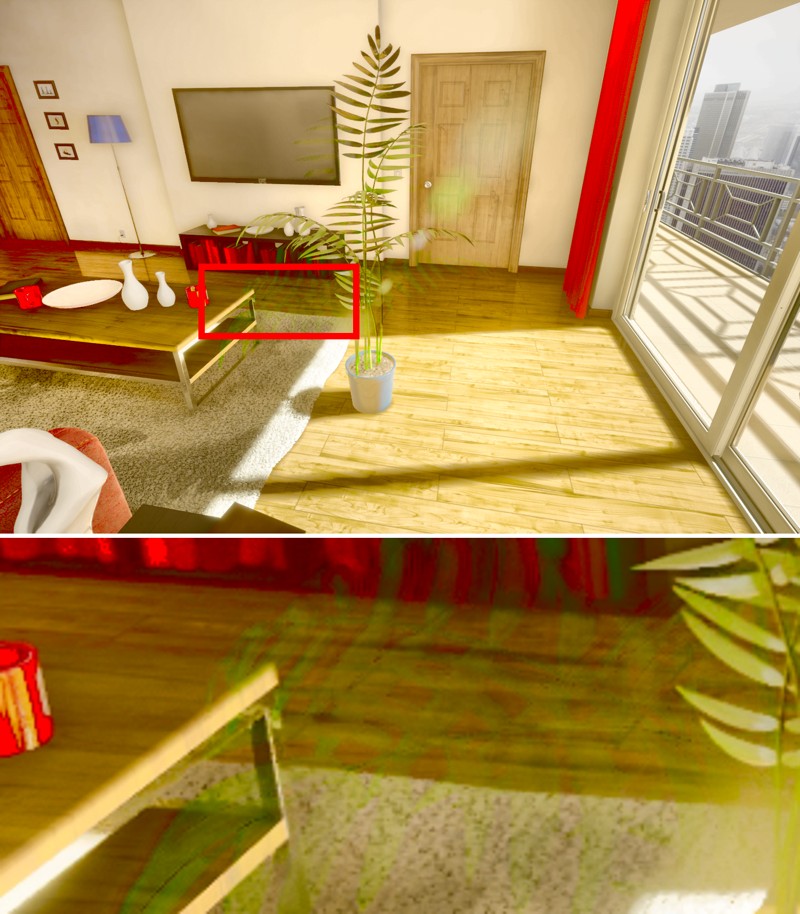}
{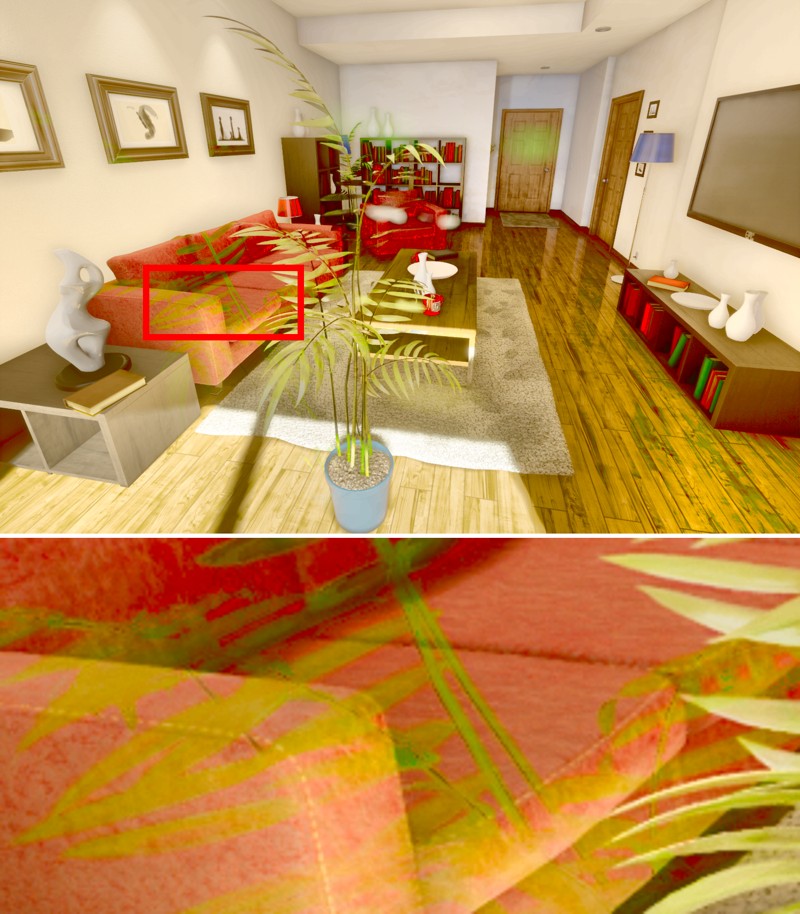}
{(a) AHDR~\cite{yan2019attention}}
&
\methodtwoscenesHu
{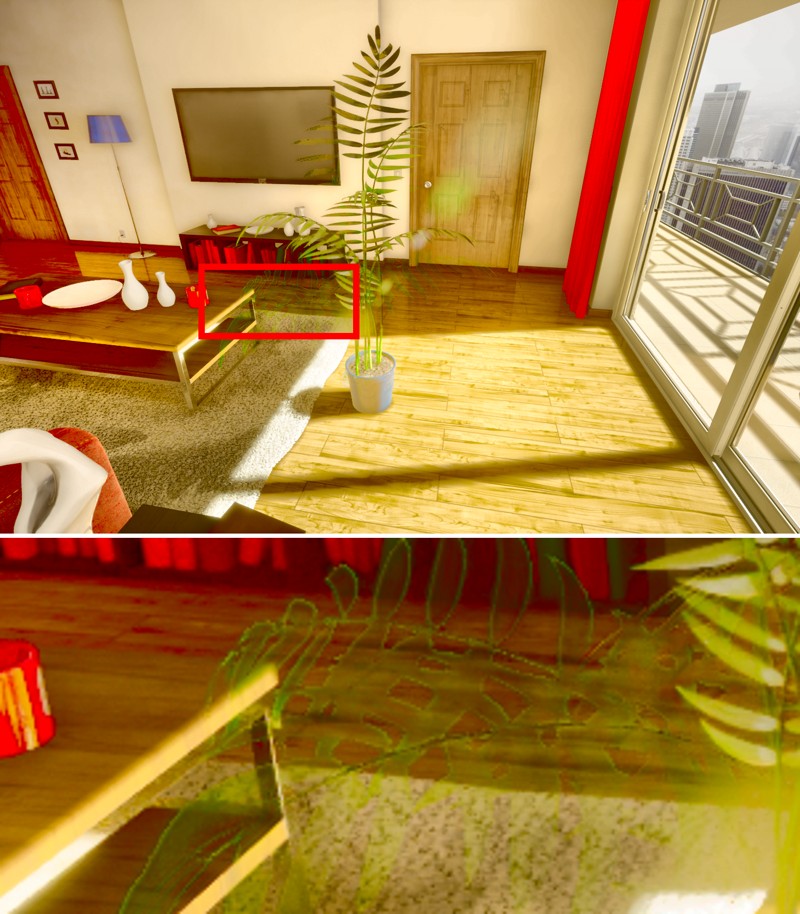}
{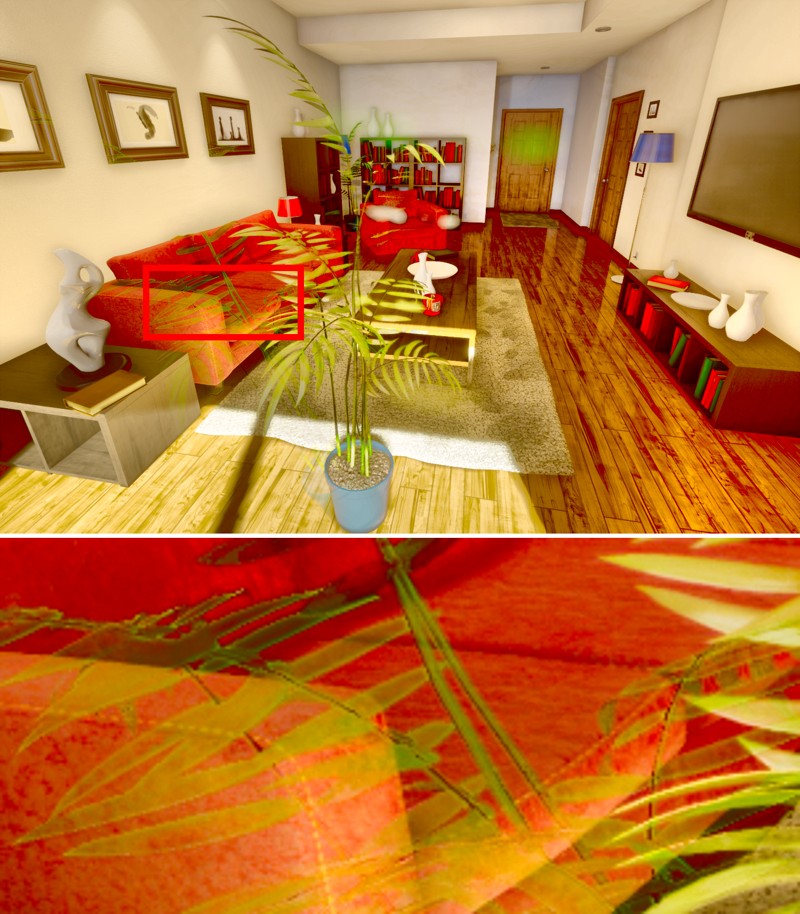}
{(b) HDRTrans~\cite{liu2022ghost}}
&
\methodtwoscenesHu
{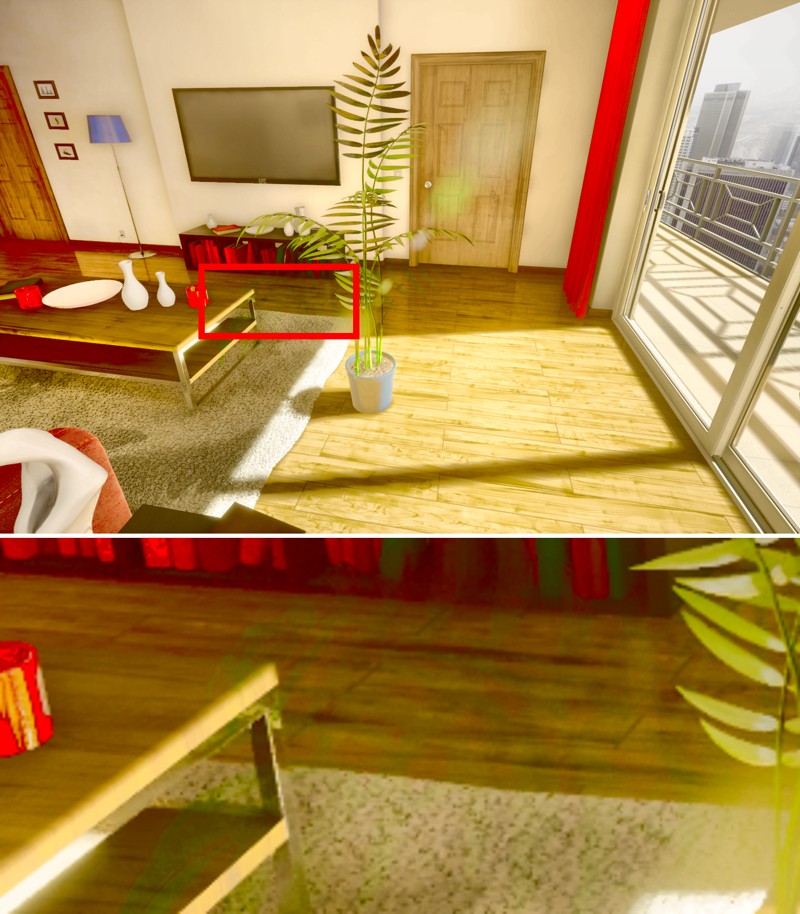}
{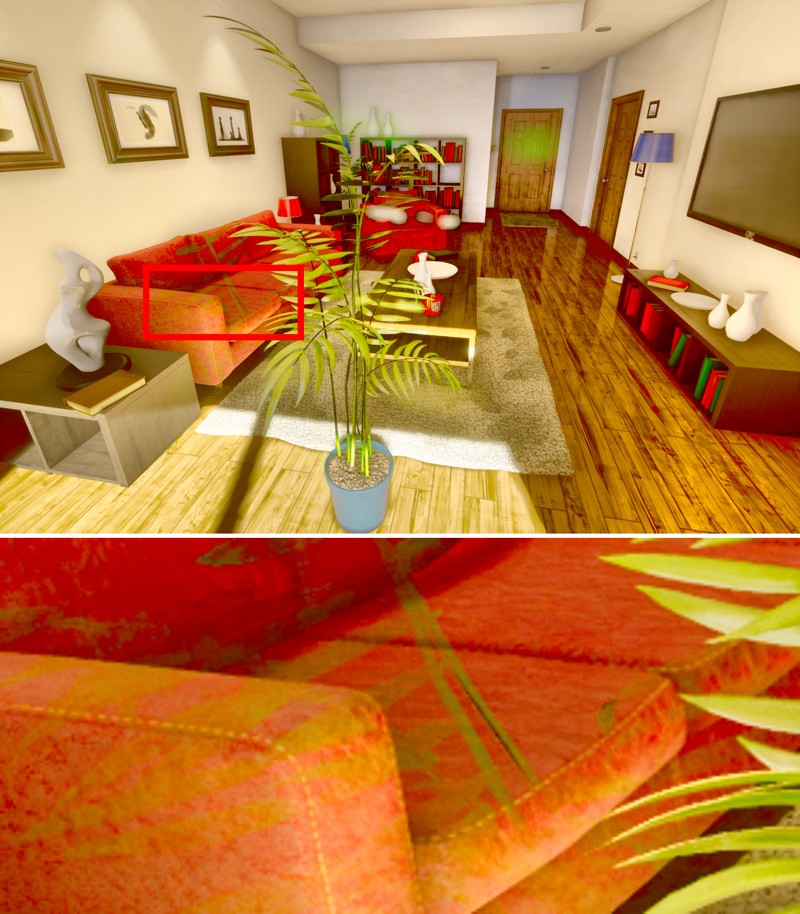}
{(c) SCTNet~\cite{tel2023alignment}}
&
\methodtwoscenesHu
{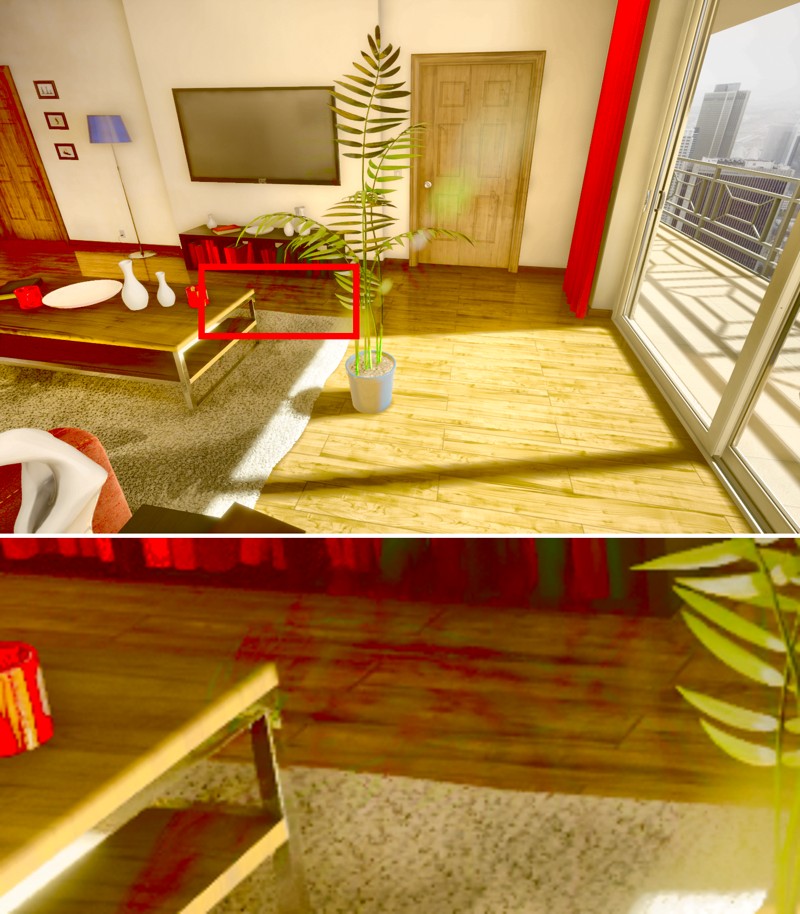}
{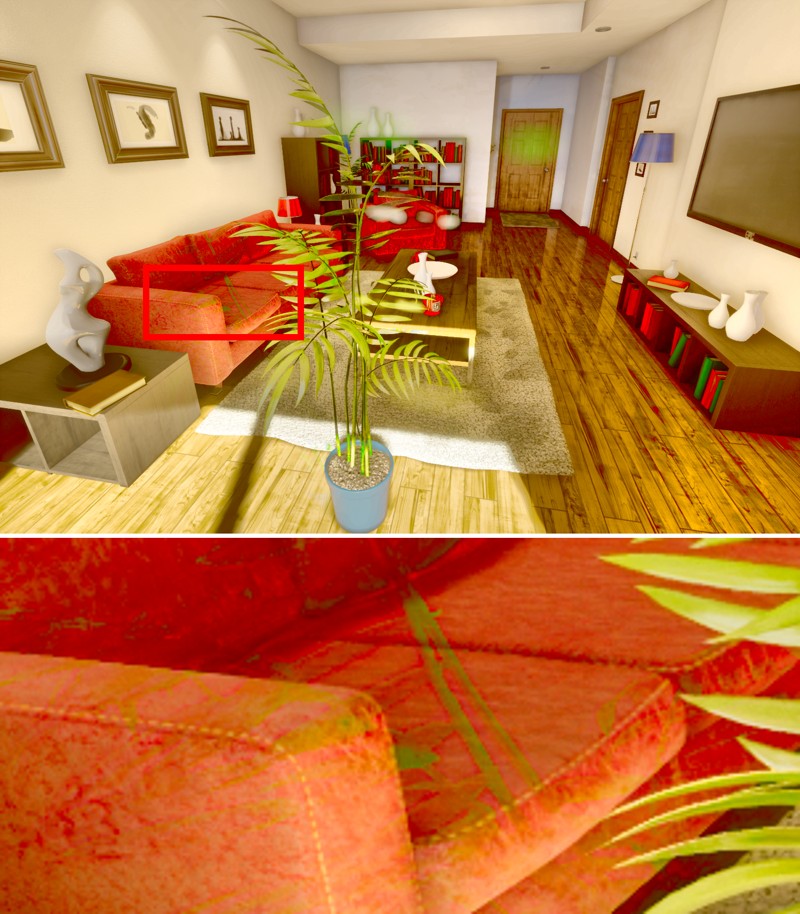}
{(d) DiffHDR~\cite{yan2023toward}}
&
\methodtwoscenesHu
{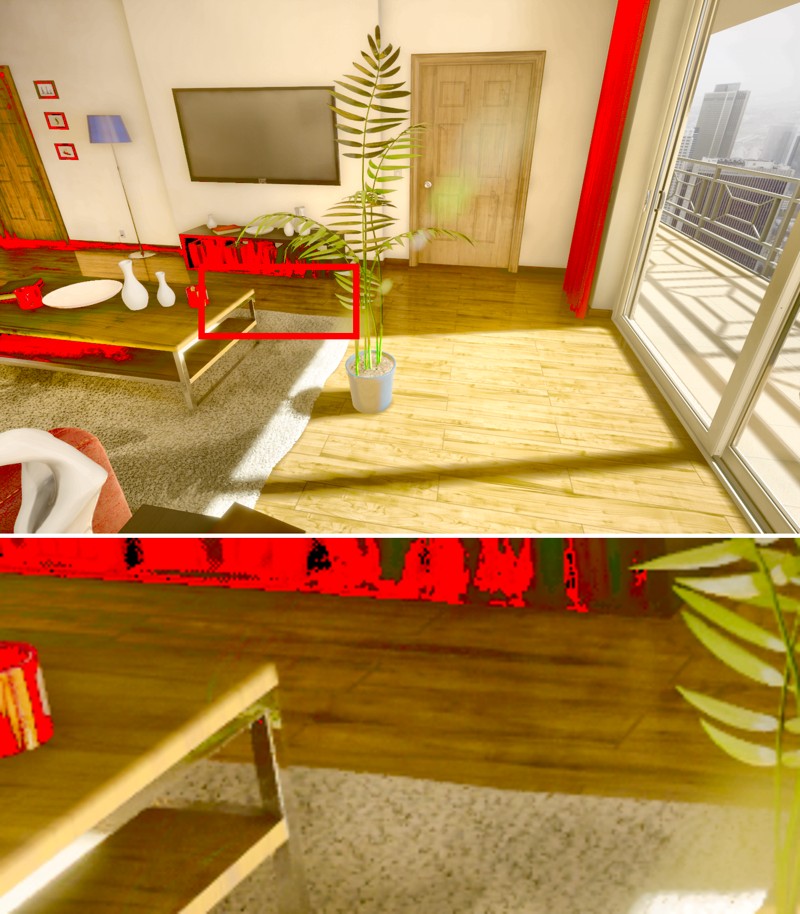}
{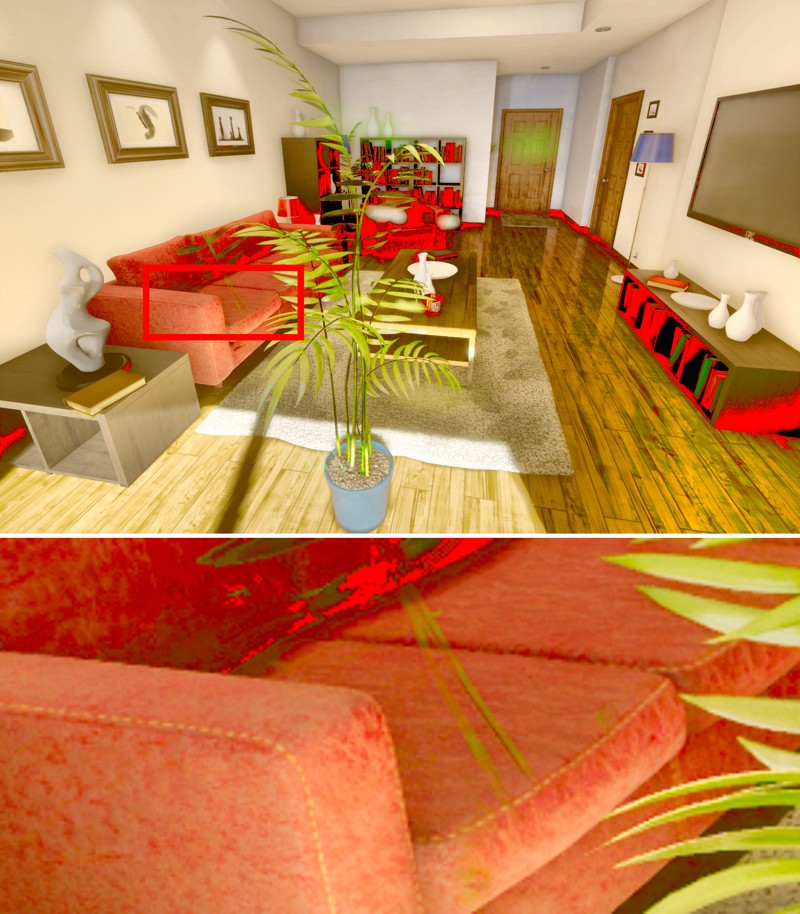}
{(e) SAFNet~\cite{kong2024safnet}}
\\[\methodrowgapHu]

\methodtwoscenesHu
{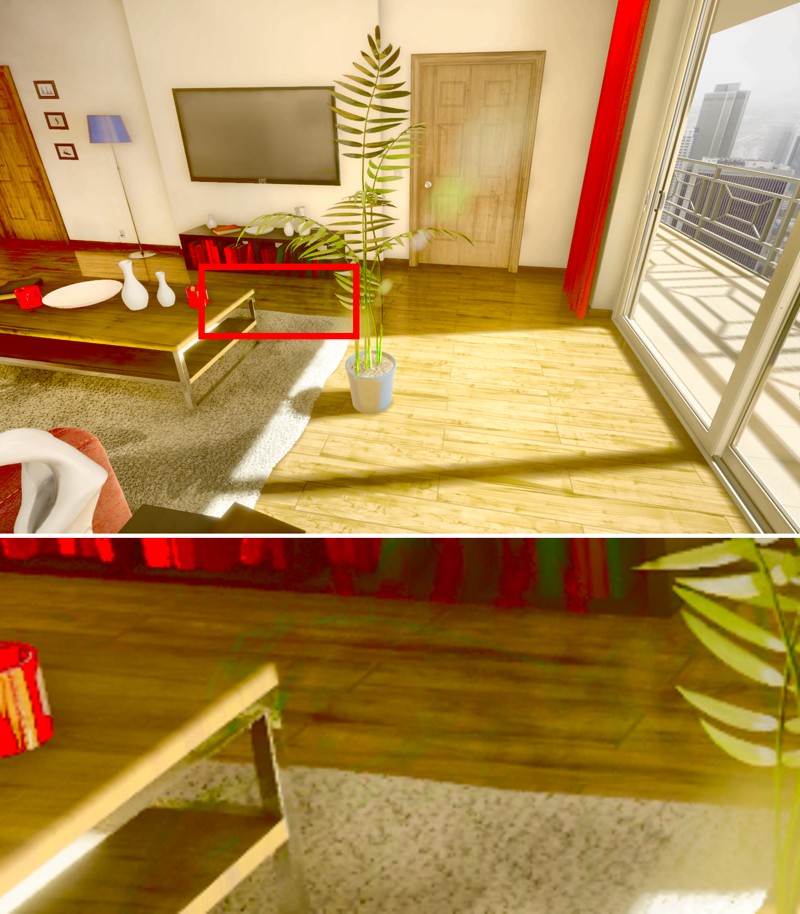}
{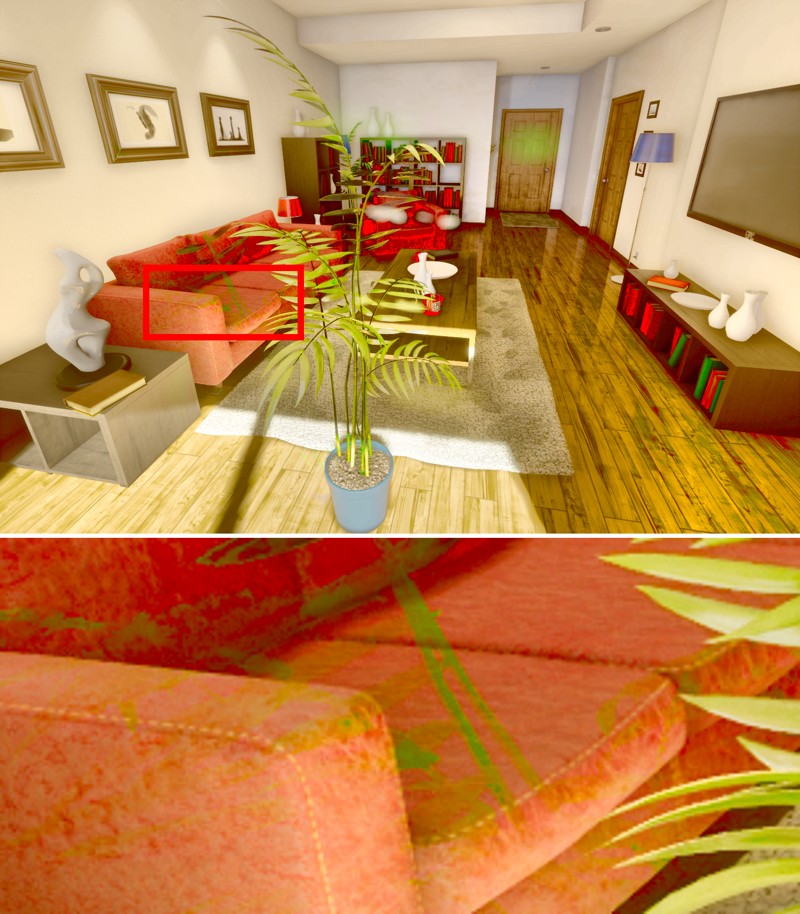}
{(f) LFDiff~\cite{hu2024generating}}
&
\methodtwoscenesHu
{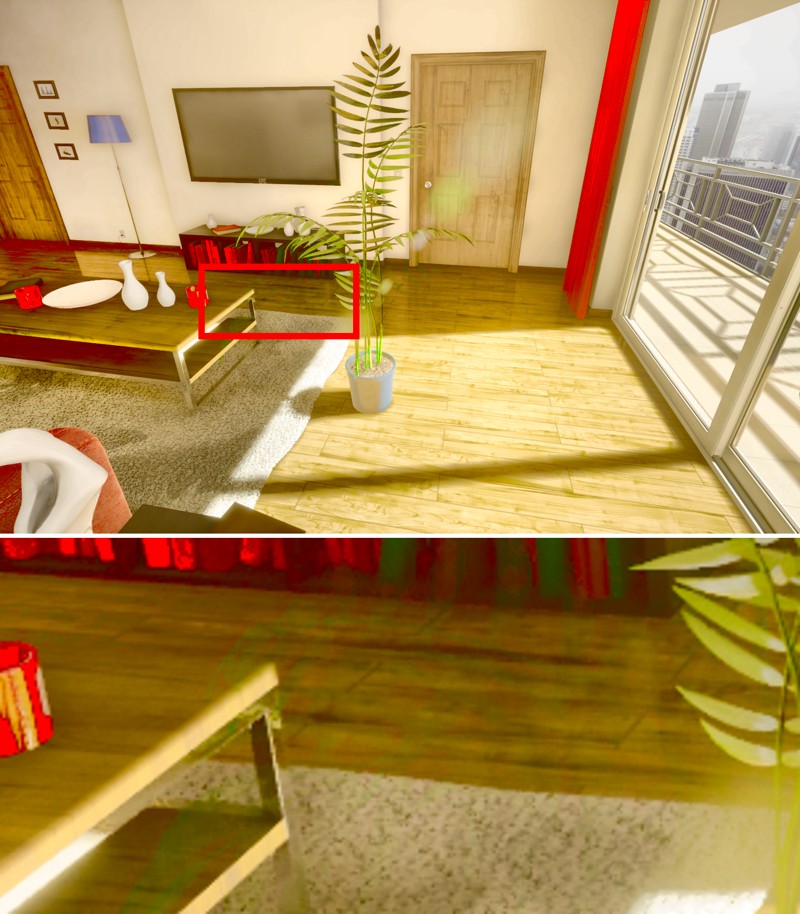}
{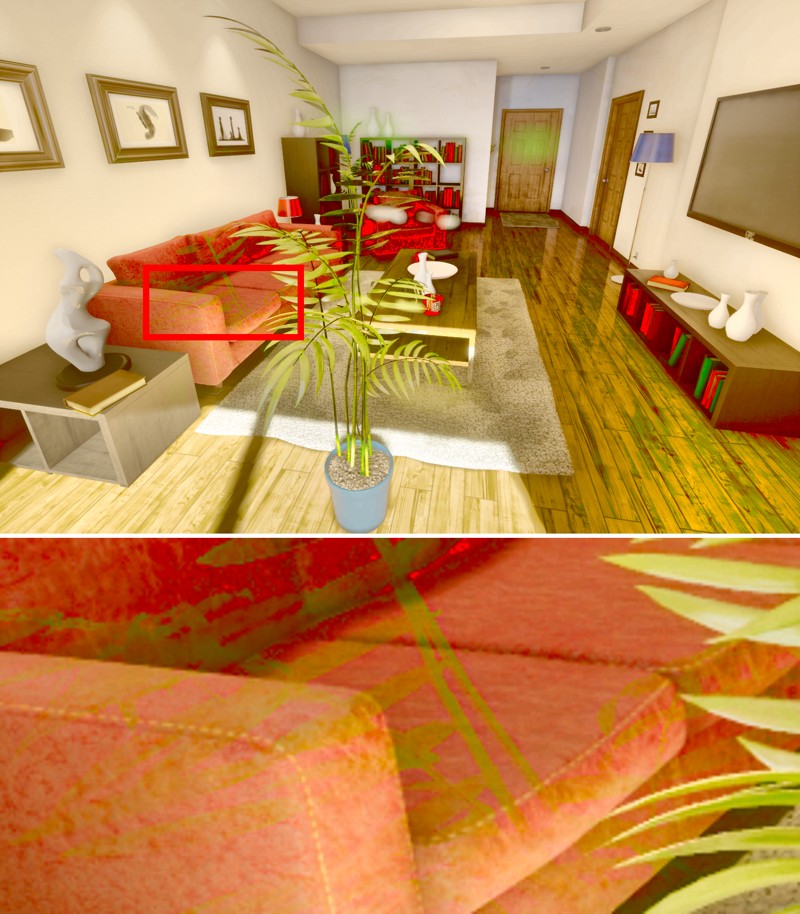}
{(g) AFUNet~\cite{li2025afunet}}
&
\methodtwoscenesHu
{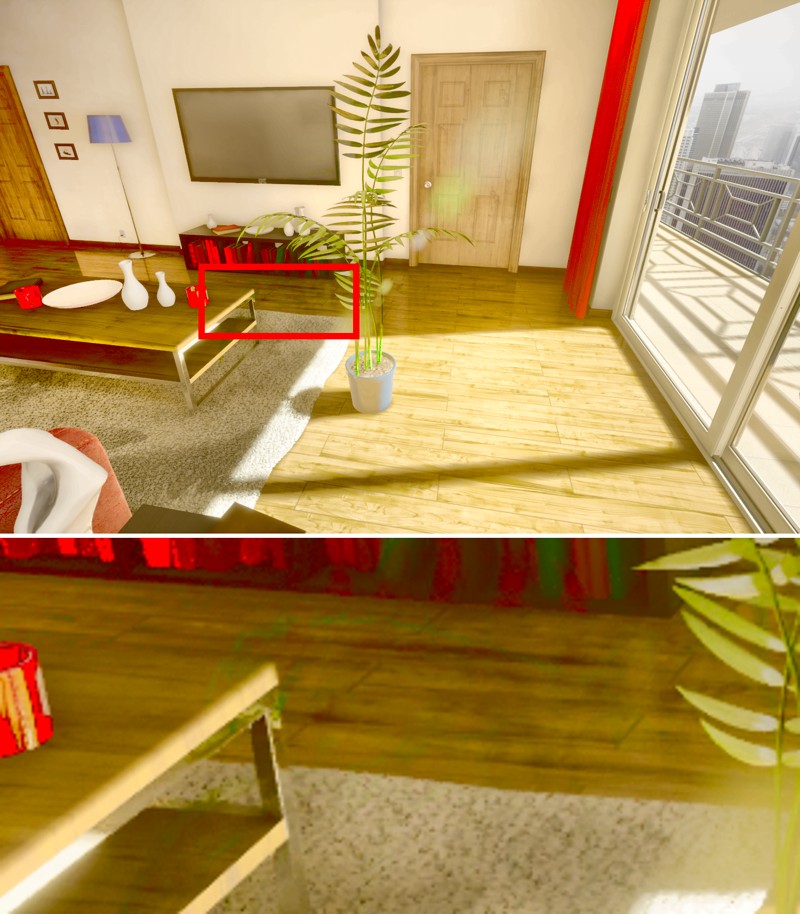}
{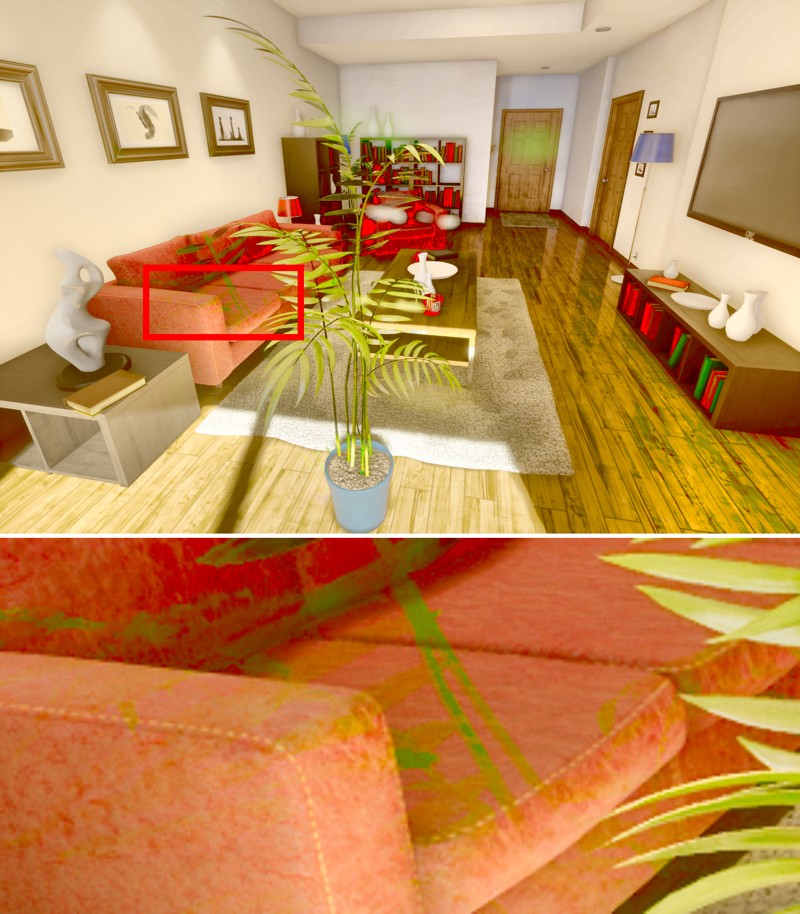}
{(h) DDPF-PR~\cite{zhou2026high}}
&
\methodtwoscenesHu
{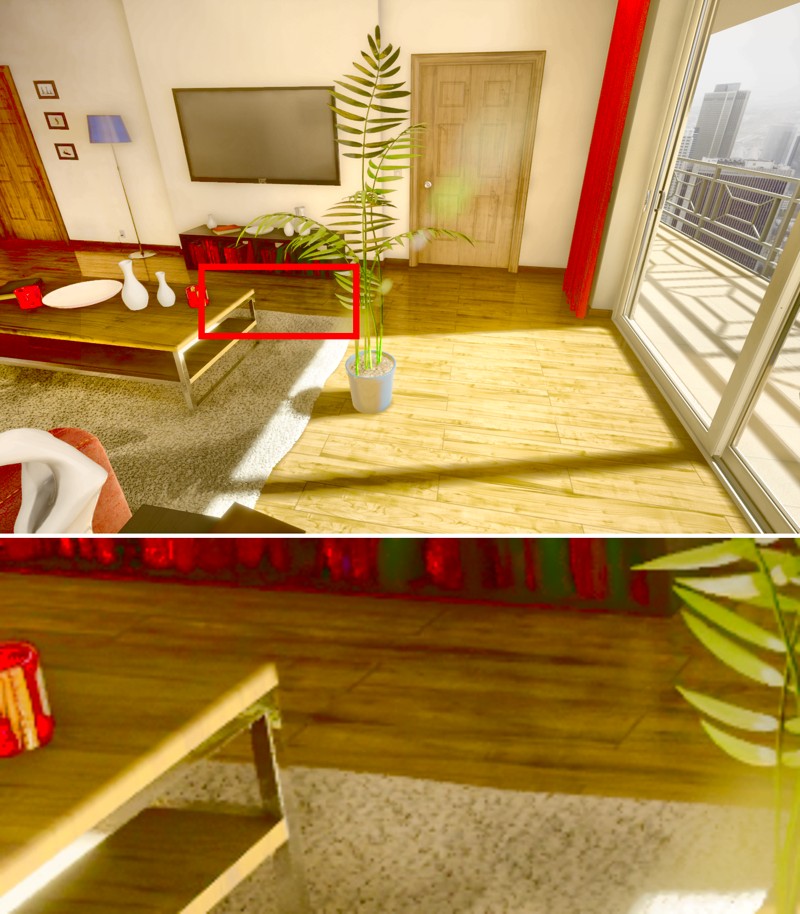}
{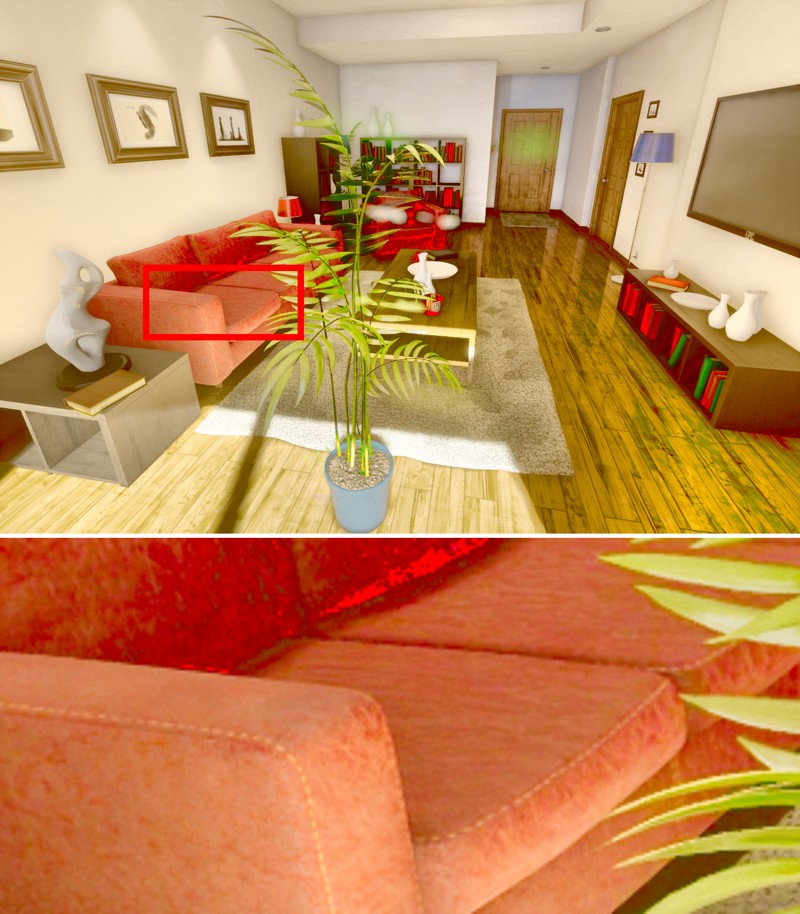}
{(i) HDRAgent}
&
\methodtwoscenesHu
{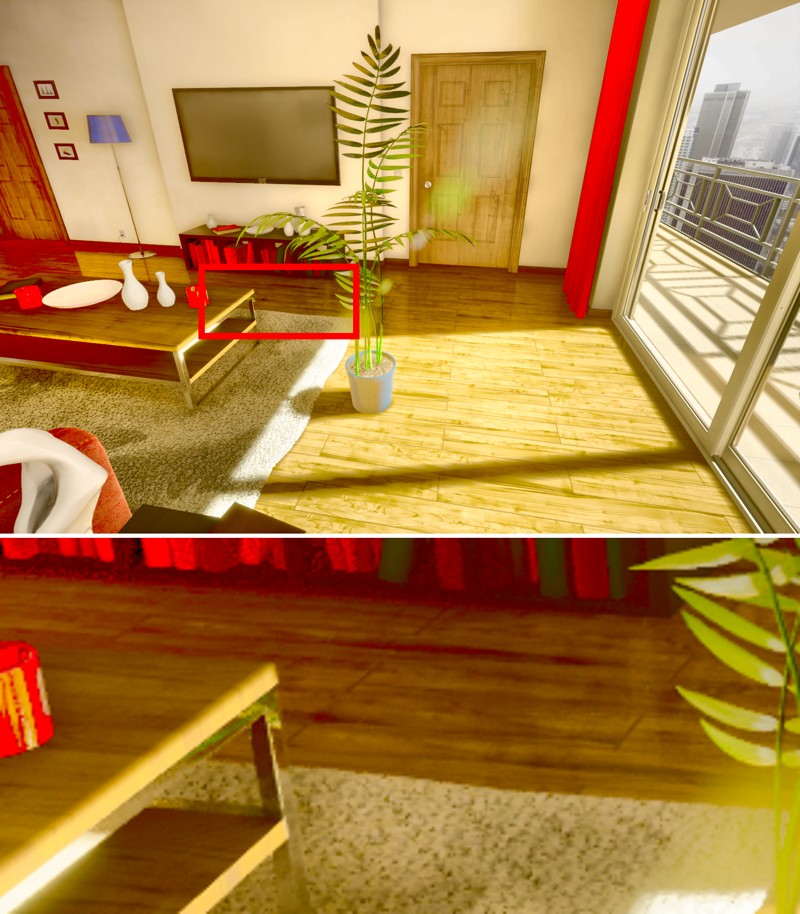}
{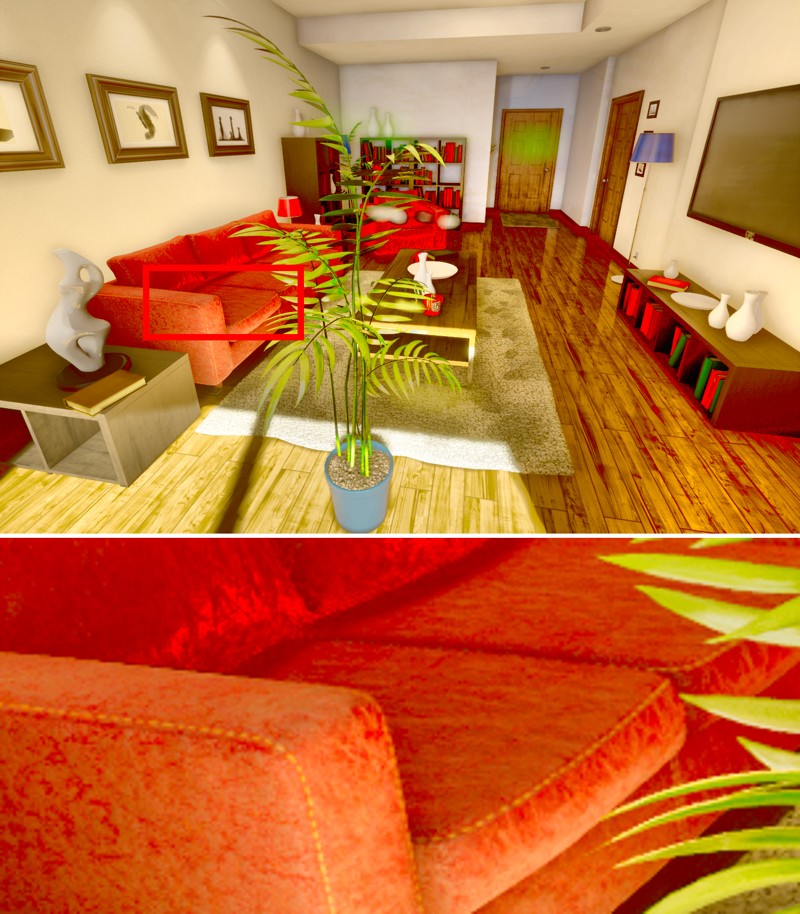}
{(j) GT}

\end{tabular}

\caption{
Visual comparison of different methods on dynamic scenes from Hu's dataset~\cite{hu2020sensor}.
Our HDRAgent yields cleaner local structures and fewer ghosting artifacts, with results closer to the reference HDR images.
}
\label{fig:visual_comparison_hu}
\vspace{-0.4cm}
\end{figure*}

\subsection{Qualitative Evaluation}

We conduct qualitative comparisons on three representative multi-exposure HDR benchmarks, as shown in Figs.~\ref{fig:visual_comparison_sig}, \ref{fig:visual_comparison_tel}, and \ref{fig:visual_comparison_hu}. These examples cover challenging dynamic scenarios with large foreground motion, saturated highlights, occlusions, and severe cross-exposure appearance changes. The cropped regions provide a closer inspection of local ghosting artifacts and structural inconsistencies.

Fig.~\ref{fig:visual_comparison_sig} presents visual comparisons on Kalantari's dataset~\cite{kalantari2017deep}. In these real dynamic scenes, the moving foreground objects introduce severe cross-exposure misalignment, especially around the body boundaries and background structures. Early CNN-based methods such as DHDR~\cite{wu2018deep}, AHDR~\cite{yan2019attention}, and NHDRR~\cite{yan2020deep} can recover the overall HDR appearance, but they still leave noticeable duplicated edges and semi-transparent ghosting in the enlarged regions. HDR-GAN~\cite{niu2021hdr} and HDRTrans~\cite{liu2022ghost} improve local contrast and global appearance, yet their results still contain residual misalignment around motion boundaries. Recent methods such as SCTNet~\cite{tel2023alignment}, SAFNet~\cite{kong2024safnet}, DiffHDR~\cite{yan2023toward}, LFDiff~\cite{hu2024generating}, and AFUNet~\cite{li2025afunet} further reduce obvious ghosting, but local artifacts remain visible in high-mismatch regions, such as blurred contours, color bleeding, or inconsistent background structures. DDPF-PR~\cite{zhou2026high} produces competitive local details, but slight residual artifacts can still be observed around challenging motion regions. In contrast, HDRAgent generates cleaner motion boundaries and more coherent local structures, with fewer ghosting traces and results closer to the reference HDR images.

Fig.~\ref{fig:visual_comparison_tel} shows the results on Tel's dataset~\cite{tel2023alignment}, where strong foreground motion, occlusion, and saturated background regions jointly degrade the reliability of cross-exposure correspondences. DHDR~\cite{wu2018deep} and NHDRR~\cite{yan2020deep} exhibit obvious ghosting around the moving subject, indicating that conventional feature-domain alignment is insufficient when the correspondence cues are weak. AHDR~\cite{yan2019attention} and HDRTrans~\cite{liu2022ghost} suppress part of the ghosting but still produce halo-like artifacts and unstable colors near high-contrast boundaries. SCTNet~\cite{tel2023alignment} and SMHDR~\cite{ni2025semantic} improve the overall visual consistency, yet semi-transparent residual structures remain around the foreground silhouette. AFUNet~\cite{li2025afunet} and DDPF-PR~\cite{zhou2026high} achieve stronger deghosting performance, but some local regions still show slight boundary ambiguity or texture inconsistency. By comparison, HDRAgent better removes the ghosting around the moving foreground and preserves a more stable local appearance. This demonstrates that the proposed scene-adaptive routing and feedback-guided correction are effective for selecting more suitable deghosting strategies under complex motion and occlusion conditions.

Fig.~\ref{fig:visual_comparison_hu} reports qualitative results on Hu's synthetic dataset~\cite{hu2020sensor}. Although the scenes are rendered synthetically, they contain strong illumination variations, saturated regions, and fine local textures, which make HDR fusion challenging. AHDR~\cite{yan2019attention} and HDRTrans~\cite{liu2022ghost} tend to preserve the global brightness but introduce visible color shifts and local fusion artifacts in the cropped regions. SCTNet~\cite{tel2023alignment} and DiffHDR~\cite{yan2023toward} alleviate some structural distortions, but residual texture inconsistency and color bleeding can still be observed around saturated and high-contrast regions. SAFNet~\cite{kong2024safnet} shows strong detail recovery in some areas, yet it may produce over-enhanced colors or unstable local textures. LFDiff~\cite{hu2024generating} and AFUNet~\cite{li2025afunet} obtain visually plausible results, but the local structures in the enlarged regions are still less consistent with the reference. DDPF-PR~\cite{zhou2026high} further improves local fidelity, while slight fusion artifacts remain around saturated object boundaries. HDRAgent achieves cleaner local reconstruction with more faithful color transitions and fewer ghosting artifacts, indicating its robustness to both synthetic illumination changes and local correspondence failures.

Overall, these qualitative comparisons show that different HDR reconstruction methods exhibit distinct failure patterns under different dynamic conditions. Methods with fixed inference paths may perform well in some regions but still struggle when motion, saturation, and occlusion jointly cause unreliable cross-exposure correspondences. HDRAgent consistently produces more coherent local structures and fewer residual artifacts across the three datasets. These visual results are consistent with the quantitative comparisons, demonstrating that the proposed agent-driven dynamic deghosting process improves both numerical fidelity and perceptual visual quality in challenging multi-exposure HDR reconstruction.

\begin{figure*}[t]
  \centering
  \includegraphics[width=\textwidth]{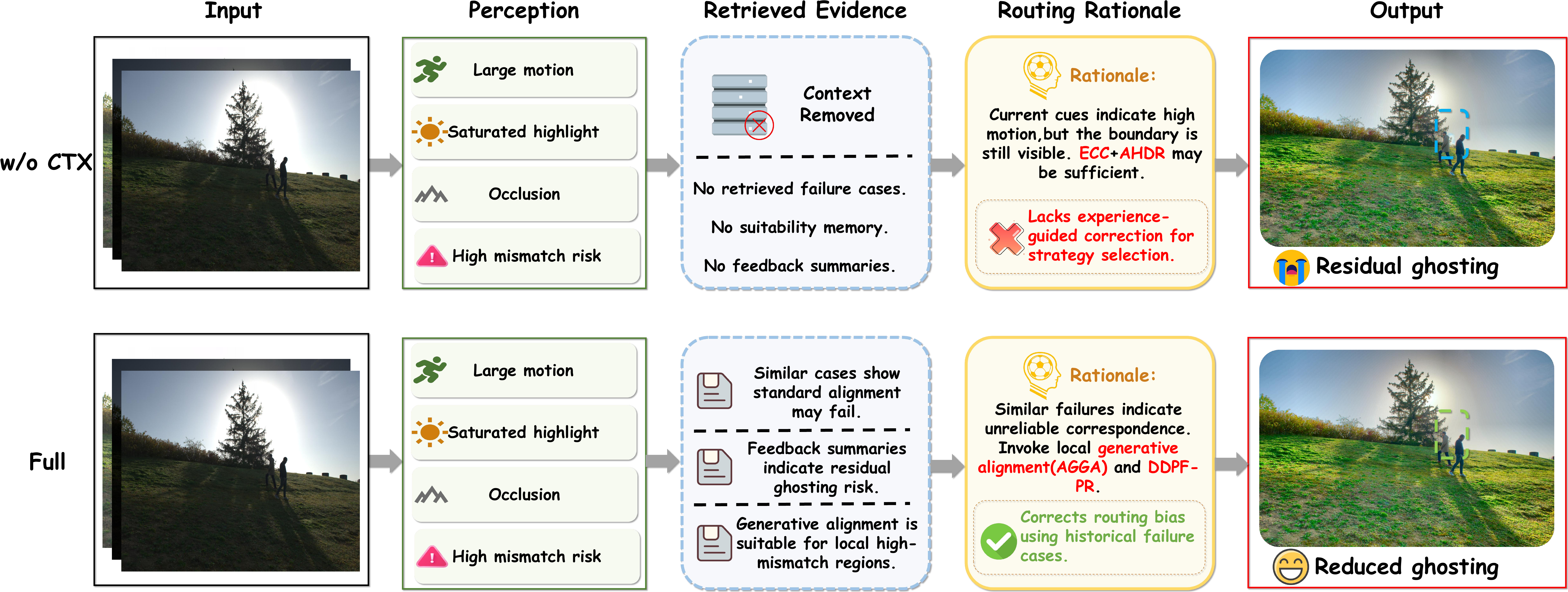}
    \caption{Ablation study on fine-grained contextual knowledge matching. The full model retrieves scene-relevant tool applicability evidence and historical feedback evidence, thereby avoiding biased strategy selection and reducing residual ghosting caused by insufficient contextual evidence.}
  \label{fig:ablation_context}
\vspace{-0.4cm}
\end{figure*}

\subsection{Ablation Study}
To analyze the contribution of each key design in HDRAgent, we conduct ablation studies on dynamic strategy selection, fine-grained contextual knowledge matching, agent-guided generative alignment, and perception--distortion feedback. The evaluated variants are summarized as follows:

\begin{itemize}
    \item \textbf{w/ RS}: A reconstruction tool is randomly selected from the HDR tool library for each input, without scene-adaptive strategy selection.

    \item \textbf{w/o CTX}: Fine-grained contextual knowledge matching is removed. The router only receives the current scene description and the basic HDR tool inventory, without scene-relevant tool applicability evidence or feedback evidence retrieved from the historical feedback memory.

    \item \textbf{w/o GA}: Agent-guided generative alignment is removed, and high-mismatch regions are reconstructed using the selected HDR reconstruction strategy.

    \item \textbf{Full}: The complete HDRAgent framework with fine-grained contextual knowledge matching, dynamic strategy selection, perception--distortion feedback, and agent-guided generative alignment.
\end{itemize}

\begin{figure}[t]
    \centering
    \includegraphics[width=\columnwidth]{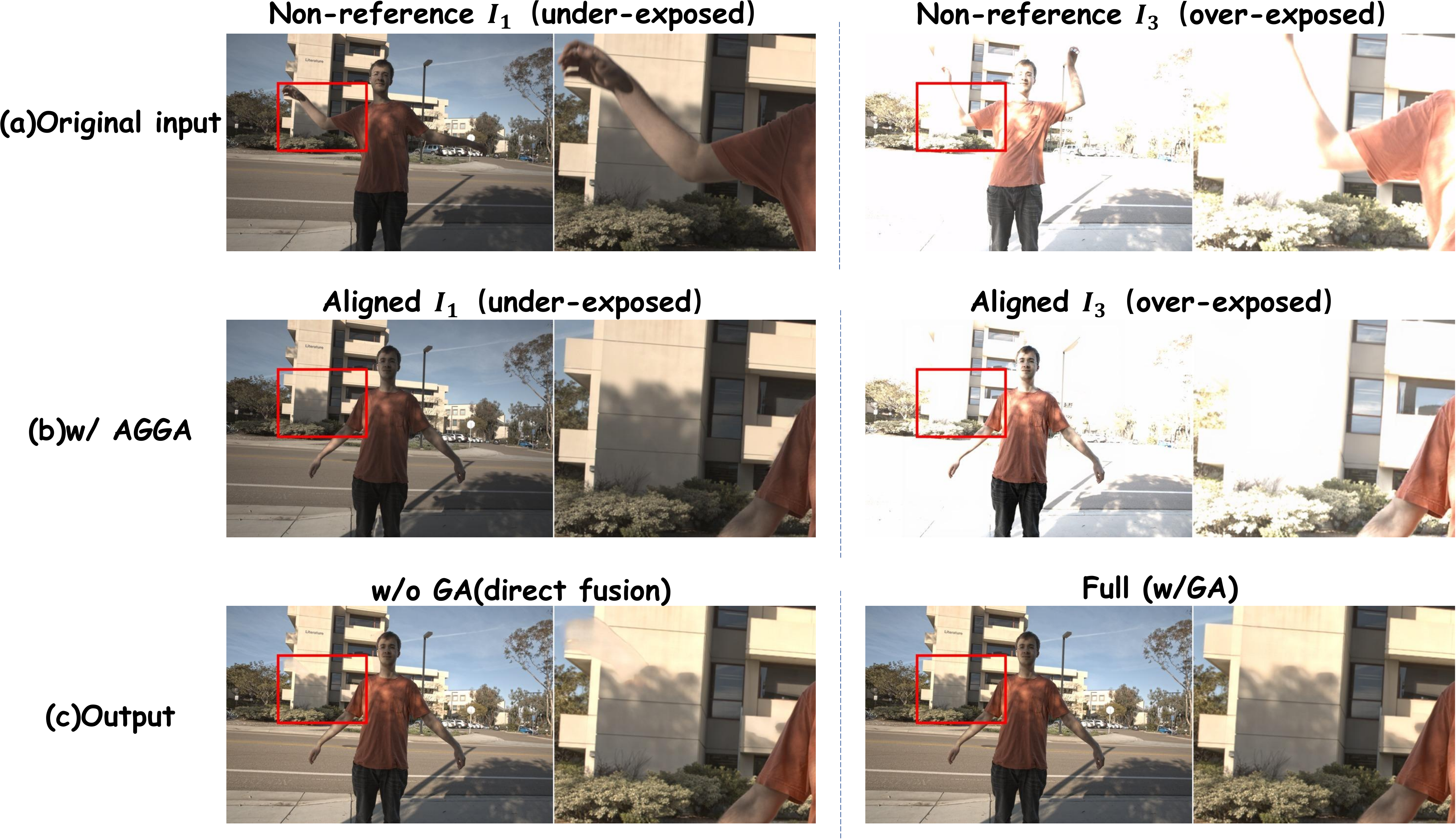}
    \caption{
    Ablation study on the agent-guided generative alignment module of HDRAgent. 
    Given two non-reference frames with under-exposure and over-exposure, AGGA updates large-motion regions before HDR fusion. 
    Without AGGA, direct fusion of the misaligned inputs produces duplicated contours and residual ghosting. 
    With AGGA, the updated non-reference frames become more structurally consistent, leading to cleaner local structures and fewer ghosting artifacts in the final HDR reconstruction.
    }
    \label{fig:ablation_ga_feedback}
    \vspace{-0.8cm}
\end{figure}

Table~\ref{tab:ablation_kalantari} reports the quantitative results of different decision-level variants, while Figs.~\ref{fig:ablation_context} and~\ref{fig:ablation_ga_feedback} provide visual comparisons. The complete HDRAgent achieves the best performance, demonstrating the effectiveness of scene-adaptive strategy selection, fine-grained contextual knowledge matching, agent-guided generative alignment, and perception--distortion feedback.

\noindent\textbf{1) Effect of Scene-Adaptive Tool Selection.}
To verify the necessity of scene-adaptive tool selection, we compare the full HDRAgent with \textbf{w/ RS}, which randomly selects a reconstruction tool from the HDR tool library for each input. As shown in Table~\ref{tab:ablation_kalantari}, \textbf{w/ RS} obtains the lowest performance among all variants, with 44.14 PSNR-$\mu$, 42.31 PSNR-$L$, 0.9917 SSIM-$\mu$, and 0.9883 SSIM-$L$. In contrast, the full HDRAgent improves these results to 45.01, 42.68, 0.9922, and 0.9911, respectively. The performance gap indicates that arbitrary tool invocation cannot reliably handle different combinations of motion, saturation, and occlusion. By selecting reconstruction tools according to the perceived scene state and context knowledge, HDRAgent achieves more robust HDR deghosting and better reconstruction fidelity.

\noindent\textbf{2) Effect of Fine-Grained Contextual Knowledge Matching.}
The variant \textbf{w/o CTX} removes fine-grained contextual knowledge matching. The router still receives the current scene description and the basic HDR tool inventory, but lacks scene-relevant tool applicability evidence and historical feedback evidence. As shown in Fig.~\ref{fig:ablation_context}, both \textbf{w/o CTX} and \textbf{Full} perceive the same challenging scene state, including large motion, saturated highlights, occlusion, and high mismatch risk. However, without contextual evidence from similar failure cases and tool applicability analysis, \textbf{w/o CTX} makes an over-optimistic routing decision and selects a conventional reconstruction path, resulting in residual ghosting around the dynamic region.

In contrast, the full HDRAgent matches the current scene evidence with the expert tool knowledge base and historical feedback memory. The retrieved evidence suggests that standard alignment is unreliable in similar large-motion and saturated regions, while AGGA is more suitable for local high-mismatch areas. Therefore, HDRAgent selects AGGA together with DDPF-PR for more reliable reconstruction, leading to clearly reduced ghosting. These results demonstrate that fine-grained contextual knowledge matching provides critical decision evidence for scene-adaptive HDR deghosting.

\noindent\textbf{3) Effect of Agent-Guided Generative Alignment.}
To evaluate the contribution of agent-guided generative alignment, we compare the full HDRAgent with \textbf{w/o GA}, where the original non-reference frames are directly processed by the selected conventional HDR reconstruction tool without reference-constrained generative alignment. As shown in Table~\ref{tab:ablation_kalantari}, removing GA degrades the reconstruction performance, indicating that conventional alignment and fusion are still insufficient for regions with severe motion and unreliable cross-exposure correspondences. 

Fig.~\ref{fig:ablation_ga_feedback} further visualizes this effect. Given the original three-frame multi-exposure input, the under-exposed non-reference frame $I_1$ and the over-exposed non-reference frame $I_3$ contain large motion around the foreground target, making them inconsistent with the reference frame $I_r$. With AGGA, HDRAgent performs reference-constrained local alignment on these non-reference frames before HDR fusion, producing aligned frames with structures more consistent with $I_r$. Without this process, \textbf{w/o GA} directly fuses the misaligned inputs and leaves duplicated contours and semi-transparent ghosting in the final HDR result. In contrast, the full model reduces local ghosting and produces cleaner structures closer to the reference HDR image. This verifies that AGGA effectively improves HDR reconstruction by correcting high-mismatch non-reference regions before fusion.

\section{Conclusion}

In this paper, we present HDRAgent, an agent-driven framework designed for robust multi-exposure HDR reconstruction in complex dynamic scenes. The cornerstone of our approach is a scene-adaptive decision-making framework that enables precise HDR tool invocation by integrating multimodal perception, fine-grained contextual knowledge matching, and perception--distortion feedback. Moreover, we propose agent-guided generative alignment to handle extreme motion and occlusion, which leverages generative priors to locally reconstruct unreliable dynamic regions. Extensive experiments demonstrate that HDRAgent effectively reduces ghosting and local fusion artifacts, achieving competitive quantitative performance and improved visual quality in challenging dynamic scenes.


{
\bibliographystyle{IEEEtran}
\bibliography{main1}
}

\vfill

\end{document}


\title{Supplementary Material}

\author{Weiyu Zhou, Tao Hu, Yijian Wang, Xiaogang Xu, Ruixing Wang, and Qingsen Yan}

\maketitle

\section{Notation Summary}
\label{sec:supp_notation}

For clarity, we summarize the key notations used in HDRAgent in Table~\ref{tab:notation}. 
These notations are used throughout the main paper and the supplementary material to describe the input sequence, contextual evidence, tool selection, reconstruction result, and feedback memory.

\begin{table}[!ht]
\centering
\caption{Summary of key notations used in HDRAgent.}
\label{tab:notation}
\renewcommand{\arraystretch}{1.25}
\setlength{\tabcolsep}{4pt}
\footnotesize
\renewcommand{\tabularxcolumn}[1]{m{#1}}
\begin{tabularx}{\columnwidth}{
>{\centering\arraybackslash}m{0.22\columnwidth}|
>{\raggedright\arraybackslash}X
}
\toprule
\textbf{Notation} & \textbf{Description} \\
\midrule
$\mathcal{I}=\{I_i\}_{i=1}^{N}$ 
& Multi-exposure LDR input sequence. \\
\hline
$\mathcal{E}=\{e_i\}_{i=1}^{N}$ 
& Exposure information of the input sequence. \\
\hline
$S$ 
& MLLM-derived scene-state description. \\
\hline
$K$ 
& Expert knowledge base of candidate reconstruction tools. \\
\hline
$\mathcal{H}_{t-1}$ 
& Historical feedback memory before round $t$. \\
\hline
$C_t$ 
& Contextual evidence generated by FCM, including scene evidence $C_t^{\mathrm{scene}}$, tool evidence $C_t^{\mathrm{tool}}$, and feedback evidence $C_t^{\mathrm{fb}}$. \\
\hline
$\mathcal{T}$ 
& Candidate tool set, including alignment and fusion/reconstruction tools. \\
\hline
$R_t \in \mathcal{T}$ 
& Selected reconstruction strategy from the candidate tool set $\mathcal{T}$ at round $t$. \\
\hline
$\hat{H}_t$ 
& Reconstructed HDR result at round $t$. \\
\hline
$F_t$ 
& Structured feedback generated by the perception--distortion evaluator. \\
\bottomrule
\end{tabularx}
\end{table}

\section{Candidate Alignment and Fusion Tool Library}
\label{sec:supp_tool_library}

HDRAgent is built upon an extensible tool library rather than a fixed reconstruction pipeline. 
The current tool library consists of candidate alignment tools and candidate fusion/reconstruction tools. 
Given the contextual evidence produced by FCM, HDRAgent can invoke different tools according to the scene condition and reconstruction requirement. 
The library can be readily extended by adding new executable tools, including traditional alignment algorithms, optical-flow estimators, HDR merging operators, exposure fusion methods, learning-based HDR reconstruction networks, and generative restoration models. 
For learning-based tools, we only consider methods with publicly available code and pretrained weights, or methods implemented and released by ourselves. 
Tables~\ref{tab:alignment_tools} and~\ref{tab:fusion_tools} summarize the candidate tools used in the current implementation.

\begin{table*}[!t]
\centering
\caption{Summary of candidate alignment tools used in HDRAgent.}
\label{tab:alignment_tools}
\footnotesize
\setlength{\tabcolsep}{4pt}
\renewcommand{\arraystretch}{1.15}
\begin{tabular}{
>{\centering\arraybackslash}m{0.24\linewidth}|
>{\centering\arraybackslash}m{0.36\linewidth}|
>{\centering\arraybackslash}m{0.32\linewidth}
}
\toprule
Tool category & Representative tools & Tool source \\
\midrule
Global / parametric alignment
& ECC
& OpenCV \\
\midrule
Exposure-robust alignment
& MTB
& OpenCV \\
\midrule
Sparse feature-based alignment
& SIFT/ORB + RANSAC
& OpenCV \\
\midrule
Traditional dense optical flow
& Farnebäck optical flow, DIS optical flow
& OpenCV \\
\midrule
Learning-based dense correspondence
& RAFT, PWC-Net
& Open-source code and pretrained weights \\
\midrule
Agent-guided generative alignment
& AGGA
& Proposed in this work \\
\bottomrule
\end{tabular}
\end{table*}

\begin{table*}[!t]
\centering
\caption{Summary of candidate fusion and reconstruction tools used in HDRAgent.}
\label{tab:fusion_tools}
\footnotesize
\setlength{\tabcolsep}{4pt}
\renewcommand{\arraystretch}{1.15}
\begin{tabular}{
>{\centering\arraybackslash}m{0.26\linewidth}|
>{\centering\arraybackslash}m{0.42\linewidth}|
>{\centering\arraybackslash}m{0.24\linewidth}
}
\toprule
Tool category & Representative tools & Tool source \\
\midrule
Traditional weighted fusion
& Hat-function weighted fusion
& Hand-crafted implementation \\
\midrule
Radiance-domain HDR merging
& Debevec--Malik, Robertson
& OpenCV / hand-crafted implementation \\
\midrule
Traditional exposure fusion
& Mertens exposure fusion, Laplacian-pyramid exposure fusion
& OpenCV / hand-crafted implementation \\
\midrule
Learning-based HDR fusion / reconstruction
& DHDR, AHDR, NHDRR, HDR-GAN, CA-ViT, SCTNet, DiffHDR, SAFNet, LFDiff, AFUNet, DDPF-PR
& Open-source code and pretrained models / our open-source implementation \\
\bottomrule
\end{tabular}
\end{table*}

\section{Example of Fine-Grained Contextual Knowledge Matching}
\label{sec:supp_fcm_example}

To better illustrate the output form of the proposed fine-grained contextual knowledge matching (FCM) module, we provide a representative example of the matched contextual evidence.
Given the MLLM-derived scene description, FCM organizes the current scene condition, relevant reconstruction-tool knowledge, and matched historical feedback cases into structured contextual evidence for dynamic strategy selection.
The contextual evidence consists of three parts: scene evidence $C_t^{\mathrm{scene}}$, tool evidence $C_t^{\mathrm{tool}}$, and feedback evidence $C_t^{\mathrm{fb}}$.
Table~\ref{tab:fcm_example} presents an illustrative example.

\begin{table*}[!t]
\centering
\caption{Illustrative example of the structured contextual evidence produced by FCM.}
\label{tab:fcm_example}
\footnotesize
\setlength{\tabcolsep}{4pt}
\renewcommand{\arraystretch}{1.18}
\begin{tabular}{
>{\centering\arraybackslash}m{0.18\linewidth}|
>{\raggedright\arraybackslash}m{0.76\linewidth}
}
\toprule
\centering Component & Example output \\
\midrule

MLLM-derived scene description
& The scene contains noticeable foreground motion around the human body and arms, strong exposure variation across the three inputs, severe saturation in the face and bright background regions, and unreliable local correspondence near the moving foreground boundaries.
Background structures such as cars and building edges are relatively stable, while the facial region and saturated highlights contain missing or weakly observable textures. \\

\midrule

$C_t^{\mathrm{scene}}$
& \textbf{Motion:} the foreground person and arms show noticeable displacement across exposures.
\textbf{Exposure variation:} the low-exposure input preserves part of the bright background but loses body and shadow details, while the high-exposure input reveals darker regions but saturates the face and background highlights.
\textbf{Saturation:} severe over-exposure appears in the face, sky, building wall, and local shirt regions.
\textbf{Foreground--background ambiguity:} moving body contours and arm boundaries introduce local mismatch between the foreground person and the background.
\textbf{Texture observability:} facial details are unreliable in the high-exposure input and weakly visible in the low-exposure input.
\textbf{Fusion risk:} direct fusion may introduce ghosting around the arms and body contour, as well as color inconsistency in saturated facial and background regions. \\

\midrule

$C_t^{\mathrm{tool}}$
& \textbf{ECC / MTB:} relevant mainly for globally stable background regions, such as cars, building edges, and parking-lot structures, where the dominant background motion is mild.
They are not reliable for the moving person, saturated face, or non-rigid foreground boundaries.
\textbf{RAFT / DIS:} relevant for dense alignment in textured and observable regions, such as visible body or background areas between low- and medium-exposure inputs.
Their reliability decreases in saturated facial regions, over-exposed highlights, and texture-missing areas.
\textbf{AHDR / DHDR:} relevant as alignment--fusion reconstruction tools for moderately dynamic regions where learned attention or fusion can suppress part of the misaligned non-reference information, but they may still leave ghosting under large foreground displacement and severe saturation.
\textbf{CA-ViT / SCTNet / AFUNet:} relevant for large foreground motion and contextual ambiguity, since stronger contextual modeling or alignment--fusion coupling can better handle non-local dependencies around the moving person and background.
\textbf{DDPF-PR:} relevant for difficult dynamic regions requiring prior-guided reconstruction and perceptual refinement, especially where conventional fusion produces local artifacts.
\textbf{AGGA:} relevant for the saturated face and severely unreliable foreground regions, where valid cross-exposure content is missing or correspondence-based alignment becomes unreliable. \\

\midrule

$C_t^{\mathrm{fb}}$
& \textbf{Matched historical case 1:} in scenes with stable outdoor backgrounds and mild camera motion, ECC/MTB-based background alignment followed by learned fusion preserved cars and building structures well, but did not correct ghosting on independently moving foreground persons.
\textbf{Matched historical case 2:} in cases with a moving person and saturated facial highlights, RAFT/DIS-based dense alignment improved textured body regions but produced unreliable warping around over-exposed faces and arms due to missing texture and non-rigid motion.
\textbf{Matched historical case 3:} in scenes with large foreground displacement, AHDR/DHDR-style fusion reduced part of the exposure inconsistency but left residual ghosting near body contours when the foreground--background correspondence was ambiguous. \\

\bottomrule
\end{tabular}
\end{table*}